\documentclass[11pt]{article}

\usepackage[preprint]{acl}

\usepackage{times}
\usepackage{latexsym}

\usepackage[T1]{fontenc}

\usepackage[utf8]{inputenc}
\usepackage{amsmath,amssymb}
\usepackage{booktabs}
\usepackage{microtype}
\usepackage{url}
\usepackage{inconsolata}
\usepackage{multirow}
\usepackage{hyperref}

\usepackage{graphicx}
\usepackage{algorithm}
\usepackage{algpseudocode}
\usepackage{pgfplots}
\pgfplotsset{compat=1.18}
\usepackage{subcaption}
\usepackage{array}
\title{TIGER: Text-conditioned Visual Gated Routing with Acceptance Alignment for Multimodal Speculative Decoding}

\author{
  Quynh Vo \quad Cong-Duy Nguyen \quad Ponhvoan Srey \quad Luu Anh Tuan \quad \textbf{Thong Nguyen}\thanks{Corresponding author} 
  \\
  National University of Singapore \quad Nanyang Technological University \\ Center of AI Research, VinUniversity \\ \href{mailto:thong.nguyen@u.nus.edu}{thong.nguyen@u.nus.edu}}

\begin{document}
\maketitle
\begin{abstract}

Speculative decoding accelerates autoregressive generation by letting a lightweight drafter propose multiple tokens that are verified by a larger target model. Although effective for text-only LLMs, speculative decoding yields limited gains in VLMs because drafters often diverge on vision-critical content, while existing multimodal acceleration methods do not directly address irrelevant visual evidence or optimize the verifier-accepted prefix length that governs speedup. We propose TIGER, a Text-conditioned vIsual GatEd Routing framework for multimodal speculative decoding. TIGER dynamically selects a sparse set of context-relevant visual tokens based on the drafter’s current textual state, rather than expose the full visual token set or a fixed compressed interface. To better align training with inference-time efficiency, we optimize the drafter with acceptance-aligned group-based policy training using verifier-derived rewards based on accepted prefix length, built on top of distillation warm start with KL anchoring. This encourages the drafter not only to imitate the target model, but also to produce speculative continuations that survive verification for longer prefixes. Experiments show that TIGER yields consistent gains in accepted prefix length and speculative speedup under exact verifier-side speculative decoding, while achieving favorable quality-latency trade-offs with comparable downstream accuracy in visual-routing analyses.
\end{abstract}

\section{Introduction}

\textit{Vision-language models} (VLMs) have rapidly become the backbone of multimodal assistants~\citep{10.5555/3666122.3667638,10.5555/3618408.3619222}, enabling instruction following over images and videos for tasks such as visual question answering, document understanding, chart reasoning, and grounded dialogue~\citep{10.5555/3737916.3740705}. Despite steady progress in model quality, real-world deployment remains constrained by \emph{autoregressive decoding} cost: generating outputs token-by-token requires repeated passes through large decoders and extensive key--value (KV) caching~\citep{DBLP:conf/iclr/SadhukhanCCTLSY25}. As VLMs scale to higher resolutions and richer visual contexts, decoding latency and serving cost increasingly dominate practical deployment.

\emph{Speculative decoding} offers a principled, lossless route to accelerate generation. A lightweight drafter proposes multiple tokens, and a large verifier checks and accepts a prefix of these proposals, thereby reducing the number of expensive verifier decoding steps while preserving the target distribution through a modified rejection-sampling test~\citep{chen2023speculative}. In text-only LLMs, speculative decoding can provide substantial speedups when the drafter closely matches the target model, and recent work further improves efficiency through better drafting structures such as dynamic draft trees~\citep{li2024eagle2}. However, directly transferring this paradigm to VLMs often yields limited gains.

The main challenge is that multimodal speculative decoding is highly sensitive to vision-critical tokens---for example, OCR strings in TextVQA-style queries~\citep{singh2019textvqa}, numerical answers in chart or science reasoning~\citep{lu2022scienceqa}, compositional relations in GQA~\citep{hudson2019gqa}, and grounded object descriptions where hallucination must be avoided~\citep{li2023pope}. On such tokens, a small drafter often diverges early from the verifier, causing short accepted prefixes and sharply reducing speculative speedup. We argue that this failure stems from two coupled mismatches. First, the drafter is often exposed to either the full visual token set or a fixed compressed visual interface, which may still contain irrelevant or noisy visual evidence for the current decoding context. Second, most existing training strategies optimize imitation quality rather than the runtime quantity that actually governs speculative efficiency: the verifier-accepted prefix length.

These observations motivate a different perspective on multimodal speculative decoding. Instead of only asking how to compress visual inputs for a small drafter, we ask: what visual evidence does the drafter actually need for the next speculative block, and how should it be trained so that the verifier accepts that block for as long as possible? Based on this view, we propose \textbf{TIGER} (\textbf{T}ext-conditioned v\textbf{I}sual \textbf{G}at\textbf{E}d \textbf{R}outing), a framework for multimodal speculative decoding built on two key ideas. First, rather than exposing the drafter to the full image token set or a fixed compressed adaptor, TIGER dynamically selects a sparse subset of context-relevant visual evidence conditioned on the drafter's current textual state. This yields a lightweight, adaptive visual interface tailored to the next speculative block. Second, rather than training the drafter only to imitate the target model, TIGER directly aligns the drafter with the verifier's acceptance mechanism.

Concretely, we first warm-start the drafter with knowledge distillation to provide a stable behavioral initialization. We then formulate drafter learning as an acceptance alignment problem: for each input, we sample multiple draft continuations, run verifier-based acceptance tests, and assign rewards based on verifier-accepted prefix length. This enables group-based policy optimization using verifier-derived supervision, without human preference annotations or learned reward models. In effect, the drafter is encouraged not merely to match the target distribution in aggregate, but to generate speculative blocks that the verifier can accept for longer prefixes. To ensure stable training, we employ KL anchoring to prevent policy drift, and we design the routing and verification pipeline to remain compatible with efficient speculative decoding at inference time.

We evaluate TIGER on a diverse suite of multimodal benchmarks spanning OCR-heavy, counting-intensive, and visually grounded reasoning tasks. Experiments show that TIGER consistently lengthens verifier-accepted prefixes, reduces target-model decoding work, and improves end-to-end speculative decoding speedup over strong multimodal speculative baselines, while preserving downstream task accuracy. Our contributions are summarized as follows:



\begin{itemize}
    \item We propose \textsc{TIGER}, a text-conditioned visual gated routing mechanism that dynamically selects context-relevant visual evidence for the drafter in each speculative block.
    \item We introduce an acceptance-aligned training objective based on verifier-accepted prefix length, directly targeting the runtime quantity that governs speculative decoding speedup.
    \item We conduct experiments that explicitly distinguish exact verifier-side speculative decoding from auxiliary visual-routing analyses, showing improved accepted prefix length and speedup in the former and favorable quality--latency trade-offs in the latter.
\end{itemize}

\section{Related Works}

\subsection{Inference-Time Acceleration via Visual Token Pruning}

A major efficiency bottleneck in VLMs is the large number of visual tokens, which increases prefill computation and KV-cache traffic. Prior work therefore accelerates VLM inference by pruning or compressing visual tokens. \citet{xing2024pyramiddrop} progressively removes image tokens across layers, while \textsc{SparseVLM} performs text-guided token sparsification with adaptive sparsity~\citep{zhang2024sparsevlm}. \textsc{VisPruner} instead selects salient and diverse visual tokens using visual cues, avoiding reliance on potentially unstable text--visual attention signals~\citep{zhang2025vispruner}. Related pruning ideas have also been extended to specialized multimodal tasks such as reasoning segmentation~\citep{10.1145/3746027.3755243}.

These methods primarily reduce the amount of visual computation processed by the model. TIGER is complementary: rather than globally pruning or compressing visual tokens, it performs \emph{text-conditioned visual routing} for speculative decoding, exposing the drafter to a sparse subset of context-relevant visual evidence for each draft block. This targets drafter--verifier mismatch on vision-critical continuations, a bottleneck not directly addressed by standard visual-token pruning methods.

\subsection{Speculative Decoding for Multimodal Models}

Speculative decoding accelerates generation by using a lightweight drafter to propose multiple tokens that are verified by a larger target model, yielding lossless sampling under the verifier distribution~\citep{chen2023speculative}. Prior work improves drafting and verification through token trees, dynamic draft structures, direct token prediction, and serving-oriented optimizations, including \textsc{SpecInfer}, \textsc{EAGLE-2}, \textsc{EAGLE-3}, \textsc{RACER}, \textsc{TokenRecycling}, and \textsc{vLLM}~\citep{miao2024specinfer,li2024eagle2,li2025eagle3,zhang2026racer,luo2025tokenrecycling,kwon2023vllm}. However, these methods are mostly developed for text-only or general LLM serving and do not directly address multimodal draft--verifier divergence on vision-critical tokens.

In VLMs, draft tokens can be rejected early when the drafter diverges on OCR strings, counts, chart values, or visually grounded entities, leading to short accepted prefixes and limited speedup. Recent multimodal methods such as \textsc{SpecFLASH} and \textsc{ViSpec} improve draft--verifier coherence through latent-aware drafting or vision-aware adaptation~\citep{wang2025flash,kang2025vispec}. TIGER differs by dynamically routing context-relevant visual evidence to the drafter and by optimizing a verifier-derived accepted-prefix objective, aligning both visual access and training with verifier acceptance rather than token imitation alone.

\section{Method}

\begin{figure*}[t]
    \centering
    \includegraphics[width=1\linewidth]{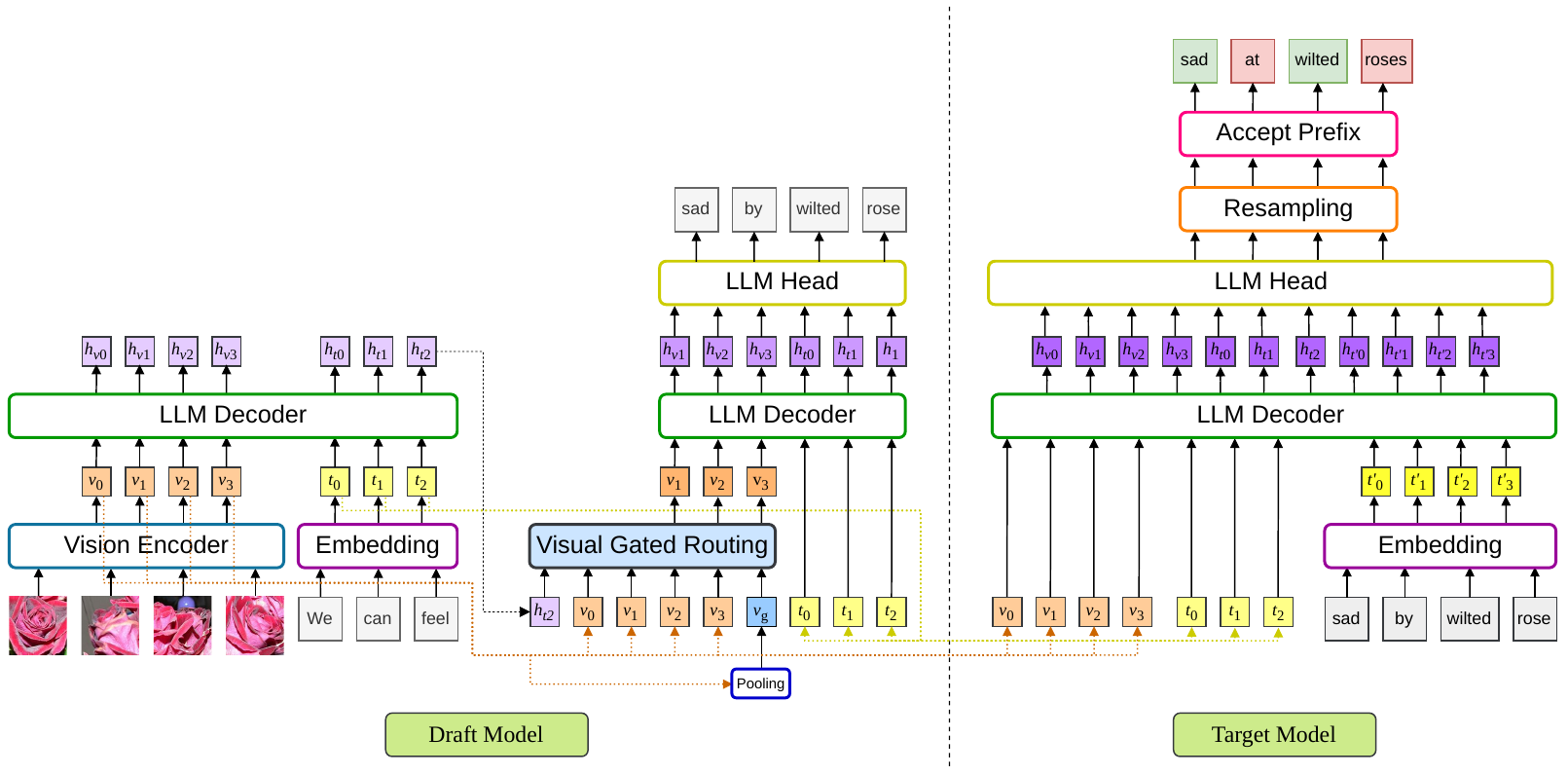}
    \caption{\textbf{Overview of TIGER for multimodal speculative decoding.}
    Given an image and a textual prefix, TIGER first performs text-conditioned visual routing to select a sparse subset of relevant visual tokens for the drafter. Conditioned on this routed visual context, the drafter proposes a speculative block of \(\gamma\) tokens. The verifier, which always operates on the full visual context, then accepts a prefix of the proposed block under the standard speculative decoding test. TIGER improves efficiency by increasing verifier-accepted prefix length while keeping the drafter-side visual interface lightweight.}
    \label{fig:tiger-overview}
\end{figure*}

Figure~\ref{fig:tiger-overview} provides an overview of TIGER. At each speculative decoding round, TIGER first routes a sparse subset of context-relevant visual tokens for the drafter, then optimizes the drafter with an acceptance-aligned objective so that the verifier accepts longer prefixes.

\subsection{Problem Setup}
We consider multimodal speculative decoding with a lightweight drafter \(q_\theta\) and a large verifier \(p_\phi\). Given an input image \(I\) and a textual prefix \(T_{1:j}\), a vision encoder produces a visual token bank
\begin{equation}
V = E_v(I) = [v_1,\dots,v_r] \in \mathbb{R}^{r \times d},
\label{eq:visual-bank}
\end{equation}
where \(r\) is the number of visual tokens and \(d\) is the hidden dimension.

Let \(E_t(T_{1:j})\) denote the embedded textual prefix, and let
\(V_j^{\mathrm{dra}}\subseteq V\) denote the visual interface exposed to the drafter at speculative round \(j\). 
For a full-context drafter, \(V_j^{\mathrm{dra}}=V\); in TIGER, this interface will be instantiated as the routed visual context \(V_j^{\mathrm{route}}\) in Section~\ref{sec:tiger-routing}. The drafter input is
\begin{equation}
X_j = [E_t(T_{1:j}); V_j^{\mathrm{dra}}],
\label{eq:drafter-input}
\end{equation}
and the drafter proposes a block of \(\gamma\) future tokens,
\begin{equation}
\hat T_{j+1:j+\gamma} \sim q_\theta(\cdot \mid X_j).
\label{eq:draft-block}
\end{equation}
The proposed block is then evaluated by the verifier under the standard speculative decoding acceptance test using the full visual token bank \(V\). Let \(a_u \in \{0,1\}\) indicate whether the \(u\)-th drafted token is accepted. Since speculative decoding accepts only a prefix of the proposed block, we define the verifier-accepted prefix length as
\begin{equation}
A(x;\theta)
=
\sum_{u=1}^{\gamma}\prod_{m=1}^{u} a_m,
\qquad x=(I,T_{1:j}).
\label{eq:accepted-prefix}
\end{equation}
This quantity measures how many drafted tokens survive verification in one speculative round.

Our goal is to train the drafter to maximize expected verifier-accepted prefix length while maintaining a lightweight visual interface. TIGER achieves this through text-conditioned visual gated routing and acceptance-aligned optimization.

\subsection{TIGER: Text-conditioned Visual Gated Routing}
\label{sec:tiger-routing}

\begin{figure}[t]
    \centering
    \includegraphics[width=0.85\columnwidth]{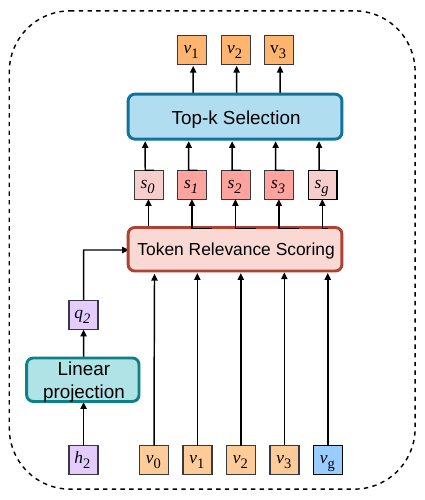}
    \caption{\textbf{TIGER visual gated routing.}
    A routing query derived from the current prefix hidden state scores each token in the visual token bank, and the top-\(k\) most relevant tokens are selected as the routed visual context for the next speculative block.}
    \label{fig:tiger-routing}
\end{figure}


Figure~\ref{fig:tiger-routing} illustrates the central idea of TIGER: dynamically routing visual evidence according to the current decoding context. At each step, a query computed from the hidden state of the textual prefix scores the visual token bank, and the top-\(k\) most relevant tokens are selected as routed visual context for the next speculative block. This mechanism is motivated by the observation that, in multimodal speculative decoding, the useful visual evidence is highly context-dependent. 
Using all visual tokens can be costly and may introduce redundant evidence, while a fixed compressed representation may miss cues needed for the current prediction. TIGER therefore performs adaptive visual token selection to provide the drafter with compact yet contextually relevant visual information.

Starting from the visual bank in Eq.~\eqref{eq:visual-bank}, we optionally append a pooled global token \(v_{\mathrm g}\), yielding
\begin{equation}
\bar V = [v_1,\dots,v_r,v_{\mathrm g}] \in \mathbb{R}^{(r+1)\times d}.
\label{eq:aug-visual-bank}
\end{equation}
Let \(h_j \in \mathbb{R}^{d}\) denote the drafter hidden state at the end of the current textual prefix. TIGER projects this state into a routing query
\begin{equation}
q_j = W_q h_j,
\label{eq:routing-query}
\end{equation}
where \(W_q \in \mathbb{R}^{d\times d}\) is a learnable projection. For each visual token \(\bar v_i \in \bar V\), we compute a relevance score
\begin{equation}
s_i = \frac{q_j^\top W_v \bar v_i}{\sqrt d},
\label{eq:routing-score}
\end{equation}
where \(W_v \in \mathbb{R}^{d\times d}\) is another learnable projection. Intuitively, this score measures how well each visual token matches the current decoding state: tokens with larger scores are expected to provide more useful visual evidence for predicting the next speculative block. TIGER then performs sparse routing by selecting the top-\(k\) visual tokens:
\begin{equation}
\mathcal S_j = \operatorname{TopK}(s_1,\dots,s_{r+1};k),
\label{eq:topk}
\end{equation}
and forms the routed visual context
\begin{equation}
V^{\mathrm{route}}_j = \{\bar v_i \mid i \in \mathcal S_j\}.
\label{eq:routed-context}
\end{equation}


The routed context is recomputed once per speculative round and reused for the entire draft block. In TIGER, the drafter-side visual interface in Eq.~\eqref{eq:drafter-input} is instantiated as
\(V_j^{\mathrm{dra}}=V_j^{\mathrm{route}}\). Thus, instead of conditioning on the full visual token set, the drafter generates the next speculative block from the routed multimodal context
\begin{equation}
X_j = [E_t(T_{1:j}); V^{\mathrm{route}}_j],
\qquad
\hat T_{j+1:j+\gamma} \sim q_\theta(\cdot \mid X_j),
\label{eq:routed-draft}
\end{equation}
where \(E_t(\cdot)\) is the text embedding function. In this way, TIGER provides a lightweight and context-adaptive visual interface for drafting.

\subsection{Acceptance-Aligned Training}
We train TIGER in two stages. We first warm-start the drafter with knowledge distillation (KD), obtaining a stable reference policy \(q_{\mathrm{ref}}\). This warm start improves optimization stability, but it is not our final objective. Our main training stage instead aligns the drafter with the verifier's acceptance outcome.

Given a training example \(x=(I,T_{1:j})\), TIGER first constructs the routed multimodal context \(X_j\) as in Eq.~\eqref{eq:routed-draft}. Conditioned on \(X_j\), the drafter samples \(K\) candidate speculative blocks,
\begin{equation}
\hat Y^{(1)},\dots,\hat Y^{(K)}
\sim q_\theta(\cdot \mid X_j),
\label{eq:k-samples}
\end{equation}
where \(\hat Y^{(k)}\) denotes the \(k\)-th sampled block of length \(\gamma\). Each candidate is then evaluated by the verifier under the standard speculative decoding acceptance test. Let \(a_u^{(k)} \in \{0,1\}\) indicate whether the \(u\)-th token in \(\hat Y^{(k)}\) is accepted. The accepted prefix length of the \(k\)-th candidate is
\begin{equation}
A^{(k)} =
\sum_{u=1}^{\gamma}\prod_{m=1}^{u} a_m^{(k)}.
\label{eq:accepted-prefix-k}
\end{equation}
We directly use this verifier-accepted prefix length as reward,
\begin{equation}
R^{(k)} = A^{(k)}.
\label{eq:reward}
\end{equation}
so that drafts that survive verification for longer prefixes receive larger rewards.

To optimize the drafter, we adopt a group-based relative policy update over the \(K\) sampled candidates. Rewards are normalized within the current group:
\begin{equation}
\hat g^{(k)} = \frac{R^{(k)} - \mu_R}{\sigma_R + \epsilon},
\label{eq:advantage}
\end{equation}
where \(\mu_R\) and \(\sigma_R\) are the mean and standard deviation of \(\{R^{(1)},\dots,R^{(K)}\}\). We refer to this stage as
\emph{acceptance-aligned optimization}, because the policy update is driven by
verifier-accepted prefix length rather than token-level imitation alone. The
drafter is then optimized with the following GRPO-style objective:
\begin{equation}
\begin{aligned}
\mathcal L_{\mathrm{accept}}
&=
-\sum_{k=1}^{K} \hat g^{(k)}
\log q_\theta\!\left(\hat Y^{(k)} \mid X_j\right) \\
&\quad + \beta\,\mathrm{KL}(q_\theta \,\|\, q_{\mathrm{ref}}),
\end{aligned}
\label{eq:grpo}
\end{equation}
where \(\beta\) controls the strength of the KL anchor.

Intuitively, this acceptance-aligned objective shifts the drafter toward
speculative blocks that the verifier can accept for longer prefixes. Unlike one-to-one imitation, the group-based formulation learns relative preferences among multiple candidate continuations under the verifier's acceptance mechanism. The full training pipeline is summarized in Algorithm~\ref{alg:tiger-training}.

\begin{algorithm}[t]
\caption{Acceptance-aligned training for TIGER}
\label{alg:tiger-training}
\begin{algorithmic}[1]
\Require Drafter \(q_\theta\), verifier \(p_\phi\), training set \(\mathcal D\), group size \(K\)
\State Warm-start \(q_\theta\) with KD to obtain \(q_{\mathrm{ref}}\)
\For{each \(x=(I,T_{1:j}) \in \mathcal D\)}
    \State Build \(V_j^{\mathrm{route}}\) via TIGER routing and set \(X_j=[E_t(T_{1:j});V_j^{\mathrm{route}}]\)
    \State Sample \(K\) speculative blocks \(\hat Y^{(1)},\dots,\hat Y^{(K)} \sim q_\theta(\cdot \mid X_j)\)
    \State Evaluate each block with the verifier and obtain rewards \(R^{(k)} = A^{(k)}\)
    \State Normalize rewards to \(\hat g^{(k)}\)
    \State Update \(q_\theta\) with the GRPO objective in Eq.~\eqref{eq:grpo}
\EndFor
\end{algorithmic}
\end{algorithm}

\subsection{Efficiency Rationale}\label{sec:efficiency_rationale}
TIGER addresses two coupled bottlenecks in multimodal speculative decoding. The first is \emph{visual mismatch}: a small drafter may condition on excessive or irrelevant visual evidence, causing early divergence on vision-critical tokens. The second is \emph{objective mismatch}: distillation improves average imitation quality, whereas speculative decoding speed is governed more directly by verifier-accepted prefix length.

Let \(A_i\) denote the accepted prefix length at speculative round \(i\), and let \(\bar A = \mathbb{E}[A_i]\). To generate a response of \(N\) tokens, the expected number of verification rounds is approximately
\begin{equation}
\mathbb{E}[M] \approx \frac{N}{\bar A}.
\label{eq:rounds}
\end{equation}
Hence, increasing \(\bar A\) directly reduces the number of expensive verifier decoding steps. TIGER improves this quantity from both sides: text-conditioned routing increases the likelihood that the drafter accesses the right visual evidence for the next block, while acceptance-aligned optimization explicitly favors continuations that survive verifier checking longer. As a result, TIGER targets the runtime quantity that most directly governs speculative decoding efficiency in VLMs, namely verifier-accepted prefix length.

\noindent\textbf{Routing overhead and cache reuse.}
TIGER does not re-encode the image at every speculative round. 
The visual token bank and projected visual representations are computed once and cached. 
At each round, TIGER computes only the current text-conditioned query, scores the cached visual tokens, and performs Top-\(k\) selection. 
The routed visual context is reused for the entire draft block. 
Importantly, routing affects only the drafter-side visual interface: the verifier remains unchanged, uses the full visual context, and follows the standard speculative decoding acceptance protocol. 
Thus, verifier-side KV-cache reuse is the same as in standard speculative decoding, while the reported end-to-end latency includes routing overhead.

\section{Experiments \& Discussion}
\subsection{Experimental Setup}
\noindent\textbf{Hardware.}
All experiments are conducted on a server equipped with 8 NVIDIA A40 GPUs (48GB each). Training is performed in mixed precision (bfloat16), with gradient accumulation to support large multimodal models.

\noindent\textbf{Models.}
We evaluate TIGER on several open-source VLMs spanning the LLaVA and Qwen-VL families: 
\href{https://huggingface.co/liuhaotian/llava-v1.6-vicuna-7b}{LLaVA-v1.6-Vicuna-7B},
\href{https://huggingface.co/liuhaotian/llava-v1.6-vicuna-13b}{LLaVA-v1.6-Vicuna-13B},
\href{https://huggingface.co/Qwen/Qwen3-VL-4B-Instruct}{Qwen3-VL-4B-Instruct},
and \href{https://huggingface.co/Qwen/Qwen3-VL-8B-Instruct}{Qwen3-VL-8B-Instruct}.


\noindent\textbf{Baselines.}
We compare TIGER with speculative decoding and visual token efficiency baselines. Speculative baselines include \textsc{ViSpec}~\citep{kang2025vispec}, \textsc{SpecFLASH}~\citep{wang2025flash}, \textsc{Medusa}~\citep{10.5555/3692070.3692273}, \textsc{EAGLE-2}~\citep{li2024eagle2}, \textsc{EAGLE-3}~\citep{li2025eagle3}, \textsc{RACER}~\citep{zhang2026racer}, and \textsc{TokenRecycling}~\citep{luo2025tokenrecycling}. For training-free or plug-and-play methods such as \textsc{RACER} and \textsc{TokenRecycling}, we instrument their draft candidates under the adapted multimodal verifier-side acceptance protocol when reporting AccLen; implementation details are provided in Appendix~\ref{app:baseline-adaptation}. We also compare with visual token pruning baselines, \textsc{SparseVLM}~\citep{zhang2024sparsevlm} and \textsc{VisPruner}~\citep{zhang2025vispruner}, for which we report task performance and latency-based efficiency rather than accepted prefix length. Detailed adaptation procedures are provided in Appendix~\ref{app:baseline-adaptation}.

\noindent\textbf{Datasets.}
We evaluate TIGER on a diverse set of multimodal benchmarks covering visual question answering, reasoning, OCR understanding, and captioning, including MMBench~\citep{liu2023mmbench}, ScienceQA~\citep{lu2022scienceqa}, TextVQA~\citep{singh2019textvqa}, MM-Vet~\citep{yu2023mmvet}, and COCO Captions~\citep{chen2015microsoftcococaptionsdata}. These benchmarks assess perception, knowledge grounding, diagram-based reasoning, text understanding in natural images, and open-ended generation. To further examine context-heavy settings, we include additional evaluations on Video-MME~\citep{fu2025videomme}, NLVR2~\citep{suhr2019nlvr2}, and MuirBench~\citep{wang2025muirbench}, covering video understanding and multi-image reasoning. Detailed dataset splits, sampling protocols, and evaluation details are provided in Appendix~\ref{app:dataset-details}.

\noindent\textbf{Evaluation Metrics.}
We evaluate both task performance and decoding efficiency. 
For speculative decoding, we report \textbf{Accepted Prefix Length} (\textbf{AccLen}), \textbf{Verifier Rounds} (\textbf{VerRds}), \textbf{End-to-End Latency} (\textbf{Lat.}), and \textbf{Speedup} (\textbf{Spd.}). 
AccLen measures the average number of drafted tokens accepted per round and is the key efficiency metric, since longer accepted prefixes reduce verifier decoding rounds. 
Lat. includes routing overhead, and Spd. is measured relative to autoregressive verifier-only decoding. 
For non-speculative visual-token efficiency baselines, AccLen and VerRds are not defined, so we report task performance and latency-based efficiency instead.

\begin{table*}[t]
\centering
\footnotesize
\setlength{\tabcolsep}{4.5pt}
\resizebox{\textwidth}{!}{%
\begin{tabular}{llcccccccccc}
\toprule
\multirow{2}{*}{VLM setting} & \multirow{2}{*}{Method}
& \multicolumn{2}{c}{MMBench}
& \multicolumn{2}{c}{ScienceQA}
& \multicolumn{2}{c}{TextVQA}
& \multicolumn{2}{c}{MM-Vet}
& \multicolumn{2}{c}{COCO Caption} \\
\cmidrule(lr){3-4} \cmidrule(lr){5-6} \cmidrule(lr){7-8} \cmidrule(lr){9-10} \cmidrule(lr){11-12}
&
& AccLen $\uparrow$ & Spd. $\uparrow$
& AccLen $\uparrow$ & Spd. $\uparrow$
& AccLen $\uparrow$ & Spd. $\uparrow$
& AccLen $\uparrow$ & Spd. $\uparrow$
& AccLen $\uparrow$ & Spd. $\uparrow$ \\
\midrule

\multirow{8}{*}{LLaVA-v1.6-7B}
& Medusa          & 0.73 & 1.02x & 0.69 & 1.39x & 0.77 & 1.47x & 0.70 & 1.43x & 0.76 & 1.63x \\
& EAGLE-2         & 2.04 & 2.11x & 2.43 & 2.15x & 0.62 & 1.26x & 0.64 & 1.45x & 1.34 & 1.87x \\
& ViSpec          & 2.55 & 2.45x & 2.87 & 2.35x & 2.96 & 2.89x & 2.82 & 2.55x & 3.34 & 3.20x \\
& SpecFLASH       & 2.34 & 2.21x & 2.76 & 2.22x & 2.27 & 2.34x & 2.59 & 2.38x & 3.12 & 3.02x \\
& EAGLE-3         & 2.43 & 2.34x & 2.71 & 2.24x & 2.97 & 2.78x & 2.85 & \textbf{2.65x} & 3.30 & 3.18x \\
& RACER           & 2.44 & 2.35x & 2.45 & 2.30x & 2.90 & 2.80x & 2.87 & 2.49x & 3.29 & 3.20x \\
& TokenRecycling  & 2.54 & 2.43x & 2.67 & 2.31x & 2.87 & 2.68x & 2.77 & 2.54x & 3.31 & 3.21x \\
& TIGER           & \textbf{2.67} & \textbf{2.57x} & \textbf{2.88} & \textbf{2.37x} & \textbf{2.98} & \textbf{2.91x} & \textbf{2.88} & \textbf{2.65x} & \textbf{3.37} & \textbf{3.24x} \\
\midrule

\multirow{8}{*}{LLaVA-v1.6-13B}
& Medusa          & 1.23 & 1.18x & 0.86 & 1.62x & 0.87 & 1.52x & 0.81 & 1.45x & 0.79 & 1.23x \\
& EAGLE-2         & 1.98 & 1.93x & 1.98 & 2.10x & 1.67 & 1.78x & 1.65 & 1.58x & 1.87 & 1.67x \\
& ViSpec          & 2.56 & 2.34x & \textbf{2.74} & 2.55x & 2.75 & 2.50x & 2.76 & 2.33x & 3.19 & 2.16x \\
& SpecFLASH       & 2.49 & 2.14x & 2.54 & 2.32x & 2.46 & 2.23x & 2.62 & 2.27x & 2.97 & 2.01x \\
& EAGLE-3         & 2.55 & 2.33x & 2.70 & 2.52x & 2.79 & 2.65x & 2.74 & 2.39x & 3.24 & 2.15x \\
& RACER           & 2.52 & 2.26x & 2.68 & 2.49x & 2.77 & 2.60x & 2.67 & 2.33x & 3.20 & 2.09x \\
& TokenRecycling  & 2.54 & 2.19x & 2.65 & 2.53x & 2.80 & 2.45x & \textbf{2.89} & 2.40x & 3.18 & 2.00x \\
& TIGER           & \textbf{2.59} & \textbf{2.46x} & 2.71 & \textbf{2.56x} & \textbf{2.86} & \textbf{2.67x} & 2.75 & \textbf{2.45x} & \textbf{3.25} & \textbf{2.19x} \\
\midrule

\multirow{8}{*}{Qwen3-VL-4B}
& Medusa          & 1.34 & 1.09x & 0.65 & 1.12x & 0.69 & 1.09x & 0.67 & 1.44x & 0.66 & 1.17x \\
& EAGLE-2         & 2.12 & 1.76x & 1.16 & 1.32x & 1.01 & 1.25x & 1.34 & 1.45x & 1.45 & 1.68x \\
& ViSpec          & 2.56 & 2.16x & 2.03 & 1.79x & 2.18 & 1.74x & 2.54 & 2.21x & 2.34 & 2.11x \\
& SpecFLASH       & 2.55 & 2.32x & 1.95 & 1.77x & 2.17 & 1.34x & 2.46 & 2.11x & 2.21 & 2.02x \\
& EAGLE-3         & 2.57 & 2.40x & 2.08 & 1.70x & 2.20 & 1.75x & 2.56 & 2.28x & 2.37 & 2.00x \\
& RACER           & 2.58 & 2.36x & 2.02 & 1.74x & 2.19 & 1.78x & 2.54 & 2.29x & 2.33 & 2.09x \\
& TokenRecycling  & 2.59 & 2.38x & 2.01 & 1.76x & 2.21 & 1.80x & 2.58 & 2.30x & 2.41 & 2.01x \\
& TIGER           & \textbf{2.63} & \textbf{2.41x} & \textbf{2.10} & \textbf{1.87x} & \textbf{2.29} & \textbf{1.85x} & \textbf{2.67} & \textbf{2.33x} & \textbf{2.47} & \textbf{2.15x} \\
\midrule

\multirow{8}{*}{Qwen3-VL-8B}
& Medusa          & 1.23 & 1.16x & 0.67 & 1.05x & 0.55 & 0.97x & 0.63 & 1.03x & 0.67 & 1.02x \\
& EAGLE-2         & 1.14 & 1.67x & 1.54 & 1.23x & 1.17 & 1.02x & 1.34 & 1.23x & 1.47 & 1.34x \\
& ViSpec          & 2.13 & 1.99x & 2.34 & 1.76x & 2.02 & 1.12x & 2.35 & 1.65x & 2.36 & 1.78x \\
& SpecFLASH       & 2.06 & 2.02x & 2.27 & 1.65x & 1.97 & 1.04x & 2.28 & 1.43x & 2.21 & 1.45x \\
& EAGLE-3         & 2.11 & 2.07x & 2.28 & \textbf{1.77x} & 2.03 & 1.12x & 2.36 & 1.67x & 2.44 & \textbf{1.79x} \\
& RACER           & 2.15 & 2.03x & 2.25 & 1.70x & 1.99 & 1.13x & 2.37 & 1.56x & 2.45 & 1.67x \\
& TokenRecycling  & 2.13 & 2.00x & 2.30 & 1.69x & 2.00 & 1.15x & 2.40 & 1.58x & 2.32 & 1.66x \\
& TIGER           & \textbf{2.23} & \textbf{2.12x} & \textbf{2.37} & \textbf{1.77x} & \textbf{2.11} & \textbf{1.18x} & \textbf{2.45} & \textbf{1.70x} & \textbf{2.54} & \textbf{1.79x} \\
\bottomrule
\end{tabular}%
}
\caption{
Main exact verifier-side speculative decoding results across multimodal benchmarks.
We report accepted prefix length (AccLen) and speedup (Spd.).
Each model block denotes the VLM setting used in the adapted speculative decoding protocol.
The verifier is kept unchanged and uses the full visual context.
}
\label{tab:main_results}
\end{table*}

\subsection{Main Results}




We separate exact speculative decoding from auxiliary visual-interface analyses. 
Table~\ref{tab:main_results} keeps the verifier unchanged with full visual context, so AccLen and speedup are the main metrics. 
Tables~\ref{tab:visual_pruning} and~\ref{tab:routing_sparsity} instead analyze quality--latency trade-offs under different drafter-side visual interfaces.

\begin{table}[t]
\centering
\small
\setlength{\tabcolsep}{5pt}
\resizebox{\columnwidth}{!}{%
\begin{tabular}{llcc}
\toprule
Model & Visual interface & Caption Acc. $\uparrow$ & Norm. Lat. $\downarrow$ \\
\midrule
\multirow{4}{*}{LLaVA-v1.6-7B}
& Full visual tokens  & 0.45 & 1.00 \\
& SparseVLM           & 0.56 & 0.82 \\
& VisPruner           & 0.69 & 0.76 \\
& TIGER routing       & \textbf{0.75} & \textbf{0.62} \\
\midrule
\multirow{4}{*}{Qwen3-VL-4B}
& Full visual tokens  & 0.67 & 1.00 \\
& SparseVLM           & 0.68 & 0.86 \\
& VisPruner           & 0.75 & 0.79 \\
& TIGER routing       & \textbf{0.78} & \textbf{0.65} \\
\bottomrule
\end{tabular}%
}
\caption{
Auxiliary visual-token efficiency analysis on COCO Caption.
We compare different drafter-side visual interfaces using caption accuracy and normalized decoding latency.
}
\label{tab:visual_pruning}
\end{table}

Table~\ref{tab:main_results} reports exact verifier-side speculative decoding results across five multimodal benchmarks and four VLM settings. TIGER achieves the tied-best accepted prefix length and speedup in most settings, suggesting that text-conditioned visual routing and acceptance-aligned training improve drafter-verifier agreement across both LLaVA and Qwen-VL model families. The gains are consistent but generally modest over strong speculative decoding baselines such as ViSpec, EAGLE-3, RACER, and TokenRecycling, which is expected because these methods already provide competitive draft quality. Nevertheless, even small improvements in accepted prefix length are meaningful, since they reduce the number of expensive verifier decoding rounds and translate into end-to-end speedup. Additional context-heavy evaluations are reported in Appendix~\ref{app:context-heavy}, where TIGER shows moderate gains on short/medium video and multi-image settings while long-video cases remain challenging.

\subsection{Drafter-Side Visual Interface Analysis}

Table~\ref{tab:visual_pruning} compares TIGER with visual token efficiency baselines on COCO Caption. 
Unlike speculative decoding methods, SparseVLM and VisPruner reduce inference cost by pruning visual tokens, so AccLen is not directly defined for them. 
We therefore report caption accuracy and normalized latency to analyze the quality-latency trade-off of different drafter-side visual interfaces.
TIGER achieves a favorable quality--latency trade-off compared with both full-token decoding and pruning-based baselines. 
This suggests that routing visual evidence according to the current textual context is more effective than static visual-token reduction alone in this setting. 
Additional adapter-style comparisons, where SparseVLM/VisPruner-style selectors are inserted into the same speculative decoding pipeline, are reported in Appendix~\ref{app:adapter-pruning}.

\begin{table}[t]
\centering
\small
\setlength{\tabcolsep}{4.5pt}
\resizebox{\columnwidth}{!}{%
\begin{tabular}{llcccc}
\toprule
Model & Variant & AccLen $\uparrow$ & Spd. $\uparrow$ & VerRds. $\downarrow$ & Lat. $\downarrow$ \\
\midrule

\multirow{4}{*}{LLaVA-v1.6-7B}
& TIGER                    & \textbf{2.67} & \textbf{2.57x} & \textbf{0.375} & \textbf{0.389} \\
& w/o Visual Gated Routing & 2.13          & 2.12x          & 0.469          & 0.472 \\
& w/o Acceptance Alignment   & 2.07          & 2.03x          & 0.483          & 0.493 \\
& w/o KD                   & 2.32          & 1.99x          & 0.431          & 0.503 \\
\midrule

\multirow{4}{*}{Qwen3-VL-4B}
& TIGER                    & \textbf{2.63} & \textbf{2.41x} & \textbf{0.380} & \textbf{0.415} \\
& w/o Visual Gated Routing & 1.98          & 1.87x          & 0.505          & 0.535 \\
& w/o Acceptance Alignment   & 1.67          & 1.56x          & 0.599          & 0.641 \\
& w/o KD                   & 1.54          & 1.03x          & 0.649          & 0.971 \\
\bottomrule
\end{tabular}%
}
\caption{
Ablation study on COCO Caption.
All three components, visual gated routing, accepted-prefix alignment, and KD warm-start, contribute to higher accepted prefix length and lower decoding cost.
}
\label{tab:ablation_coco}
\end{table}

\subsection{Ablation Study}



We conduct ablation experiments on COCO Caption to analyze the contribution of TIGER's main components: text-conditioned visual gated routing, acceptance-aligned training, and KD warm-start initialization. 
Table~\ref{tab:ablation_coco} reports results on LLaVA-v1.6-7B and Qwen3-VL-4B. On LLaVA-v1.6-7B, removing visual gated routing reduces AccLen and speedup, while removing accepted alignment gives a similar drop to 2.07 AccLen and 2.03x speedup. 
On Qwen3-VL-4B, the degradation is more pronounced: removing accepted alignment reduces AccLen, and removing KD warm-start gives the lowest speedup, only 1.03x, with much higher normalized latency. 
These results indicate that visual routing improves drafter--verifier agreement, accepted-prefix optimization provides gains beyond KD-only imitation, and KD warm-start is important for stable policy optimization. Additional analyses are provided in the appendix: Appendix~\ref{app:routing-ablation} gives finer-grained routing ablations, and Appendix~\ref{app:objective-comparison} compares the accepted-prefix reward against KD-only, binary acceptance reward, and token-agreement reward.








\begin{table}[t]
\centering
\small
\setlength{\tabcolsep}{6pt}
\resizebox{\columnwidth}{!}{%
\begin{tabular}{cccc}
\toprule
Visual interface & AccLen $\uparrow$ & Spd. $\uparrow$ & Caption Acc. $\uparrow$ \\
\midrule
\(k=4\)  & 3.12 & 2.87x & 0.68 \\
\textbf{\(k=8\)} & \textbf{3.37} & \textbf{3.24x} & \textbf{0.75} \\
\(k=16\) & 2.76 & 2.01x & 0.56 \\
\(k=32\) & 2.54 & 1.78x & 0.53 \\
Full visual tokens & 2.23 & 1.23x & 0.45 \\
\bottomrule
\end{tabular}%
}
\caption{
Routing-sparsity analysis on COCO Caption using LLaVA-v1.6-Vicuna-7B.
This table studies the quality--efficiency trade-off of the drafter-side visual interface.
}
\label{tab:routing_sparsity}
\end{table}

\subsection{Routing Sparsity Analysis}

We analyze the effect of routing sparsity \(k\), the number of visual tokens routed to the drafter. 
Unless otherwise stated, TIGER uses \(k=8\) for both training and inference in all main experiments. 
Table~\ref{tab:routing_sparsity} reports a compact sensitivity analysis on COCO Caption with LLaVA-v1.6-Vicuna-7B. The results show that moderate sparsity gives the best quality--efficiency trade-off. 
When \(k=4\), the drafter may miss useful visual evidence, reducing caption accuracy and acceptance. 
When \(k\ge16\), the drafter receives more redundant visual context, which lowers accepted prefix length and speedup. 
The default setting \(k=8\) achieves the best trade-off among AccLen, speedup, and caption accuracy. 
Extended sensitivity results across OCR, VQA, captioning, and counting, including matched and mismatched \(k_{\text{train}}\) / \(k_{\text{infer}}\) settings, are reported in Appendix~\ref{app:k-sensitivity}.

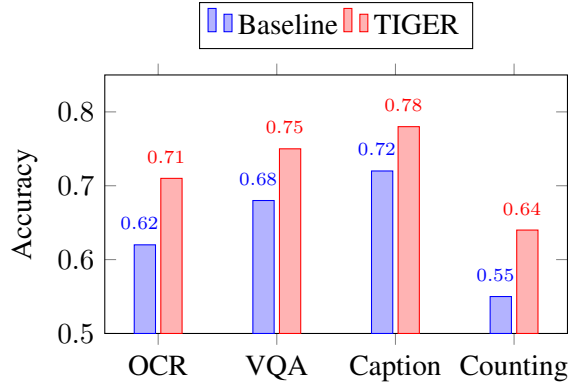
\begin{figure}[t]
\centering
\begin{tikzpicture}
\begin{axis}[
    ybar,
    bar width=8pt,
    width=\columnwidth,
    height=5cm,
    legend style={at={(0.5,1.1)},anchor=south,legend columns=-1},
    symbolic x coords={OCR,VQA,Caption,Counting},
    xtick=data,
    ymin=0.5,
    ymax=0.85,
    ylabel={Accuracy},
    enlarge x limits=0.15,
    nodes near coords,
    every node near coord/.append style={font=\scriptsize, yshift=2pt, xshift=-2pt},
    point meta=y
]

\addplot coordinates {(OCR,0.62) (VQA,0.68) (Caption,0.72) (Counting,0.55)};
\addplot coordinates {(OCR,0.71) (VQA,0.75) (Caption,0.78) (Counting,0.64)};

\legend{Baseline,TIGER}

\end{axis}
\end{tikzpicture}
\caption{
Capability analysis on LLaVA-v1.6-7B across different multimodal tasks.} 
\label{fig:capability_analysis}
\end{figure}

\subsection{Capability Analysis.}
\label{sec:capability-analysis}
We further analyze TIGER across different multimodal capability categories, including OCR understanding, visual question answering, caption generation, and counting.  This analysis is intended as a capability-preservation sanity check: the main speculative decoding metrics remain AccLen, verifier rounds, latency, and speedup, while task accuracy is used here to verify that efficiency gains do not come from degrading downstream capability. Details of the evaluation datasets and sampling protocol are provided in Appendix~\ref{app:dataset-details}. Figure~\ref{fig:capability_analysis} shows that TIGER improves task-level accuracy across the evaluated capability categories while improving decoding efficiency. 
The gains are more visible in OCR and counting, where accurate generation depends on selecting precise visual evidence. 
This supports the role of text-conditioned visual routing in helping the drafter focus on evidence that is relevant to the current decoding context.

\section{Conclusion}

We study speculative decoding for vision-language models and show that draft quality alone is not sufficient; decoding efficiency depends directly on verifier acceptance. 
We propose \textbf{TIGER}, which combines text-conditioned visual gated routing with acceptance-aligned training. 
TIGER routes context-relevant visual tokens to the drafter and optimizes continuations that yield longer verifier-accepted prefixes. 
Experiments show improved accepted-prefix efficiency and decoding speed while maintaining competitive task performance, highlighting verifier-acceptance alignment as an effective direction for efficient multimodal generation.

\section*{Limitations}

Despite the promising results, our work has several limitations. First, TIGER focuses primarily on improving speculative decoding efficiency rather than fundamentally modifying the architecture of vision-language models. 
As a result, the overall performance of the system still depends on the capabilities of the underlying drafter and verifier models. 
If the base models exhibit strong hallucination or reasoning errors, TIGER may not fully resolve these issues. Second, the proposed acceptance-aligned training requires sampling multiple speculative blocks and evaluating them with the verifier during training. 
Although we employ block-level verification to reduce computational overhead, this training procedure can still be more expensive than standard distillation-based approaches. Third, our routing mechanism selects a sparse subset of visual tokens based on the current textual prefix. 
While this design improves efficiency in practice, it may occasionally omit useful visual evidence, especially in complex scenes requiring global visual reasoning. Finally, our experiments focus on widely used multimodal benchmarks and model families such as LLaVA and Qwen-VL. 
Future work could explore the generality of TIGER across a broader range of architectures, larger-scale models, and more diverse multimodal tasks.

\bibliography{custom}

\appendix

\section{Method Details}

\subsection{Cost and Complexity Analysis}
We provide a simple cost decomposition to clarify why TIGER can improve speculative decoding efficiency despite introducing an additional routing step.

\paragraph{Expected number of verification rounds.}
For completeness, we briefly restate the relationship between accepted prefix length and verifier cost introduced in Section~\ref{sec:efficiency_rationale}. 
Let \(A_i\) denote the verifier-accepted prefix length at speculative round \(i\), and let
\begin{equation}
\bar A = \mathbb{E}[A_i].
\label{eq:app-avg-accepted}
\end{equation}
Assuming that the accepted lengths are sufficiently stable across rounds, the total number of verifier rounds required to generate a response of \(N\) tokens can be approximated by the total token budget divided by the average accepted progress per round:
\begin{equation}
\mathbb{E}[M] \approx \frac{N}{\bar A}.
\label{eq:app-rounds}
\end{equation}
This makes explicit why accepted prefix length is the central efficiency quantity in speculative decoding: larger \(\bar A\) implies fewer verifier invocations, and thus lower overall decoding cost.

\paragraph{Per-round speculative decoding cost.}
Let \(C_{\mathrm{draft}}\) and \(C_{\mathrm{verify}}\) denote the average drafting and verification costs per speculative round, respectively. The expected end-to-end speculative decoding cost can then be approximated as
\begin{equation}
\mathbb{E}[T_{\mathrm{SD}}]
\approx
\frac{N}{\bar A}
\bigl(C_{\mathrm{draft}} + C_{\mathrm{verify}}\bigr).
\label{eq:app-sd-cost}
\end{equation}
This expression makes explicit the central role of \(\bar A\): even when \(C_{\mathrm{draft}}\) remains fixed, larger verifier-accepted prefixes reduce the total number of rounds and therefore reduce overall decoding cost.

\paragraph{Full-visual versus routed drafting.}
In a conventional multimodal drafter, each speculative round conditions on the full visual token bank of size \(r\). Abstracting away model-specific constants, we write its drafting cost as
\begin{equation}
C_{\mathrm{draft}}^{\mathrm{full}} = C_{\mathrm{text}} + C_{\mathrm{vis}}(r),
\label{eq:app-full-draft}
\end{equation}
where \(C_{\mathrm{text}}\) denotes the text-side decoding cost and \(C_{\mathrm{vis}}(r)\) denotes the cost of interacting with the full visual context.

In TIGER, the drafter-side cost is factorized into two parts: a routing stage over the visual bank and a reduced drafting stage over only the selected subset of size \(k\), where \(k \ll r\):
\begin{equation}
C_{\mathrm{draft}}^{\mathrm{TIGER}}
=
C_{\mathrm{text}}
+
C_{\mathrm{route}}(r)
+
C_{\mathrm{vis}}(k).
\label{eq:app-tiger-draft}
\end{equation}
Here \(C_{\mathrm{route}}(r)\) is the cost of scoring the visual bank and selecting the top-\(k\) tokens, while \(C_{\mathrm{vis}}(k)\) is the visual interaction cost after routing.

Substituting Eq.~\eqref{eq:app-tiger-draft} into Eq.~\eqref{eq:app-sd-cost} gives the expected cost of TIGER:
\begin{equation}
\begin{aligned}
\mathbb{E}[T_{\mathrm{TIGER}}]
&\approx
\frac{N}{\bar A_{\mathrm{TIGER}}}
\Bigl(
C_{\mathrm{text}}
+
C_{\mathrm{route}}(r) \\
&\qquad\qquad
+
C_{\mathrm{vis}}(k)
+
C_{\mathrm{verify}}
\Bigr).
\end{aligned}
\label{eq:app-tiger-cost}
\end{equation}
Compared with a full-visual drafter, TIGER is beneficial whenever the improvement in \(\bar A\) outweighs the additional routing overhead.

\paragraph{Routing complexity.}
For a visual bank of size \(r\) with hidden dimension \(d\), the routing score in Eq.~\eqref{eq:routing-score} is computed once per visual token. The score computation therefore scales as
\begin{equation}
O(rd).
\label{eq:app-routing-score-complexity}
\end{equation}
Selecting the top-\(k\) tokens from the resulting scores can be implemented in \(O(r \log k)\) time using a standard partial sort, or \(O(r)\) using specialized selection routines. Hence, the routing overhead is linear (or near-linear) in the visual bank size and is incurred only once per speculative block.

By contrast, the subsequent visual conditioning cost of the drafter depends only on the selected subset size \(k\), rather than the full bank size \(r\). This is the main computational advantage of TIGER: routing introduces a small per-block overhead, but it reduces the amount of visual computation required during drafting.

\paragraph{Discussion.}
Eq.~\eqref{eq:app-tiger-cost} highlights the central trade-off in TIGER. On one hand, routing adds an extra cost term \(C_{\mathrm{route}}(r)\). On the other hand, it can both reduce the drafter-side visual interaction cost from \(C_{\mathrm{vis}}(r)\) to \(C_{\mathrm{vis}}(k)\) and, more importantly, increase \(\bar A\) by exposing more relevant visual evidence to the drafter. In practice, TIGER improves efficiency when the acceptance gain and reduced visual load together dominate the routing overhead. This is precisely the regime targeted by our design: sparse, text-conditioned routing for the drafter, combined with direct optimization of verifier-accepted prefix length.

\section{Implementation Details}

\subsection{Model Configurations}
\label{app:model-configs}

We evaluate TIGER on four open-source vision--language backbones: 
LLaVA-v1.6-Vicuna-7B, LLaVA-v1.6-Vicuna-13B, Qwen3-VL-4B-Instruct, and Qwen3-VL-8B-Instruct. 
Detailed technical specifications are reported in Table~\ref{tab:model_configs}.

In speculative decoding, the target model \(p_\phi\) also serves as the verifier, while the drafter \(q_\theta\) proposes candidate tokens before verification. 
Depending on the method, draft candidates may be produced by a smaller VLM, a lightweight method-specific draft module, or a training-free candidate-construction mechanism. 
For the two-model TIGER setting, we use LLaVA-v1.6-Vicuna-7B as the drafter with LLaVA-v1.6-Vicuna-13B as the verifier, and Qwen3-VL-4B-Instruct as the drafter with Qwen3-VL-8B-Instruct as the verifier.

Unless otherwise stated, we use the official pretrained models for all experiments. 
The verifier architecture is kept unchanged throughout training and evaluation and always uses the full visual context. 
TIGER is applied only to the drafter side, where we introduce text-conditioned visual gated routing and acceptance-aligned training. 
The original pretrained vision encoder is preserved, and routing is performed over the drafter-side visual token bank produced by the corresponding drafter backbone.

\begin{table*}[t]
\centering
\small
\setlength{\tabcolsep}{4pt}
\renewcommand{\arraystretch}{1.08}
\resizebox{0.98\textwidth}{!}{%
\begin{tabular}{lp{5.2cm}p{6.2cm}p{2.8cm}}
\toprule
Model & Text / LLM configuration & Vision configuration & Context \\
\midrule
LLaVA-v1.6-Vicuna-7B 
& hidden size 4096, 32 layers, 32 attention heads, vocabulary size 32{,}000, BF16
& CLIP ViT-L/14-336 vision tower, \texttt{anyres} + \texttt{spatial\_unpad}, projector \texttt{mlp2x\_gelu}
& max position 4096 \\

LLaVA-v1.6-Vicuna-13B 
& hidden size 5120, 40 layers, 40 attention heads, vocabulary size 32{,}000, BF16
& CLIP ViT-L/14-336 vision tower, \texttt{anyres} + \texttt{spatial\_unpad}, projector \texttt{mlp2x\_gelu}
& max position 4096 \\

Qwen3-VL-4B-Instruct
& hidden size 2560, 36 layers, 32 attention heads (8 KV heads), vocabulary size 151{,}936, BF16
& ViT depth 24, hidden size 1024, patch size 16, temporal patch size 2, DeepStack
& max position 262{,}144 \\

Qwen3-VL-8B-Instruct
& hidden size 4096, 36 layers, 32 attention heads (8 KV heads), vocabulary size 151{,}936, BF16
& ViT depth 27, hidden size 1152, patch size 16, temporal patch size 2, DeepStack
& max position 262{,}144 \\
\bottomrule
\end{tabular}%
}
\caption{
Technical specifications of the backbone models used in our experiments.
For two-model TIGER experiments, the smaller backbone in each family is used as the drafter and the larger backbone is used as the verifier.
}
\label{tab:model_configs}
\end{table*}

\subsection{Training Hyperparameters}

We summarize the training configuration of TIGER in Table~\ref{tab:tiger-training-config}. 
TIGER is trained in two stages: a KD warm-start stage that provides a stable reference policy, followed by an acceptance-aligned optimization stage using verifier-derived accepted-prefix rewards. 
The verifier is frozen throughout both stages.

\begin{table}[t]
\centering
\small
\setlength{\tabcolsep}{5pt}
\resizebox{\columnwidth}{!}{%
\begin{tabular}{lcc}
\toprule
Hyperparameter & KD warm-start & Acceptance alignment \\
\midrule
Training examples & 92.5K & 80K \\
Epochs & 1 & 1 \\
Learning rate & \(5 \times 10^{-6}\) & \(1 \times 10^{-5}\) \\
Per-device batch size & 1 & 1 \\
Gradient accumulation & 4 & 8 \\
Effective batch size & 4 & 8 \\
Max sequence length & 2048 & 2048 \\
Routing sparsity \(k\) & 8 & 8 \\
Draft length \(\gamma\) & -- & 32 \\
Candidate group size \(K\) & -- & 4 \\
KL coefficient \(\beta\) & -- & 0.02 \\
Verifier & frozen & frozen \\
Precision & BF16 & BF16 \\
\bottomrule
\end{tabular}%
}
\caption{
TIGER training configuration.
The drafter is first warm-started with KD and then optimized with verifier-derived accepted-prefix rewards.
The verifier is frozen in both stages.
}
\label{tab:tiger-training-config}
\end{table}

For the compared baselines, we distinguish methods with trainable acceleration modules from plug-and-play inference-time methods. 
Medusa, EAGLE-style methods, ViSpec, and SpecFLASH require training or retraining of method-specific drafting modules when adapted to our VLM backbones. 
RACER and TokenRecycling are training-free speculative decoding methods, while VisPruner and SparseVLM are inference-time visual-token reduction methods and do not require additional training under the standard acceleration pipeline. 
Detailed baseline adaptation procedures are provided in Appendix~\ref{app:baseline-adaptation}.
\noindent\textbf{ViSpec.}
ViSpec adopts a two-stage training pipeline~\citep{kang2025vispec}. 
In the first stage, the model is initialized through the standard vision-aware training procedure; in the second stage, it is further optimized with a speculative decoding head. 
According to the released implementation, the representative run configuration uses BF16 training, batch size \(=1\), and maximum sequence length \(=4096\) in both stages, with learning rate \(3\times 10^{-5}\) for Stage~1 and \(3\times 10^{-6}\) for the speculative stage. 
The speculative configuration further introduces \texttt{num\_q}=2. 
In addition, the default parser settings specify \texttt{num\_epochs}=20, gradient accumulation \(=1\), and configurable data-loading workers.

\noindent\textbf{SpecFLASH.}
SpecFLASH requires additional training for its acceleration modules and provides multiple training branches in the released codebase, including multimodal speculative decoding, vision token compression, and flash-style speculative acceleration~\citep{wang2025flash}. 
Across these branches, the typical setup uses batch size \(=2\) and gradient accumulation \(=1\). 
The default number of epochs is \(20\) for the main speculative branches, including \texttt{train\_llava}, \texttt{train\_qwenvl}, and \texttt{train\_sar}, while the compression branch \texttt{train\_compress} uses \(3\) epochs. 
The learning rate is typically set to \(3\times 10^{-5}\) for \texttt{train\_llava} and \texttt{train\_sar}, and \(2\times 10^{-5}\) for \texttt{train\_qwenvl} and \texttt{train\_compress}. 
The maximum sequence length is usually \(2048\), except for \texttt{train\_sar}, which uses \(4096\). 
Mixed precision is BF16 for the main speculative branches and FP16 for the compression branch.

\noindent\textbf{VisPruner and SparseVLM.}
VisPruner\citep{zhang2025vispruner} and SparseVLM~\citep{zhang2024sparsevlm} are plug-and-play inference-time acceleration methods. 
Their standard usage focuses on evaluation-time control of visual token reduction rather than additional training of the acceleration module itself. 
For VisPruner, inference behavior is governed by token-retention settings such as the retained token ratio and important-token selection. 
For SparseVLM, sparsity is controlled by the retained-token budget through the \texttt{RETAIN\_TOKN} setting. 
Although both repositories still include pretraining or finetuning scripts inherited from the original LLaVA codebase, these are used for reproducing or rebuilding the backbone rather than for the default acceleration pipeline.

\subsection{Decoding Configuration}

For a fair comparison, we follow the official or default decoding configurations provided in each method's released evaluation scripts whenever possible, rather than enforcing a unified decoding setup across all baselines. 
This choice is important because several compared methods are sensitive to method-specific decoding hyperparameters, especially in the speculative setting.

For visual-token reduction baselines such as \textbf{VisPruner} and \textbf{SparseVLM}, evaluation is conducted under deterministic decoding with greedy generation, i.e., single-beam decoding (\texttt{num\_beams}=1) and zero temperature in the benchmark scripts. 
For speculative decoding baselines such as \textbf{ViSpec} and \textbf{SpecFLASH}, we retain their default speculative tree configurations and sampling hyperparameters as specified in the corresponding official implementations.

More concretely, \textbf{VisPruner} uses \texttt{num\_beams}=1 and \texttt{max\_new\_tokens}=128, with benchmark evaluation scripts setting the temperature to \(0\) for greedy decoding, although the model loader exposes a default temperature of \(0.2\). 
\textbf{SparseVLM} follows the standard LLaVA-style greedy evaluation protocol with temperature \(0\), while the degree of visual sparsification is controlled by the retained-token budget. 
For \textbf{ViSpec}, the speculative decoding configuration uses the default tree-based setup with total-token \(=30\), depth \(=3\), top-k \(=8\), num-q \(=2\), \texttt{tree\_choices} set to \texttt{mc\_sim\_7b\_63}, \texttt{max\_new\_token}=1024, and sampling temperature \(=1.0\); its non-speculative baseline evaluation scripts also use temperature \(=1.0\). 
For \textbf{SpecFLASH}, the speculative evaluation scripts likewise use \texttt{max\_new\_token}=1024, temperature \(=1.0\), and \texttt{tree\_choices}=\texttt{mc\_sim\_7b\_63}, while the underlying speculative generator additionally exposes temperature, top-\(p\), and top-\(k\) controls.

Overall, our evaluation protocol is designed to respect the intended inference settings of each baseline while keeping the benchmark setup and backbone pairing fixed.

\subsection{Baseline Adaptation Details}
\label{app:baseline-adaptation}

We compare TIGER against representative baselines from multimodal speculative decoding, recent speculative decoding, and visual-token efficiency, including Medusa, EAGLE-2, EAGLE-3, RACER, TokenRecycling, ViSpec, SpecFLASH, SparseVLM, and VisPruner.
A key practical issue is that the officially released checkpoints of these methods do not consistently match the exact backbone pairs used in our experiments, namely LLaVA-v1.6-Vicuna-7B/13B and Qwen3-VL-4B/8B-Instruct.
Therefore, to ensure a fair comparison, we do not rely on mismatched off-the-shelf checkpoints.
Instead, for each baseline, we preserve the original method design as closely as possible while re-instantiating and, when necessary, retraining the method-specific components on top of the corresponding backbone used in our study.
For LLaVA-v1.6, we use the Vicuna/LLaMA-compatible tokenizer, chat template, and hidden-state interface; for Qwen3-VL, we use the Qwen-style tokenizer, chat template, and model configuration.
This protocol ensures that performance differences are attributable to the method itself rather than to differences in backbone family or released checkpoint quality.

\paragraph{Medusa.}
Medusa was originally introduced for language-only models and does not natively support multimodal inputs.
To adapt it to our setting, we attach Medusa-style multi-token prediction heads to the multimodal drafter and condition these heads on the same multimodal hidden states produced by the backbone VLM, including the original visual embeddings and textual prefix.
The verifier model is kept unchanged and verifies proposed tokens using the full visual context.
Since no official multimodal checkpoints are available for our exact drafter backbones, we train the Medusa heads directly on top of LLaVA-v1.6-Vicuna-7B and Qwen3-VL-4B-Instruct using the corresponding multimodal training data.
This adaptation preserves the original Medusa drafting principle while making it applicable to vision--language speculative decoding.

\paragraph{EAGLE-2.}
EAGLE-2 is adapted from the language-only setting by attaching an EAGLE-style drafting module to the multimodal hidden states of each VLM drafter.
For LLaVA-v1.6, we use the Vicuna/LLaMA-compatible EAGLE path; for Qwen3-VL, we use the Qwen-style tokenizer, chat template, and model configuration.
In both cases, the drafter receives the same image-conditioned prefix representation as the base VLM, while the larger verifier is kept unchanged and verifies proposed tokens with the full visual context.
Because the official checkpoints do not match our exact VLM backbone pairs, we retrain the EAGLE-2 drafting module on top of LLaVA-v1.6-Vicuna-7B and Qwen3-VL-4B-Instruct, with model-dependent hidden size, vocabulary size, attention heads, and RoPE settings matched to each backbone.
We follow the official EAGLE-2 training configuration, including learning rate \(3\times10^{-5}\), batch size 4, 20 epochs, maximum length 2048, uniform noise with standard deviation 0.2, bf16 mixed precision, and loss weights \(p_w=0.1\), \(v_w=1.0\), and \(head_w=0.1\).
At inference, we use the official tree-based decoding configuration with total draft tokens 60, depth 5, top-\(k=10\), and the corresponding EAGLE tree structure, while matching TIGER's verifier, prompt format, visual preprocessing, and decoding budget.

\paragraph{EAGLE-3.}
EAGLE-3 extends EAGLE-style speculative decoding with direct token prediction, multi-layer feature fusion, and a compact draft vocabulary.
We adapt EAGLE-3 to both LLaVA/Vicuna-style and Qwen-style VLMs by constructing its draft module on top of the multimodal hidden states from the corresponding drafter backbone.
For LLaVA-v1.6, we use the Vicuna/LLaMA-compatible tokenizer and configuration; for Qwen3-VL, we use the Qwen-specific tokenizer, chat template, and model configuration.
The verifier is kept unchanged and always uses the full visual context for verification.
Following the official EAGLE-3 setup, the draft module is a lightweight one-layer transformer-style head, with hidden size, attention configuration, vocabulary size, RoPE settings, and tokenizer-dependent IDs matched to each backbone rather than hard-coded from a single LLM family.
When supported by the backbone tokenizer, we retain the compact draft-vocabulary design with \(\texttt{draft\_vocab\_size}=32000\).
We train the EAGLE-3 draft module using the official training recipe: learning rate \(5\times10^{-5}\), batch size 1 per GPU with gradient accumulation 2, 40 epochs, maximum length 2048, gradient checkpointing, fp16 mixed precision, and DeepSpeed ZeRO-2.
At inference, we enable EAGLE-3 mode and use the same EAGLE-style tree-decoding parameters, including total draft tokens 60, depth 5, and top-\(k=10\).
All evaluation settings, including verifier model, prompt format, visual preprocessing, maximum generation length, and acceptance protocol, are matched with TIGER.

\paragraph{RACER.}
RACER is a training-free speculative decoding method that constructs draft candidates using retrieval-augmented future-token prediction.
It does not train a separate drafter; instead, it builds an automaton over recent token histories and combines retrieved exact patterns with logit-driven candidate expansion.
In our VLM adaptation, the multimodal input is processed by the target VLM with the original visual context, and RACER operates over the generated-token stream while preserving the same verifier and visual preprocessing as TIGER.
For Qwen3-VL, we use the native Qwen-style chat template and supported KV-model path; for LLaVA/Vicuna, we use the corresponding Vicuna-style prompt template and KV-compatible model wrapper.
We use the official default decoding configuration, including \(\texttt{max\_num\_draft}=64\), \(\texttt{max\_breadth}=8\), \(\texttt{ngram}=10\), \(\texttt{max\_nodes}=10000\), greedy decoding with temperature 0, and the same maximum generation budget as the other baselines.
Since RACER is training-free and supports batch-size-1 decoding in its native implementation, we evaluate it under the same latency-oriented setting and report results with the adapted multimodal protocol.

\paragraph{TokenRecycling.}
TokenRecycling is a plug-and-play speculative decoding method that accelerates inference without training an external draft model or using a datastore.
It reuses candidate tokens produced during decoding and organizes them into a fixed draft tree for verification by the target model.
To adapt it to VLMs, we keep the target VLM and visual encoder unchanged, process each multimodal input with the full visual context, and apply token recycling only to the language decoding stream.
We follow the official static-tree configuration, using \(\texttt{matrix\_top\_k}=8\) and the hardcoded draft tree of depth 6.
No additional multimodal drafter is trained for this baseline.
All candidate tokens are verified by the same target VLM under the matched prompt format, visual preprocessing, decoding budget, and latency measurement protocol.

\paragraph{ViSpec.}
ViSpec is a multimodal speculative decoding method whose drafter explicitly incorporates visual information through a lightweight vision-aware adaptor.
Its main design compresses visual tokens and injects a global image feature into the drafting process before verification by the target model.
We follow this design as closely as possible and use the official implementation when compatible.
However, because released weights are not uniformly available for all backbones considered in our experiments, we reproduce ViSpec on our evaluation backbones by retaining its vision-adaptor-based architecture and retraining the adaptor and speculative drafting components on the corresponding LLaVA-v1.6 and Qwen3-VL drafters.
The verifier remains unchanged and uses the full visual context.

\paragraph{SpecFLASH.}
SpecFLASH is reproduced using its early-exit speculative design, where an auxiliary draft head is attached to an intermediate layer of the base VLM to predict speculative tokens more cheaply before full verification.
We follow the original training pipeline as closely as possible, including the construction of hidden-state supervision for the early-exit head.
Because the released setup does not directly cover every model pair in our evaluation, we train the method-specific early-exit head on top of the exact LLaVA-v1.6 and Qwen3-VL drafters used in our experiments, while keeping the verifier unchanged.

\paragraph{SparseVLM.}
SparseVLM is not a speculative decoding method but a prompt-conditioned visual token sparsification baseline.
Its key idea is to retain only a subset of visual tokens based on text--vision interactions, thereby reducing visual computation during generation.
The official implementation is primarily centered on LLaVA-style backbones, so for LLaVA-v1.6 we follow the original sparsification design directly.
For Qwen3-VL, we adapt the same retention principle to the corresponding visual token interface of the backbone and tune the retained-token budget under matched evaluation settings.
Because SparseVLM does not include a verifier-based drafting stage, accepted-prefix metrics are not applicable.

\paragraph{VisPruner.}
VisPruner is another non-speculative efficiency baseline that accelerates VLM inference through visual token pruning.
Its pruning procedure first identifies important visual tokens using visual attention signals and then removes redundant tokens based on token similarity so that the retained subset remains both informative and diverse.
The original method is primarily developed for LLaVA-style backbones, and thus we reproduce its two-stage important-and-diverse pruning mechanism on the model families used in our experiments rather than comparing against unmatched released checkpoints.
As with SparseVLM, this baseline is evaluated as a visual-token reduction method rather than as a speculative decoder, so accepted-prefix-based metrics are not reported.

Across all baselines, we use the same evaluation benchmarks, prompt format, decoding budget, and verifier backbone corresponding to each model family whenever applicable.
For speculative decoding methods with a matched draft--verify protocol, we report accepted prefix length, verifier rounds, end-to-end speedup, latency, and task performance.
For training-free or plug-and-play speculative decoding methods, we use the adapted multimodal protocol above and report accepted-prefix metrics when their draft--verify interface is matched; otherwise, we additionally report task performance and latency-oriented metrics.
For visual-token efficiency methods without speculative verification, we report task accuracy together with latency-oriented efficiency metrics only.

\section{Dataset Details}
\label{app:dataset-details}
\subsection{Training Data}
\paragraph{KD Training Data.}
As the first stage of training, we warm-start the drafter using knowledge distillation on a mixed multimodal corpus constructed from six public vision--language datasets: LLaVA-1.5 COCO, TextVQA, ScienceQA, DocVQA, TallyQA, and ChartQA. 
The resulting mixture covers a broad range of multimodal capabilities, including general VQA, OCR, document understanding, chart reasoning, counting, and science-oriented visual QA.

The goal of this stage is to provide the drafter with a strong and diverse multimodal initialization before acceptance-aligned optimization. 
Rather than optimizing for a single benchmark, we expose the model to heterogeneous image--text interaction patterns so that it can learn to mimic the target model across different visual contexts and response formats. 
This diversity improves the drafter's generality and makes it a more reliable starting point for the later speculative decoding objective.

Each example in the KD pool consists of an image \(I\), a textual instruction or question \(X\), and a reference target response \(Y\). 
The drafter is trained to imitate the target model on this mixed corpus using standard distillation-based supervision. 
Unless otherwise stated, all datasets are merged into a single pool and uniformly sampled during KD training. 
The final KD mixture contains approximately \(92.5\)K multimodal examples; detailed dataset statistics are reported in Table~\ref{tab:kd_training_data}.

\begin{table*}[t]
\centering
\small
\begin{tabular}{lllc}
\toprule
Dataset & Split / Source & Main capability & \#Samples \\
\midrule
LLaVA-1.5 COCO & COCO-train2017 subset & General VQA / caption-style grounding & 40,000 \\
TextVQA        & train & OCR-centric VQA & 34,000 \\
ScienceQA      & train  & Science visual reasoning & 6,000 \\
DocVQA         & validation & Document understanding / OCR & 5,000 \\
TallyQA        & COCO-5k subset & Counting & 5,000 \\
ChartQA        & test & Chart understanding & 2,500 \\
\midrule
Total          & -- & Mixed multimodal KD corpus & 92,500 \\
\bottomrule
\end{tabular}%
\caption{
Composition of the multimodal training mixture used for KD warm-start of the drafter. 
The corpus combines six vision--language datasets covering complementary capability types, including general VQA, OCR, document understanding, counting, chart reasoning, and science-oriented multimodal QA.
}
\label{tab:kd_training_data}
\end{table*}

\paragraph{Acceptance-Aligned Training Data}
For the acceptance-aligned stage, we construct a dedicated multimodal training pool of approximately \(80\)K examples from six vision--language data sources: TextVQA, DocVQA, ChartQA, TallyQA, ScienceQA, and LLaVA-1.5 COCO. 
Compared with the KD warm-start corpus, this pool is designed specifically for reward-based optimization, where the goal is not merely to imitate the target model, but to expose the drafter to prompts that can induce meaningful variation in verifier acceptance outcomes.

To support stable preference- or reward-based training, we apply a lightweight heuristic filter to retain medium-difficulty prompts. 
In particular, we remove trivially short prompts (fewer than five words), excessively long prompts (more than sixty words), and short pure yes/no questions, since these cases often lead to degenerate or low-variance reward signals. 
The remaining prompts are assigned coarse capability tags, including \texttt{count}, \texttt{ocr}, \texttt{reasoning}, \texttt{description}, and \texttt{visual\_qa}, which provide a simple characterization of the training mixture. 
This filtering strategy is intended to encourage speculative samples with diverse acceptance lengths, which is important for effective GRPO-style optimization.

Each example in the GRPO pool consists of an image \(I\), a prompt \(X\), and an optional reference answer \(Y\) stored as \texttt{ground\_truth}. 
During training, the drafter generates multiple speculative continuations conditioned on the routed multimodal context, and the verifier evaluates these continuations to produce acceptance-based rewards. 
Thus, supervision is derived directly from verifier feedback rather than from manually curated preference annotations or external reward models.

We build the accepted-alignment pool by merging all retained examples into a single training set and sampling uniformly from the resulting mixture. 
Because the publicly available Hugging Face versions of ChartQA and DocVQA provide only limited test/validation splits, their contribution is capped by actual availability, and the LLaVA-COCO portion is correspondingly increased to preserve the target pool size. 
The final GRPO training pool contains approximately \(80\)K multimodal examples; detailed dataset statistics are reported in Table~\ref{tab:grpo_training_data}.

\begin{table*}[t]
\centering
\small
\begin{tabular}{lllc}
\toprule
Dataset & Split / Source & Main capability & \#Samples \\
\midrule
TextVQA      & train & OCR-centric VQA & 20,000 \\
DocVQA       & validation + test & Document understanding / OCR & 10,000 \\
ChartQA      & test & Chart understanding / reasoning & 2,500 \\
TallyQA      & COCO-5k subset & Counting & 5,000 \\
ScienceQA    & train & Science visual reasoning & 5,000 \\
LLaVA-1.5 COCO & COCO-train2017 subset & General visual QA / grounding & 37,500 \\
\midrule
Total        & -- & Accepted-alignment training pool & 80,000 \\
\bottomrule
\end{tabular}%
\caption{
Composition of the multimodal training pool used for acceptance-aligned optimization.
The pool is filtered to retain medium-difficulty prompts that are more likely to induce diverse verifier acceptance outcomes during GRPO training.
Because the public Hugging Face releases of ChartQA and DocVQA provide limited split availability, LLaVA-COCO is expanded to preserve the overall pool size.
}
\label{tab:grpo_training_data}
\end{table*}

\subsection{Capability Analysis Data}
\begin{table}[t]
\centering
\small
\setlength{\tabcolsep}{5pt}
\begin{tabular}{lll}
\toprule
Capability & Dataset & \# Test Samples \\
\midrule
OCR                & TextVQA      & 1,000 \\
VQA                & MMBench      & 1,000 \\
Captioning         & COCO Caption & 1,000 \\
Counting           & TallyQA        & 1,000 \\
\midrule
\multicolumn{2}{l}{Total} & 4,000 \\
\bottomrule
\end{tabular}
\caption{
Capability-oriented evaluation subsets used in our analysis.
Each capability is represented by 1,000 test examples from a corresponding benchmark.
}
\label{tab:capability_datasets}
\end{table}
To analyze which multimodal abilities benefit most from TIGER, we construct a capability-oriented evaluation suite.
As shown in Table~\ref{tab:capability_datasets}, we sample 1,000 test examples for each capability from a representative benchmark dataset.
This setup enables a controlled comparison of TIGER across OCR, VQA, captioning, and reasoning-oriented tasks.

\subsection{Evaluation Benchmarks}

We evaluate TIGER on five widely used multimodal benchmarks that cover complementary capability types, including general multimodal reasoning, science-oriented question answering, scene-text understanding, integrated visual reasoning, and open-ended caption generation.

\paragraph{MMBench.}
MMBench is a comprehensive multimodal benchmark designed to evaluate broad visual understanding and reasoning abilities under a unified multiple-choice format. 
It covers diverse capability categories such as object recognition, attribute understanding, spatial reasoning, commonsense inference, and cross-modal grounding. 
Because it includes a wide range of visual question types, MMBench provides a useful testbed for measuring whether a speculative decoding method can preserve general-purpose multimodal performance while improving inference efficiency.

\paragraph{ScienceQA.}
ScienceQA is a multimodal question answering benchmark built from science problems spanning multiple grade levels and subject areas. 
Many examples are accompanied by diagrams, figures, or natural images, requiring models to integrate visual evidence with textual reasoning and domain knowledge. 
This benchmark is particularly useful for assessing whether efficient decoding methods remain robust on reasoning-intensive multimodal tasks rather than only on surface-level perception.

\paragraph{TextVQA.}
TextVQA focuses on visual question answering in images that contain scene text. 
To answer correctly, a model must not only recognize objects and scenes, but also read and interpret textual content embedded in the image, such as signs, labels, numbers, and printed words. 
This makes TextVQA a strong benchmark for evaluating visual grounding accuracy in OCR-intensive settings, where speculative errors can easily arise if the drafter conditions on incomplete or irrelevant visual evidence.

\paragraph{MM-Vet.}
MM-Vet is a challenging open-ended benchmark designed to probe advanced multimodal capabilities that often require the combination of several skills at once, such as recognition, grounding, OCR, knowledge-based reasoning, and detailed comparison. 
Compared with standard single-skill benchmarks, MM-Vet places greater emphasis on compositional and fine-grained understanding. 
It therefore provides a more demanding evaluation setting for testing whether TIGER can maintain performance under complex multimodal reasoning requirements.

\paragraph{COCO Caption.}
COCO Caption is a standard image captioning benchmark based on natural images from the MS COCO dataset. 
The task requires the model to generate fluent and semantically grounded descriptions that reflect salient objects, attributes, interactions, and overall scene composition. 
We include this benchmark to evaluate TIGER in an open-ended generation setting, complementing question answering benchmarks and testing whether improved speculative efficiency transfers to caption generation quality.

Beyond the five main benchmarks above, we further evaluate TIGER on three context-heavy benchmarks to assess whether the method extends beyond standard single-image VQA and captioning settings. These additional evaluations cover video understanding, two-image reasoning, and broader multi-image reasoning.

\paragraph{Video-MME.}
Video-MME is a comprehensive video understanding benchmark designed to evaluate multimodal models on videos of different durations, including short, medium, and long videos. 
Compared with single-image tasks, Video-MME requires models to process temporally distributed visual evidence and maintain consistency across multiple sampled frames. 
We include this benchmark to test whether TIGER remains effective when the visual context is larger and more temporally extended. 
In our additional evaluation, short and medium videos are sampled at 1 fps, while long videos are sampled at 0.5 fps under a fixed maximum-frame budget.

\paragraph{NLVR2.}
NLVR2 is a two-image visual reasoning benchmark where each example contains a pair of images and a natural-language statement. 
The task is to determine whether the statement is true or false with respect to the paired visual input. 
This benchmark is useful for testing cross-image grounding and compositional reasoning, since the model must compare visual evidence across two images rather than answer from a single image alone. 
We use NLVR2 to evaluate whether TIGER's routed visual interface can support multi-image evidence selection in a controlled reasoning setting.

\paragraph{MuirBench.}
MuirBench is a robust multi-image understanding benchmark containing diverse multiple-choice questions across multiple multi-image reasoning tasks. 
Unlike NLVR2, which focuses on paired-image true/false reasoning, MuirBench covers broader multi-image relations and requires models to integrate information across several images. 
We include MuirBench to evaluate TIGER under a more general multi-image setting, where the relevant visual evidence may be distributed across multiple inputs and the drafter must preserve verifier agreement under a richer visual context.

\section{Evaluation Metrics Details}

We evaluate both task performance and decoding efficiency.

\paragraph{Task performance metrics.}
For benchmark performance, we follow the standard evaluation protocol of each dataset. 
For multiple-choice benchmarks such as MMBench and ScienceQA, we report \textbf{Accuracy} (\textbf{Acc.}). 
For captioning tasks such as COCO Caption, we report the standard captioning score used in the corresponding evaluation setting. 
For capability-oriented analysis, we use the task-specific evaluation metric associated with each benchmark subset.

\paragraph{Decoding efficiency metrics.}
For speculative decoding methods, we report four efficiency metrics. 
\textbf{Accepted Prefix Length} (\textbf{AccLen}) is the average number of drafted tokens accepted by the verifier in each speculative decoding round. 
Formally, if \(A_i\) denotes the accepted prefix length at speculative round \(i\), then
\begin{equation}
\text{AccLen} = \bar A = \mathbb{E}[A_i].
\label{eq:metric-acclen}
\end{equation}

\textbf{Verifier Rounds} (\textbf{VerRds}) measures the average number of verifier decoding rounds required to generate a complete response. 
Let \(N\) denote the total number of generated tokens. 
Following Eq.~\eqref{eq:rounds}, the expected number of verifier rounds is approximately
\begin{equation}
\text{VerRds} = \mathbb{E}[M] \approx \frac{N}{\bar A}.
\label{eq:metric-verrds}
\end{equation}

\textbf{End-to-End Latency} (\textbf{Lat.}) denotes the average wall-clock decoding time per sample. 
Because latency may be reported under different normalizations in different tables, we distinguish two variants when needed:
(i) \emph{AR-normalized latency}, where latency is normalized by standard autoregressive decoding with the verifier model; and
(ii) \emph{full-token-normalized latency}, where latency is normalized by the full-token decoding baseline within the same model family.
We explicitly state the normalization used in each table caption.

\textbf{Speedup} (\textbf{Spd.}) is defined as the ratio between baseline autoregressive decoding time and speculative decoding time:
\begin{equation}
\text{Spd.} = \frac{T_{\mathrm{AR}}}{T_{\mathrm{SD}}}.
\label{eq:metric-speedup}
\end{equation}
When \textbf{Lat.} is reported as AR-normalized latency, it satisfies
\begin{equation}
\text{Lat.} = \frac{T_{\mathrm{SD}}}{T_{\mathrm{AR}}} = \frac{1}{\text{Spd.}}.
\label{eq:metric-latency}
\end{equation}

Among these metrics, AccLen is the key quantity governing speculative decoding efficiency, since longer accepted prefixes reduce the number of expensive verifier decoding steps. 
Together, AccLen, VerRds, Lat., and Spd. characterize how improvements in verifier acceptance translate into end-to-end inference acceleration.

\paragraph{Applicability of metrics.}
AccLen and VerRds are defined only for speculative decoding methods, since they rely on the draft--verify process and the verifier acceptance test. 
For non-speculative visual token efficiency baselines such as SparseVLM and VisPruner, we report task performance together with latency-based efficiency metrics, but not AccLen or VerRds.

\section{Additional Experiments}

\subsection{Accepted Prefix Distribution Analysis}
\begin{table*}[t!]
\centering
\begin{tabular}{lcccc}
\toprule
Method & AccLen $\uparrow$ & $P(A \ge 2)$ (\%) $\uparrow$ & $P(A \ge 4)$ (\%) $\uparrow$ & $P(A \ge 8)$ (\%) $\uparrow$ \\
\midrule
Medusa    & 0.76 & 12.4 & 5.1  & 1.2 \\
EAGLE-2   & 1.34 & 35.2 & 15.8 & 4.5 \\
ViSpec    & 3.34 & 68.5 & 32.1 & 12.4 \\
SpecFLASH & 3.12 & 65.4 & 29.8 & 10.9 \\
\textbf{TIGER} & \textbf{3.37} & \textbf{70.2} & \textbf{35.6} & \textbf{15.3} \\
\bottomrule
\end{tabular}%
\caption{
Distribution of verifier-accepted prefix lengths on COCO Caption with LLaVA-v1.6-7B. 
Beyond improving the average accepted prefix length, TIGER consistently increases the probability of longer accepted prefixes, which are the primary driver of speculative decoding speedup.
}
\label{tab:prefix_distribution}
\end{table*}

We further analyze the distribution of verifier-accepted prefix lengths on the COCO Caption benchmark using LLaVA-v1.6-7B as the drafter. 
Although the average accepted prefix length (AccLen) is a useful summary metric, it does not fully reveal how often a method achieves long accepted continuations, which are the main source of practical speculative decoding gains. 
A method may obtain a similar mean accepted length while exhibiting very different behavior in the tail of the distribution, especially in the frequency of long accepted prefixes.

Table~\ref{tab:prefix_distribution} reports both the average accepted prefix length and the probability that the verifier accepts at least \(2\), \(4\), or \(8\) tokens in a speculative round. 
Compared with baseline speculative decoding methods, TIGER not only achieves the highest mean accepted prefix length, but also consistently shifts the distribution toward longer accepted prefixes. 
In particular, TIGER yields the highest probabilities at all reported thresholds, indicating that it produces verifier-compatible continuations more reliably across speculative rounds.

This distributional improvement is important because speculative decoding speedup is driven disproportionately by long accepted prefixes: when the verifier accepts a larger block, more tokens are generated per expensive verification step, thereby reducing the overall number of verifier rounds. 
The results therefore support our central claim that TIGER improves efficiency not only on average, but also by increasing the frequency of high-value speculative rounds that contribute most to end-to-end acceleration.

\subsection{Temperature Sensitivity}

\begin{table}[th!]
\centering
\small
\setlength{\tabcolsep}{5pt}
\begin{tabular}{lcccc}
\toprule
Temperature & AccLen $\uparrow$ & Spd. $\uparrow$ & VerRds. $\downarrow$ & Lat. $\downarrow$ \\
\midrule
0.0 & \textbf{3.37} & \textbf{3.24x} & \textbf{0.37}  & \textbf{0.389} \\
0.5 & 2.98          & 2.57x          & 0.39            & 0.399 \\
0.7 & 2.67          & 2.43x          & 0.41            & 0.412 \\
1.0 & 2.02          & 1.96x          & 0.494           & 0.432 \\
\bottomrule
\end{tabular}%

\caption{
Temperature sensitivity of TIGER on COCO Caption with LLaVA-v1.6-7B. 
Higher temperature weakens drafter--verifier agreement, resulting in shorter accepted prefixes, more verifier rounds, higher latency, and lower speculative decoding speedup.
}
\label{tab:temp_sensitivity}
\end{table}
We analyze the sensitivity of TIGER to decoding temperature on the COCO Caption benchmark using LLaVA-v1.6-7B. 
Speculative decoding relies on agreement between the drafter and the verifier, and this agreement typically becomes weaker under higher-temperature sampling. 
As the temperature increases, the drafter produces more diverse continuations, but these continuations are also more likely to deviate from the verifier, leading to shorter accepted prefixes and reduced speculative gains.

Table~\ref{tab:temp_sensitivity} reports the effect of temperature on accepted prefix length, speedup, verifier rounds, and latency. 
We observe a clear monotonic degradation in speculative decoding efficiency as temperature increases. 
Specifically, the average accepted prefix length drops from \(3.37\) at temperature \(0\) to \(2.02\) at temperature \(1.0\), while the corresponding speedup decreases from \(3.24\times\) to \(1.96\times\). 
At the same time, verifier rounds increase from \(0.37\) to \(0.494\), and latency rises from \(0.389\) to \(0.432\), indicating that weaker drafter--verifier alignment causes the verifier to intervene more frequently. 
Overall, these results confirm that TIGER is most effective in lower-temperature settings, where speculative continuations remain better aligned with verifier preferences.

\subsection{Additional Context-Heavy Evaluation}
\label{app:context-heavy}

We further evaluate TIGER on context-heavy multimodal settings beyond standard single-image VQA and captioning. 
This evaluation tests whether text-conditioned routing and acceptance-aligned training remain effective when relevant visual evidence is distributed across multiple frames or multiple images.

Table~\ref{tab:context_heavy_eval} reports results on Video-MME, NLVR2, and MuirBench under both LLaVA-v1.6-13B and Qwen3-VL-8B settings. 
For Video-MME, we evaluate the short-, medium-, and long-duration test splits separately, using 900 samples for each split. 
For NLVR2 and MuirBench, we evaluate 1,000 test samples from each benchmark. 
These settings cover video understanding, two-image reasoning, and broader multi-image reasoning.

Overall, this context-heavy evaluation provides a stronger test of TIGER beyond single-image tasks. 
The short/medium video and multi-image settings evaluate whether the drafter can benefit from compact, context-relevant visual evidence, while the long-video setting is more challenging because the relevant evidence may be temporally sparse and distributed across many frames.

\begin{table*}[t]
\centering
\footnotesize
\setlength{\tabcolsep}{4.2pt}
\resizebox{\textwidth}{!}{%
\begin{tabular}{llcccccccccc}
\toprule
\multirow{2}{*}{Model} & \multirow{2}{*}{Method}
& \multicolumn{2}{c}{Video-MME Short}
& \multicolumn{2}{c}{Video-MME Medium}
& \multicolumn{2}{c}{Video-MME Long}
& \multicolumn{2}{c}{NLVR2}
& \multicolumn{2}{c}{MuirBench} \\
\cmidrule(lr){3-4}
\cmidrule(lr){5-6}
\cmidrule(lr){7-8}
\cmidrule(lr){9-10}
\cmidrule(lr){11-12}
&
& AccLen $\uparrow$ & Spd. $\uparrow$
& AccLen $\uparrow$ & Spd. $\uparrow$
& AccLen $\uparrow$ & Spd. $\uparrow$
& AccLen $\uparrow$ & Spd. $\uparrow$
& AccLen $\uparrow$ & Spd. $\uparrow$ \\
\midrule

\multirow{8}{*}{LLaVA-v1.6-13B}
& Medusa          & 1.62 & 1.48x & 1.41 & 1.32x & 1.20 & 1.14x & 1.39 & 1.28x & 1.32 & 1.22x \\
& EAGLE-2         & 1.95 & 1.81x & 1.72 & 1.63x & 1.31 & 1.22x & 1.64 & 1.51x & 1.55 & 1.44x \\
& ViSpec          & 2.01 & 1.86x & 1.78 & 1.69x & 1.34 & 1.25x & 1.70 & 1.58x & 1.61 & 1.50x \\
& SpecFLASH       & 2.08 & 1.93x & 1.83 & 1.72x & 1.36 & 1.27x & 1.74 & 1.62x & 1.66 & 1.54x \\
& EAGLE-3         & 2.21 & 2.08x & 1.96 & 1.84x & 1.42 & 1.31x & 1.82 & 1.76x & 1.69 & 1.61x \\
& RACER           & 2.24 & 2.10x & 1.99 & 1.86x & 1.43 & 1.32x & 1.85 & 1.79x & 1.72 & 1.64x \\
& TokenRecycling  & 2.22 & 2.09x & 2.01 & 1.88x & \textbf{1.45} & 1.33x & 1.83 & 1.77x & 1.74 & 1.65x \\
& TIGER           & \textbf{2.35} & \textbf{2.19x}
                  & \textbf{2.10} & \textbf{1.96x}
                  & \textbf{1.45} & \textbf{1.34x}
                  & \textbf{1.94} & \textbf{1.86x}
                  & \textbf{1.80} & \textbf{1.72x} \\
\midrule

\multirow{8}{*}{Qwen3-VL-8B}
& Medusa          & 1.51 & 1.39x & 1.32 & 1.24x & 1.15 & 1.09x & 1.31 & 1.21x & 1.24 & 1.16x \\
& EAGLE-2         & 1.84 & 1.70x & 1.63 & 1.52x & 1.26 & 1.18x & 1.55 & 1.43x & 1.47 & 1.37x \\
& ViSpec          & 1.90 & 1.75x & 1.69 & 1.58x & 1.29 & 1.21x & 1.61 & 1.49x & 1.52 & 1.41x \\
& SpecFLASH       & 1.96 & 1.82x & 1.74 & 1.63x & 1.31 & 1.23x & 1.66 & 1.54x & 1.57 & 1.46x \\
& EAGLE-3         & 2.08 & 1.94x & 1.85 & 1.73x & 1.37 & 1.26x & 1.73 & 1.64x & 1.61 & 1.51x \\
& RACER           & 2.11 & 1.97x & 1.88 & 1.76x & \textbf{1.39} & 1.28x & 1.76 & 1.68x & 1.64 & 1.53x \\
& TokenRecycling  & 2.10 & 1.95x & 1.89 & 1.77x & 1.38 & 1.27x & 1.75 & 1.66x & 1.65 & 1.55x \\
& TIGER           & \textbf{2.22} & \textbf{2.06x}
                  & \textbf{1.98} & \textbf{1.85x}
                  & \textbf{1.39} & \textbf{1.29x}
                  & \textbf{1.84} & \textbf{1.76x}
                  & \textbf{1.71} & \textbf{1.62x} \\
\bottomrule
\end{tabular}%
}
\caption{
Additional context-heavy exact verifier-side speculative decoding results beyond single-image VQA and captioning.
We evaluate Video-MME using 900 test samples for each duration split: short, medium, and long.
We also evaluate 1,000 test samples each from NLVR2 and MuirBench.
For Video-MME, short and medium videos are sampled at 1 fps, while long videos are sampled at 0.5 fps under a fixed maximum-frame budget.
We report accepted prefix length (AccLen) and speedup (Spd.); best results in each model block and setting are boldfaced.
}
\label{tab:context_heavy_eval}
\end{table*}

\subsection{Batch-Size Scalability}
\label{app:batch-size}

To examine whether TIGER remains effective beyond batch-size-1 inference, we evaluate batch sizes 1, 8, and 16 on the capability-oriented evaluation suite. We report normalized per-sample latency, speedup over target-only decoding at the same batch size, peak GPU memory usage, and task accuracy. The results show that TIGER is most effective in latency-sensitive small-batch settings, while still providing positive speedups under moderate batching.

\begin{table*}[t]
\centering
\small
\setlength{\tabcolsep}{5pt}
\resizebox{\textwidth}{!}{%
\begin{tabular}{lccccccc}
\toprule
Capability & Batch size 
& Target Lat. $\downarrow$ 
& TIGER Lat. $\downarrow$ 
& Spd. $\uparrow$ 
& Target Mem. (GB) 
& TIGER Mem. (GB) 
& Acc. $\uparrow$ \\
\midrule
OCR & 1  & 1.000 & \textbf{0.412} & \textbf{2.43x} & 18.7 & 20.4 & 0.71 \\
OCR & 8  & 0.684 & \textbf{0.372} & \textbf{1.84x} & 31.6 & 34.9 & 0.70 \\
OCR & 16 & 0.603 & \textbf{0.385} & \textbf{1.57x} & 45.8 & 50.7 & 0.70 \\
\midrule
VQA & 1  & 1.000 & \textbf{0.438} & \textbf{2.28x} & 18.9 & 20.7 & 0.75 \\
VQA & 8  & 0.701 & \textbf{0.398} & \textbf{1.76x} & 32.1 & 35.4 & 0.74 \\
VQA & 16 & 0.622 & \textbf{0.411} & \textbf{1.51x} & 46.3 & 51.2 & 0.74 \\
\midrule
Caption & 1  & 1.000 & \textbf{0.389} & \textbf{2.57x} & 19.2 & 21.1 & 0.78 \\
Caption & 8  & 0.672 & \textbf{0.358} & \textbf{1.88x} & 32.8 & 36.0 & 0.77 \\
Caption & 16 & 0.596 & \textbf{0.374} & \textbf{1.59x} & 47.1 & 52.0 & 0.77 \\
\bottomrule
\end{tabular}%
}
\caption{
Batch-size scalability analysis on the capability-oriented evaluation suite.
Latency is normalized per sample, and speedup is measured against target-only decoding at the same batch size.
TIGER achieves the largest gains at batch size 1, but remains faster than target-only decoding at batch sizes 8 and 16 with moderate additional GPU memory usage.
}
\label{tab:batch-size-scalability}
\end{table*}

\subsection{Extended Routing-Sparsity Sensitivity}
\label{app:k-sensitivity}

We further analyze the effect of the routing sparsity \(k\), i.e., the number of visual tokens routed to the drafter. Unless otherwise stated, TIGER uses \(k=8\) for both training and inference in the main experiments. To verify this choice, we evaluate matched settings with \(k_{\text{train}}=k_{\text{infer}}\in\{4,8,16,32\}\), as well as mismatched settings where the model is trained with \(k_{\text{train}}=8\) but inferred with \(k_{\text{infer}}\in\{4,16\}\). We use the same capability-oriented evaluation suite as in Section~4.6, covering OCR, VQA, captioning, and counting.

\begin{table*}[t]
\centering
\scriptsize
\setlength{\tabcolsep}{3.5pt}
\resizebox{\textwidth}{!}{%
\begin{tabular}{lcc|ccc|ccc|ccc|ccc}
\toprule
\multirow{2}{*}{Model} 
& \multirow{2}{*}{$k_{\text{train}}$} 
& \multirow{2}{*}{$k_{\text{infer}}$}
& \multicolumn{3}{c|}{OCR}
& \multicolumn{3}{c|}{VQA}
& \multicolumn{3}{c|}{Caption}
& \multicolumn{3}{c}{Counting} \\
\cmidrule(lr){4-6}
\cmidrule(lr){7-9}
\cmidrule(lr){10-12}
\cmidrule(lr){13-15}
& & 
& AccLen $\uparrow$ & Spd. $\uparrow$ & Acc. $\uparrow$
& AccLen $\uparrow$ & Spd. $\uparrow$ & Acc. $\uparrow$
& AccLen $\uparrow$ & Spd. $\uparrow$ & Acc. $\uparrow$
& AccLen $\uparrow$ & Spd. $\uparrow$ & Acc. $\uparrow$ \\
\midrule

\multirow{6}{*}{LLaVA-v1.6-7B}
& 4  & 4  & 3.10 & 2.88x & 0.66 & 2.52 & 2.41x & 0.70 & 3.12 & 2.87x & 0.68 & 2.46 & 2.35x & 0.59 \\
& 8  & 8  & \textbf{3.36} & \textbf{3.21x} & \textbf{0.71} & \textbf{2.74} & \textbf{2.62x} & \textbf{0.75} & \textbf{3.37} & \textbf{3.24x} & \textbf{0.78} & \textbf{2.68} & \textbf{2.55x} & \textbf{0.64} \\
& 8  & 4  & 3.02 & 2.74x & 0.65 & 2.43 & 2.28x & 0.69 & 3.05 & 2.79x & 0.67 & 2.39 & 2.21x & 0.58 \\
& 8  & 16 & 2.84 & 2.36x & 0.64 & 2.31 & 2.08x & 0.68 & 2.76 & 2.01x & 0.56 & 2.25 & 2.02x & 0.57 \\
& 16 & 16 & 2.79 & 2.29x & 0.63 & 2.26 & 2.03x & 0.67 & 2.76 & 2.01x & 0.56 & 2.20 & 1.97x & 0.56 \\
& 32 & 32 & 2.58 & 1.92x & 0.60 & 2.10 & 1.81x & 0.64 & 2.54 & 1.78x & 0.53 & 2.04 & 1.76x & 0.53 \\
\midrule

\multirow{6}{*}{Qwen3-VL-4B}
& 4  & 4  & 2.73 & 2.42x & 0.63 & 2.26 & 2.12x & 0.67 & 2.86 & 2.51x & 0.66 & 2.18 & 2.06x & 0.56 \\
& 8  & 8  & \textbf{2.95} & \textbf{2.66x} & \textbf{0.68} & \textbf{2.47} & \textbf{2.32x} & \textbf{0.72} & \textbf{3.04} & \textbf{2.74x} & \textbf{0.73} & \textbf{2.39} & \textbf{2.25x} & \textbf{0.61} \\
& 8  & 4  & 2.65 & 2.32x & 0.62 & 2.18 & 2.01x & 0.66 & 2.76 & 2.39x & 0.65 & 2.11 & 1.96x & 0.55 \\
& 8  & 16 & 2.48 & 2.04x & 0.61 & 2.03 & 1.86x & 0.65 & 2.53 & 2.05x & 0.55 & 1.98 & 1.82x & 0.54 \\
& 16 & 16 & 2.43 & 1.98x & 0.60 & 1.99 & 1.81x & 0.64 & 2.47 & 1.99x & 0.54 & 1.94 & 1.78x & 0.53 \\
& 32 & 32 & 2.21 & 1.71x & 0.57 & 1.82 & 1.62x & 0.61 & 2.26 & 1.69x & 0.51 & 1.77 & 1.58x & 0.50 \\
\bottomrule
\end{tabular}%
}
\caption{
Extended routing-sparsity sensitivity across capability subsets.
TIGER uses \(k=8\) for both training and inference in the main experiments.
The matched setting \(k_{\text{train}}=k_{\text{infer}}=8\) gives the best overall trade-off across accepted prefix length, speedup, and task accuracy.
Mismatched settings, such as \(k_{\text{train}}=8,k_{\text{infer}}=4\) and \(k_{\text{train}}=8,k_{\text{infer}}=16\), underperform the matched default.
}
\label{tab:k-sensitivity}
\end{table*}

The results confirm the trend observed in the main routing-sparsity analysis. Smaller \(k\) values can omit useful visual evidence, while larger \(k\) values introduce more redundant visual context and reduce speculative acceptance. The matched default \(k_{\text{train}}=k_{\text{infer}}=8\) consistently provides the strongest trade-off, supporting our use of the same routing sparsity during training and inference.

\subsection{Comparison with SparseVLM/VisPruner-style Selectors}
\label{app:adapter-pruning}

SparseVLM and VisPruner are relevant visual-token sparsification methods, but they are not speculative decoding methods by themselves. Their primary goal is to reduce visual-token computation while preserving downstream task quality. In contrast, TIGER's visual routing is designed for multimodal speculative decoding: the selected visual tokens should help the drafter better match the verifier's accepted continuation, thereby increasing verifier-accepted prefix length.

To make the comparison more direct, we conduct an adapter-style experiment where SparseVLM-style and VisPruner-style selectors replace TIGER's visual router inside the same drafter--verifier speculative decoding pipeline. All methods use the same verifier, decoding budget, and evaluation subset. This isolates the effect of the drafter-side visual selector while keeping the speculative decoding protocol fixed.

\begin{table}[t]
\centering
\small
\setlength{\tabcolsep}{5pt}
\resizebox{\columnwidth}{!}{%
\begin{tabular}{lcccc}
\toprule
Drafter visual selector 
& AccLen $\uparrow$ 
& Spd. $\uparrow$ 
& Lat. $\downarrow$ 
& Acc. $\uparrow$ \\
\midrule
SparseVLM-style selector & 2.86 & 2.48x & 0.65 & 0.72 \\
VisPruner-style selector & 2.79 & 2.39x & 0.66 & 0.71 \\
TIGER routing-only       & 3.02 & 2.76x & 0.63 & 0.73 \\
TIGER full               & \textbf{3.37} & \textbf{3.24x} & \textbf{0.62} & \textbf{0.75} \\
\bottomrule
\end{tabular}%
}
\caption{
Adapter-style comparison with SparseVLM/VisPruner-style visual selectors on COCO Caption.
All selectors are inserted into the same drafter-side speculative decoding pipeline.
SparseVLM/VisPruner-style selectors improve visual-token efficiency, but TIGER routing better optimizes drafter--verifier agreement and achieves higher accepted prefix length.
}
\label{tab:adapter-pruning}
\end{table}

The results show that SparseVLM/VisPruner-style selectors provide useful visual-token reduction baselines, but they underperform TIGER in accepted prefix length and speedup. This supports our hypothesis that the visual tokens most useful for speculative decoding are not necessarily the globally most salient tokens for task accuracy. Instead, the selector should expose visual evidence that helps the drafter match the verifier under the current textual prefix. TIGER routing-only already improves over the adapted pruning selectors, and the full model further benefits from acceptance-aligned training.

\subsection{Objective Comparison for Accepted-Prefix Reward}
\label{app:objective-comparison}

We further compare the proposed accepted-prefix reward with simpler training objectives. The goal is to test whether optimizing verifier-accepted prefix length provides benefits beyond standard imitation-based distillation or simpler verifier-derived rewards.

We compare five variants. The \emph{vanilla drafter} is trained without KD warm-start or acceptance-aligned optimization. The \emph{KD-only} variant corresponds to the drafter after the distillation warm-start stage, without the second-stage accepted-prefix optimization. We also consider two simpler verifier-derived rewards: a \emph{binary acceptance reward}, which gives reward 1 if at least one drafted token is accepted and 0 otherwise, and a \emph{token-agreement reward}, which rewards token-level agreement with the verifier without explicitly requiring a contiguous accepted prefix. TIGER uses the proposed accepted-prefix reward, where the reward equals the number of consecutively accepted tokens from the beginning of the speculative block.

\begin{table*}[t]
\centering
\small
\setlength{\tabcolsep}{6pt}
\resizebox{\textwidth}{!}{%
\begin{tabular}{llcccc}
\toprule
Model & Method & AccLen $\uparrow$ & Spd. $\uparrow$ & Lat. $\downarrow$ & Acc. $\uparrow$ \\
\midrule
\multirow{5}{*}{LLaVA-v1.6-7B}
& Vanilla drafter & 1.06 & 1.04x & 0.534 & 0.44 \\
& KD-only (w/o Accepted Alignment) & 2.07 & 2.03x & 0.493 & 0.60 \\
& KD + binary acceptance reward & 2.17 & 2.19x & 0.476 & 0.61 \\
& KD + token-agreement reward & 2.28 & 2.33x & 0.462 & 0.65 \\
& TIGER & \textbf{2.67} & \textbf{2.57x} & \textbf{0.389} & \textbf{0.75} \\
\midrule
\multirow{5}{*}{Qwen3-VL-4B}
& Vanilla drafter & 1.12 & 1.02x & 0.768 & 0.46 \\
& KD-only (w/o Accepted Alignment) & 1.67 & 1.56x & 0.641 & 0.62 \\
& KD + binary acceptance reward & 2.23 & 1.87x & 0.601 & 0.65 \\
& KD + token-agreement reward & 2.34 & 2.01x & 0.557 & 0.69 \\
& TIGER & \textbf{2.63} & \textbf{2.41x} & \textbf{0.415} & \textbf{0.78} \\
\bottomrule
\end{tabular}%
}
\caption{
Objective comparison for accepted-prefix training.
KD-only improves over the vanilla drafter, confirming the value of the distillation warm-start.
However, simpler verifier-derived rewards still underperform TIGER.
The accepted-prefix reward directly optimizes the contiguous verifier-accepted prefix, which is the runtime quantity governing speculative decoding efficiency.
}
\label{tab:objective-comparison}
\end{table*}

The results show that KD-only provides a strong warm-start over the vanilla drafter, but it does not fully align the drafter with the verifier's acceptance behavior. Binary acceptance provides only a coarse block-level signal, while token-agreement reward captures local token matches without explicitly optimizing the contiguous accepted prefix that speculative decoding can commit. In contrast, TIGER directly rewards longer verifier-accepted prefixes, yielding the best accepted prefix length, speedup, latency, and task accuracy across both model families.

\subsection{Fine-Grained Visual Routing Ablations}
\label{app:routing-ablation}

We further ablate the internal design of TIGER's visual gated routing module.
The goal is to test whether the gains come merely from reducing the number of visual tokens, or from the proposed text-conditioned and visual-conditioned routing design.
We compare TIGER with several routing variants: random Top-\(k\) selection, static image-only Top-\(k\) selection, a text-only query variant, TIGER routing without acceptance-aligned training, and the full TIGER model.

\begin{table}[t]
\centering
\small
\setlength{\tabcolsep}{4pt}
\resizebox{\columnwidth}{!}{%
\begin{tabular}{llcccc}
\toprule
Variant & Query / selector & AccLen $\uparrow$ & Spd. $\uparrow$ & Lat. $\downarrow$ & Acc. $\uparrow$ \\
\midrule
Full visual context & all visual tokens & 2.24 & 1.38x & 0.74 & 0.72 \\
Random Top-\(k\) & random visual tokens & 2.41 & 1.77x & 0.69 & 0.66 \\
Static image-only Top-\(k\) & image-only saliency & 2.55 & 2.05x & 0.66 & 0.69 \\
Text-only query & text-only hidden state & 2.83 & 2.41x & 0.64 & 0.72 \\
TIGER routing-only & visual-conditioned query & 3.02 & 2.76x & 0.63 & 0.73 \\
TIGER full & visual-conditioned query & \textbf{3.37} & \textbf{3.24x} & \textbf{0.62} & \textbf{0.75} \\
\bottomrule
\end{tabular}%
}
\caption{
Fine-grained ablation of visual gated routing on COCO Caption.
Random and static selectors test whether reducing visual tokens alone is sufficient.
The text-only query variant tests whether multimodal query construction is necessary.
TIGER routing-only removes acceptance-aligned training, while TIGER full combines visual-conditioned routing with acceptance-aligned optimization.
}
\label{tab:routing-ablation}
\end{table}

The results show that token reduction alone is not sufficient: random Top-\(k\) and static image-only Top-\(k\) underperform text-conditioned variants.
The text-only query improves over non-conditioned selection, but still lags behind TIGER's visual-conditioned query, suggesting that multimodal query construction is useful for selecting evidence that supports drafter--verifier agreement.
TIGER full performs best overall, indicating that visual-conditioned routing and acceptance-aligned training are complementary.

\subsection{Qualitative Execution Traces}
\label{app:qualitative-trace}

\begin{figure*}[t]
\centering
\small
\renewcommand{\arraystretch}{1.18}
\setlength{\tabcolsep}{6pt}

\begin{tabular}{m{0.28\textwidth} m{0.67\textwidth}}
\toprule

\multicolumn{2}{l}{\textbf{Question:} What emotions might someone feel when seeing the flowers in this photo, and why?} \\
\midrule

\begin{minipage}[c]{\linewidth}
    \centering
    \includegraphics[width=0.92\linewidth]{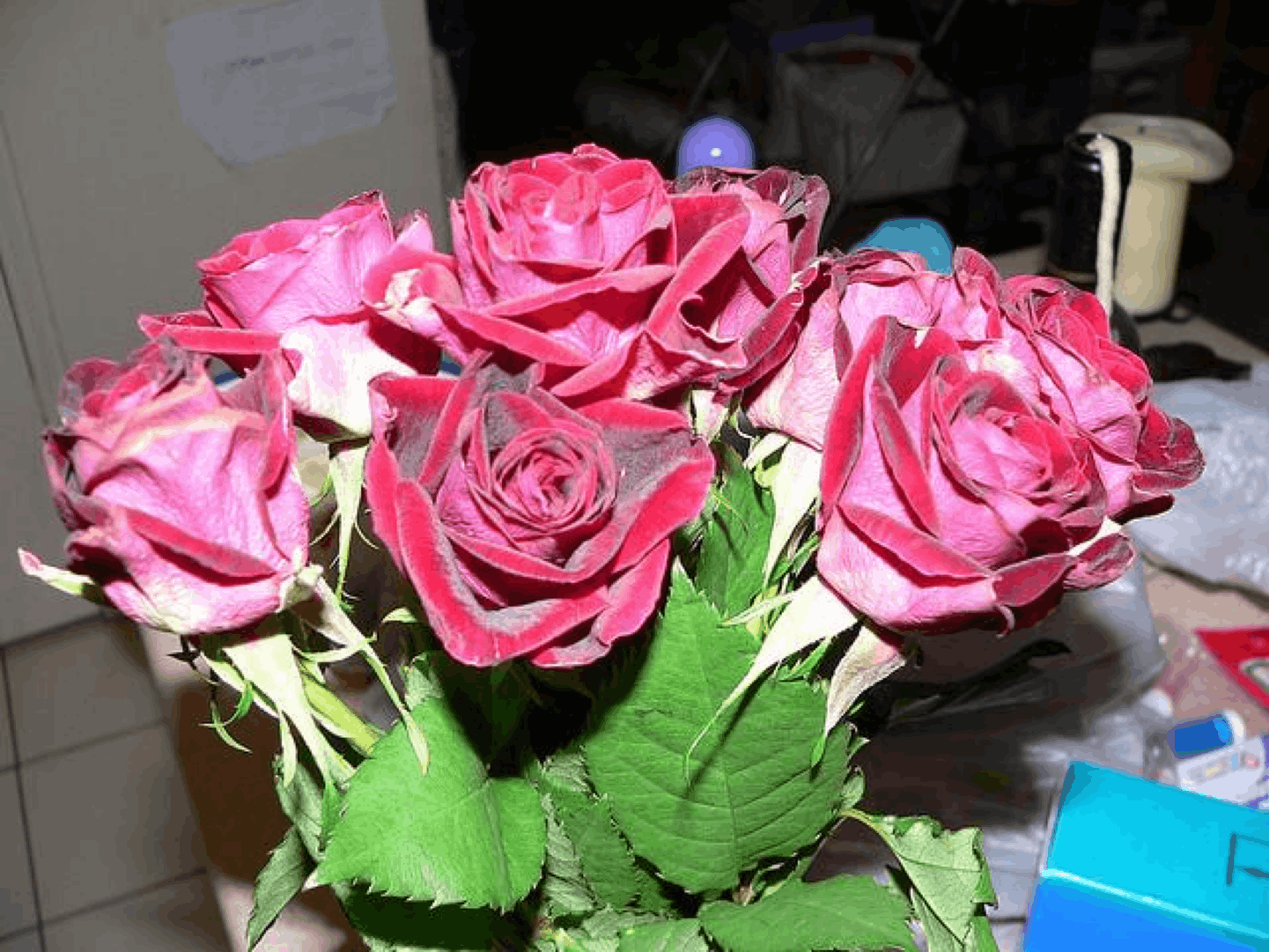}
    
    \vspace{2pt}
    \textbf{Input image}
\end{minipage}
&
\begin{minipage}[c]{\linewidth}
The input image shows a bouquet of red roses. We visualize how TIGER's routed visual tokens evolve across speculative decoding rounds and how verifier feedback shapes the committed output.
\end{minipage}
\\
\midrule

\begin{minipage}[c]{\linewidth}
    \centering
    \includegraphics[width=0.92\linewidth]{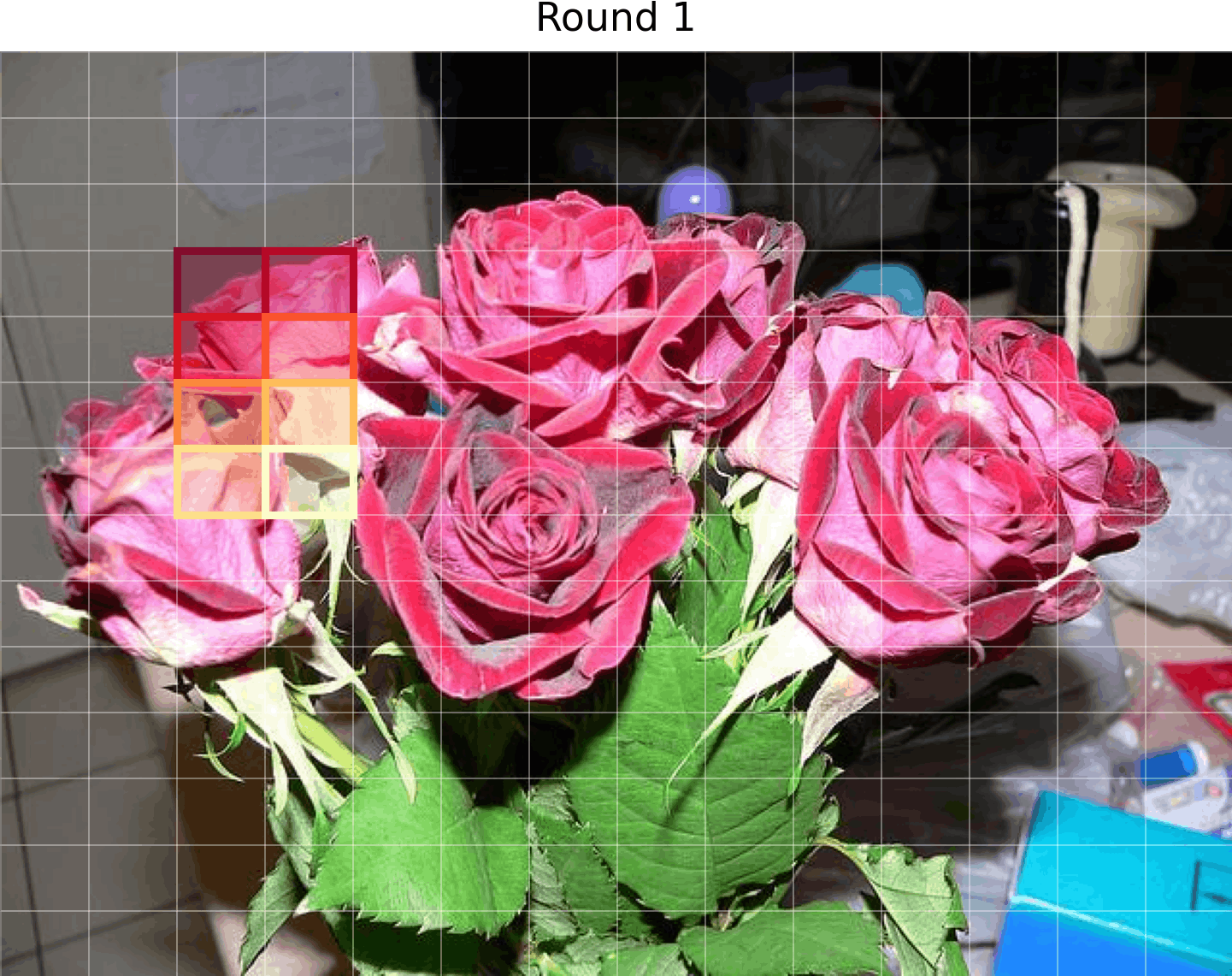}
    
    \vspace{2pt}
    \textbf{Round 1 routing}
\end{minipage}
&
\begin{minipage}[c]{\linewidth}
\textbf{Draft:} Someone might feel \emph{sadness}. \\
\textbf{Verifier:} Accepts only a short prefix; the emotionally loaded interpretation diverges early from the visual evidence. \\
\textbf{Committed output:} Someone might feel \dots
\end{minipage}
\\
\midrule

\begin{minipage}[c]{\linewidth}
    \centering
    \includegraphics[width=0.92\linewidth]{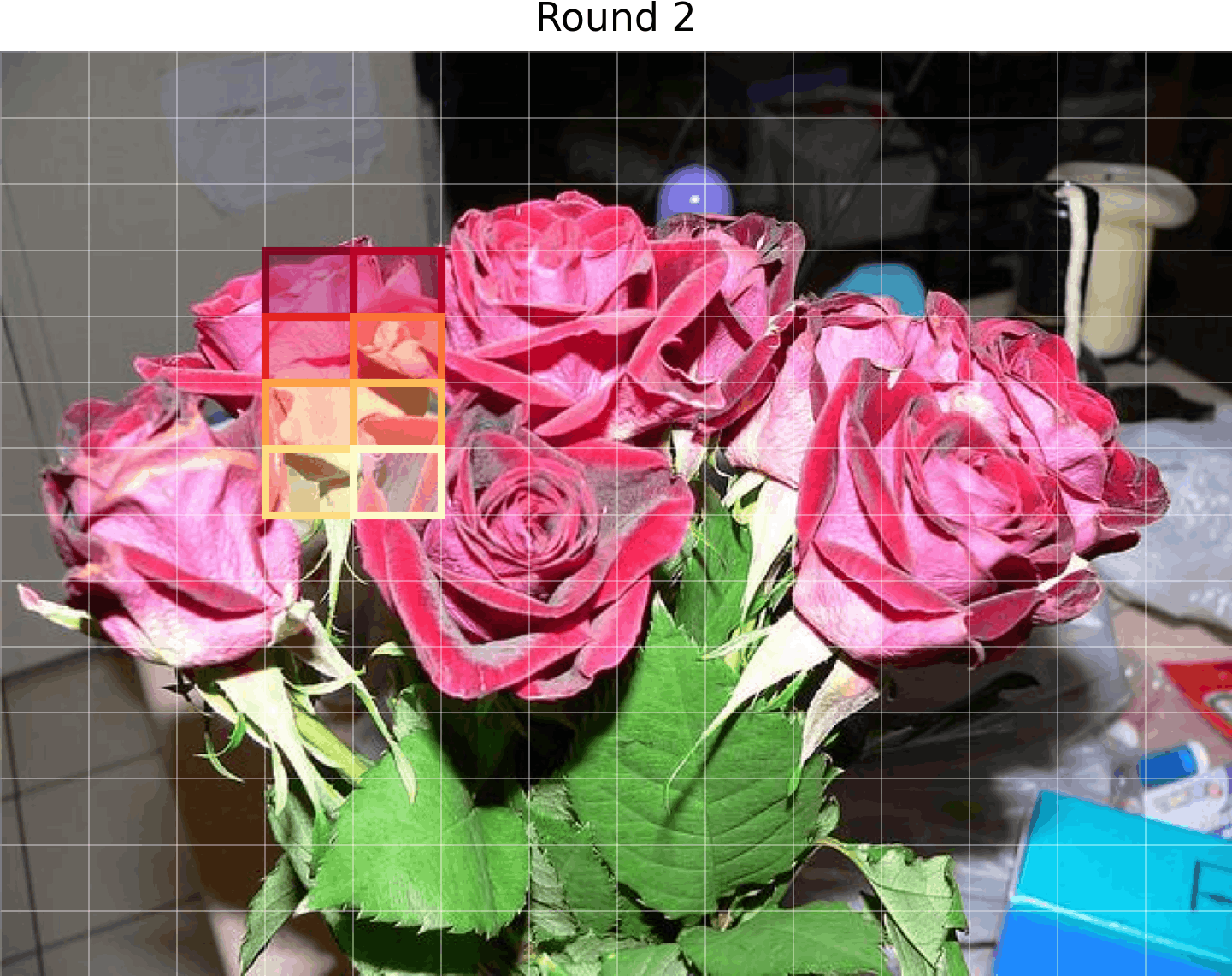}
    
    \vspace{2pt}
    \textbf{Round 2 routing}
\end{minipage}
&
\begin{minipage}[c]{\linewidth}
\textbf{Draft:} because the image shows \\
\textbf{Verifier:} Accepts the full drafted phrase, since it is grounded in the visible content of the image. \\
\textbf{Committed output:} Someone might feel \dots because the image shows
\end{minipage}
\\
\midrule

\begin{minipage}[c]{\linewidth}
    \centering
    \includegraphics[width=0.92\linewidth]{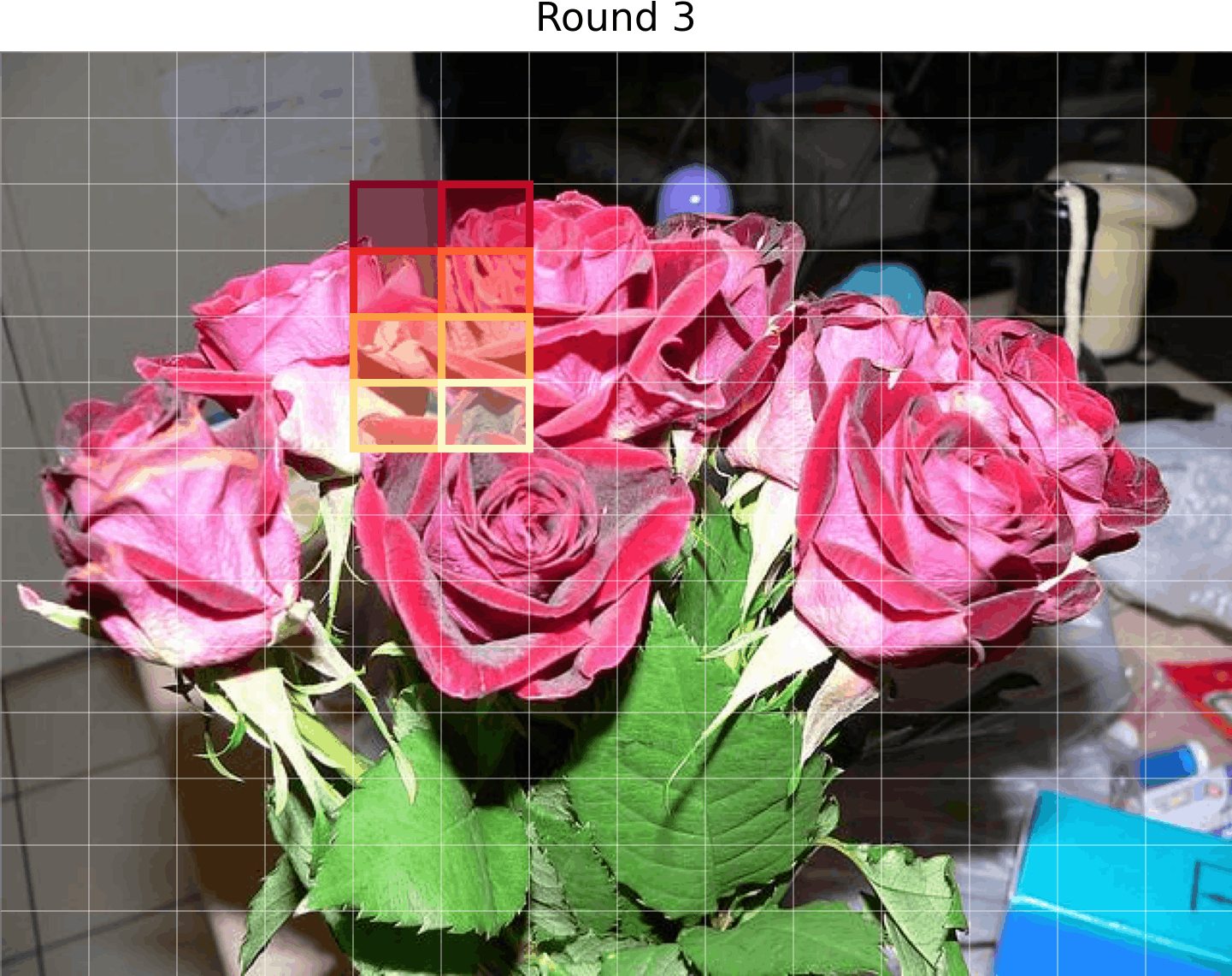}
    
    \vspace{2pt}
    \textbf{Round 3 routing}
\end{minipage}
&
\begin{minipage}[c]{\linewidth}
\textbf{Draft:} a bouquet of red \\
\textbf{Verifier:} Accepts the full drafted phrase; routing remains concentrated on flower-related regions. \\
\textbf{Committed output:} Someone might feel \dots because the image shows a bouquet of red
\end{minipage}
\\
\midrule

\begin{minipage}[c]{\linewidth}
    \centering
    \includegraphics[width=0.92\linewidth]{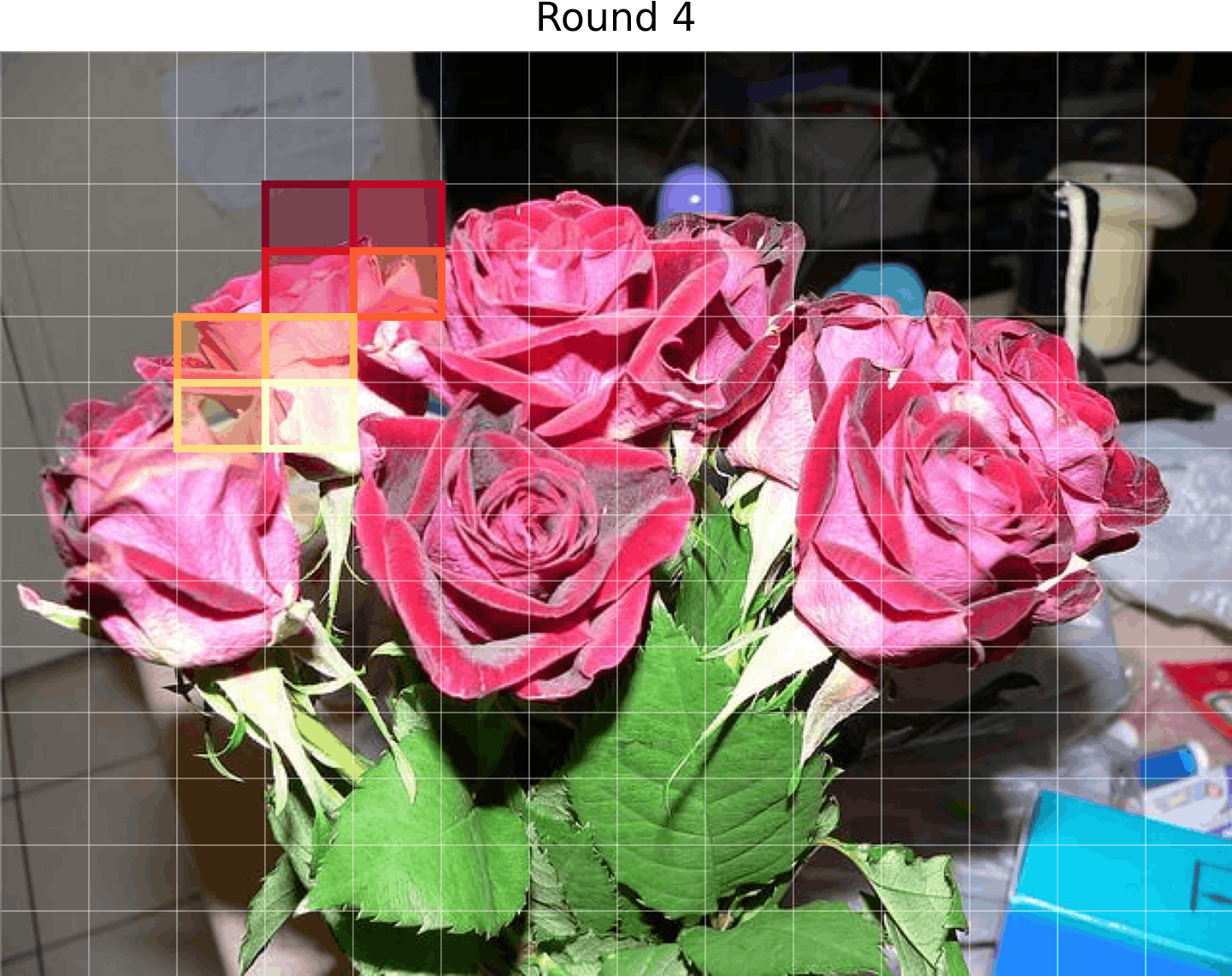}
    
    \vspace{2pt}
    \textbf{Round 4 routing}
\end{minipage}
&
\begin{minipage}[c]{\linewidth}
\textbf{Draft:} roses which may evoke \\
\textbf{Verifier:} Accepts a longer prefix than the baseline, preserving fluent and visually grounded continuation. \\
\textbf{Final output:} Someone might feel happiness or appreciation because the image shows a bouquet of red roses, which may evoke feelings of beauty or affection.
\end{minipage}
\\
\bottomrule
\end{tabular}

\vspace{6pt}

\caption{
Execution trace of TIGER on a qualitative example.
Each row shows the routed visual tokens selected at one speculative round together with the corresponding drafter output, verifier response, and committed text.
The routed regions shift dynamically with the evolving textual prefix, leading to progressively better visual grounding and longer verifier-accepted prefixes.
}
\label{fig:tiger_execution_trace}
\end{figure*}

\begin{figure*}[t]
\centering
\small
\renewcommand{\arraystretch}{1.18}
\setlength{\tabcolsep}{6pt}

\begin{tabular}{m{0.28\textwidth} m{0.67\textwidth}}
\toprule

\multicolumn{2}{l}{\textbf{Prompt:} Please carefully observe the image and come up with a caption for the image.} \\
\midrule

\begin{minipage}[c]{\linewidth}
    \centering
    \includegraphics[width=0.92\linewidth]{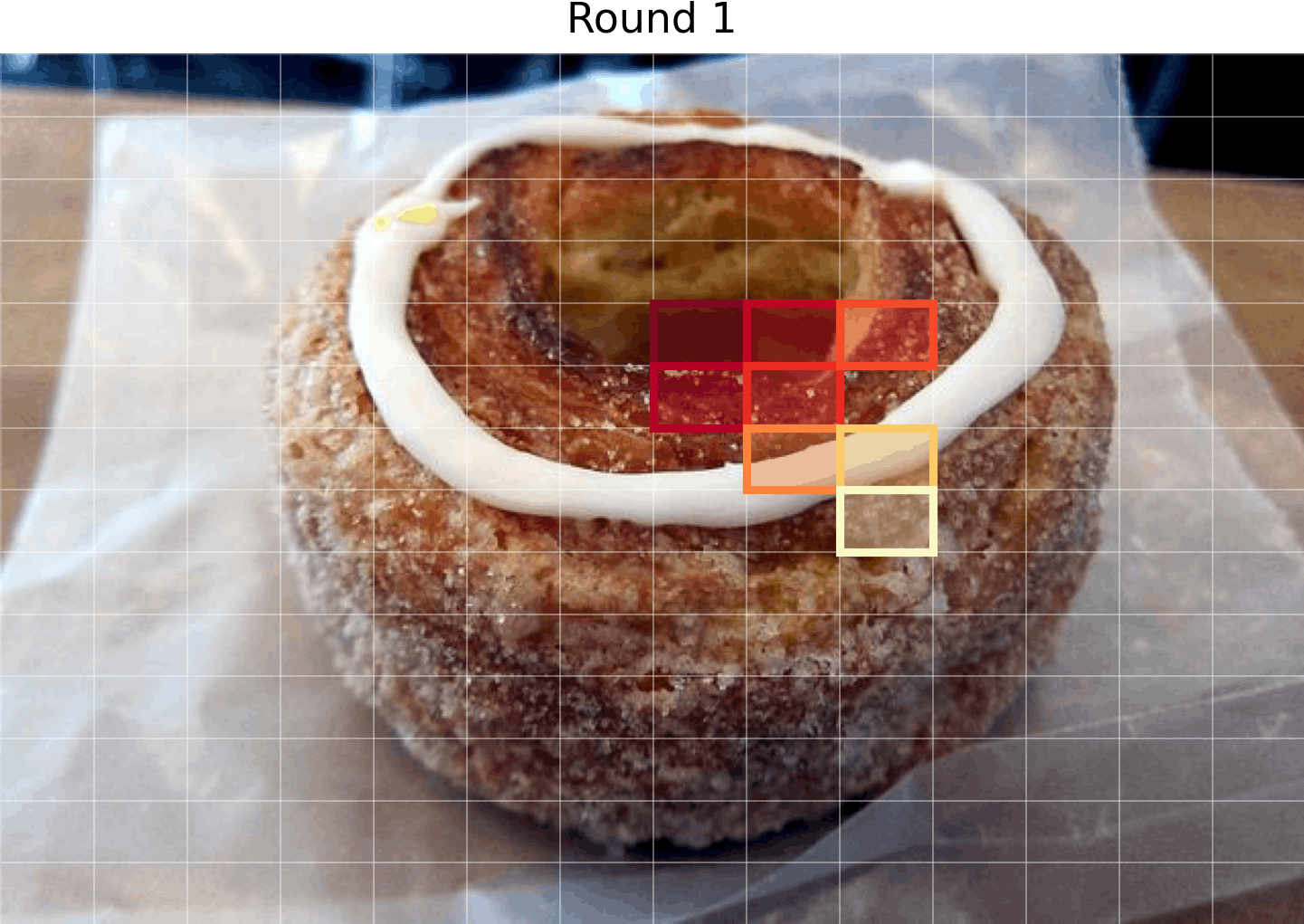}
    
    \vspace{2pt}
    \textbf{Round 1 routing}
\end{minipage}
&
\begin{minipage}[c]{\linewidth}
\textbf{Prefix:} Please carefully observe the image and come up with a caption for the image. \\[2pt]
\textbf{Draft block:} Cinnamon roll with icing \\
\textbf{Accepted prefix:} Cinnamon roll with icing \\
\textbf{Committed output:} Please carefully observe the image and come up with a caption for the image. Cinnamon roll with icing
\end{minipage}
\\
\midrule

\begin{minipage}[c]{\linewidth}
    \centering
    \includegraphics[width=0.92\linewidth]{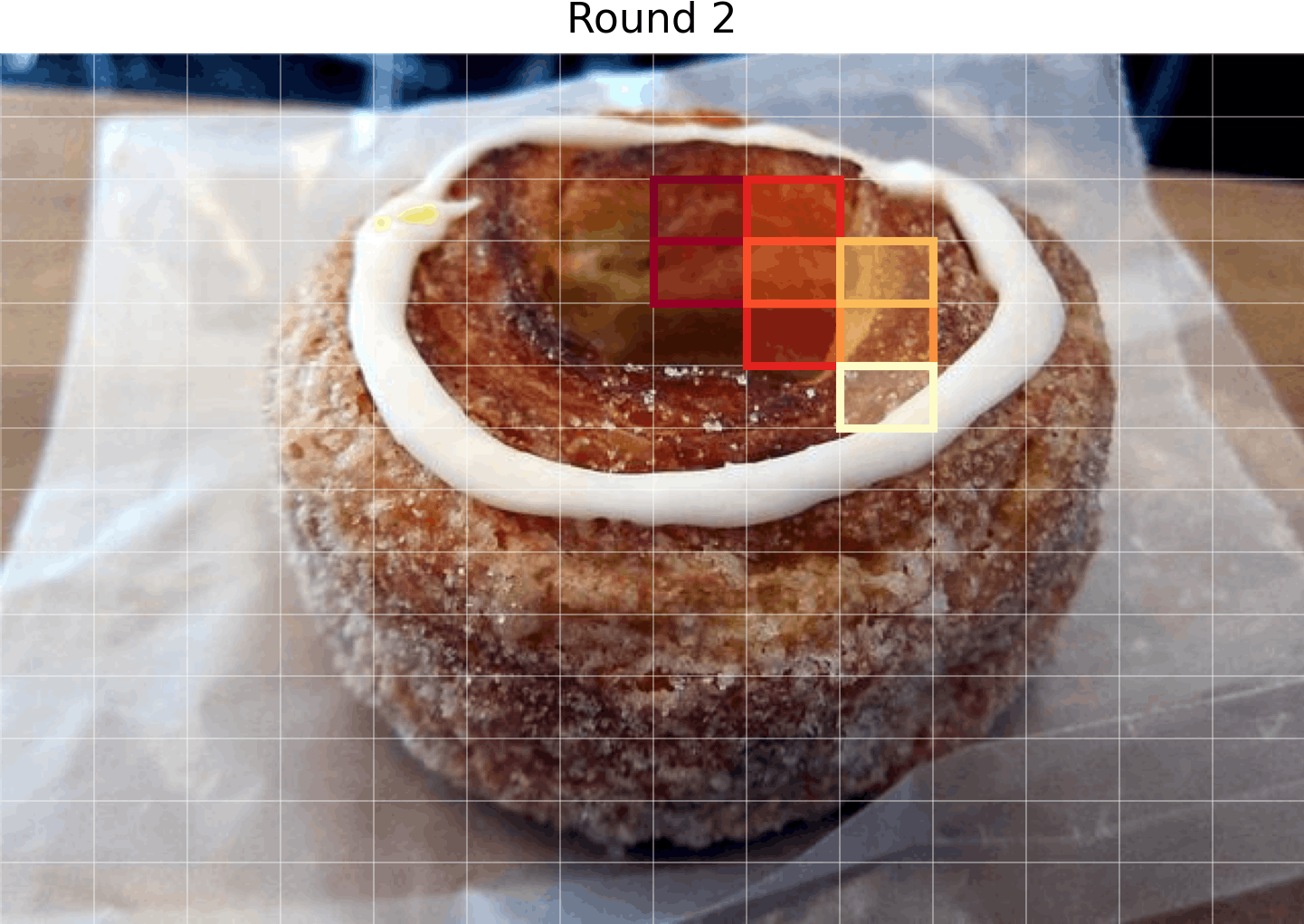}
    
    \vspace{2pt}
    \textbf{Round 2 routing}
\end{minipage}
&
\begin{minipage}[c]{\linewidth}
\textbf{Prefix:} Please carefully observe the image and come up with a caption for the image. Cinnamon roll with icing \\[2pt]
\textbf{Draft block:} on a red square \\
\textbf{Accepted prefix:} on a \\
\textbf{Rejected suffix:} red square \\
\textbf{Verifier replacement:} square \\
\textbf{Committed output:} Please carefully observe the image and come up with a caption for the image. Cinnamon roll with icing on a square
\end{minipage}
\\
\midrule

\begin{minipage}[c]{\linewidth}
    \centering
    \includegraphics[width=0.92\linewidth]{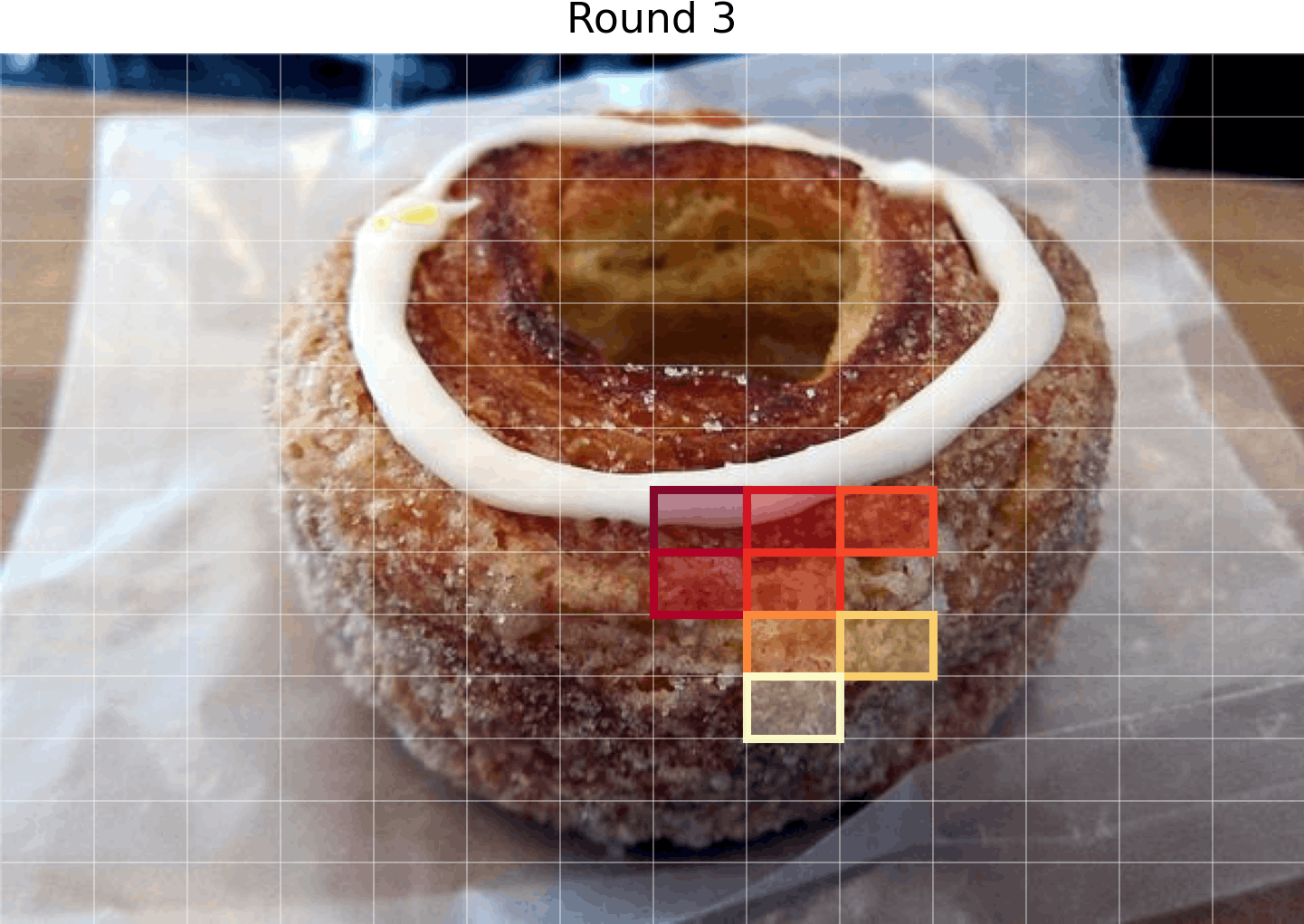}
    
    \vspace{2pt}
    \textbf{Round 3 routing}
\end{minipage}
&
\begin{minipage}[c]{\linewidth}
\textbf{Prefix:} Please carefully observe the image and come up with a caption for the image. Cinnamon roll with icing on a square \\[2pt]
\textbf{Draft block:} of wax paper. \\
\textbf{Accepted prefix:} of wax paper. \\
\textbf{Committed output:} Please carefully observe the image and come up with a caption for the image. Cinnamon roll with icing on a square of wax paper.
\end{minipage}
\\
\bottomrule
\end{tabular}

\caption{
Execution trace of TIGER across speculative rounds on a captioning example.
Each row shows the routed visual tokens selected at one round together with the current prefix, draft block, verifier decision, and committed output.
The routed regions change with the evolving textual prefix, helping the drafter produce visually grounded continuations such as ``cinnamon roll with icing'' and ``square of wax paper.''
}
\label{fig:tiger_cinnamon_trace}
\end{figure*}

\begin{figure*}[t]
\centering
\small
\renewcommand{\arraystretch}{1.18}
\setlength{\tabcolsep}{6pt}

\begin{tabular}{m{0.28\textwidth} m{0.67\textwidth}}
\toprule

\multicolumn{2}{l}{\textbf{Question:} Select the organism in the same genus as the Goliath heron.} \\
\multicolumn{2}{l}{\textbf{Options:} 0) Falco sparverius \quad 1) Ardea herodias \quad 2) Tigrisoma mexicanum} \\
\midrule

\begin{minipage}[c]{\linewidth}
    \centering
    \includegraphics[width=0.92\linewidth]{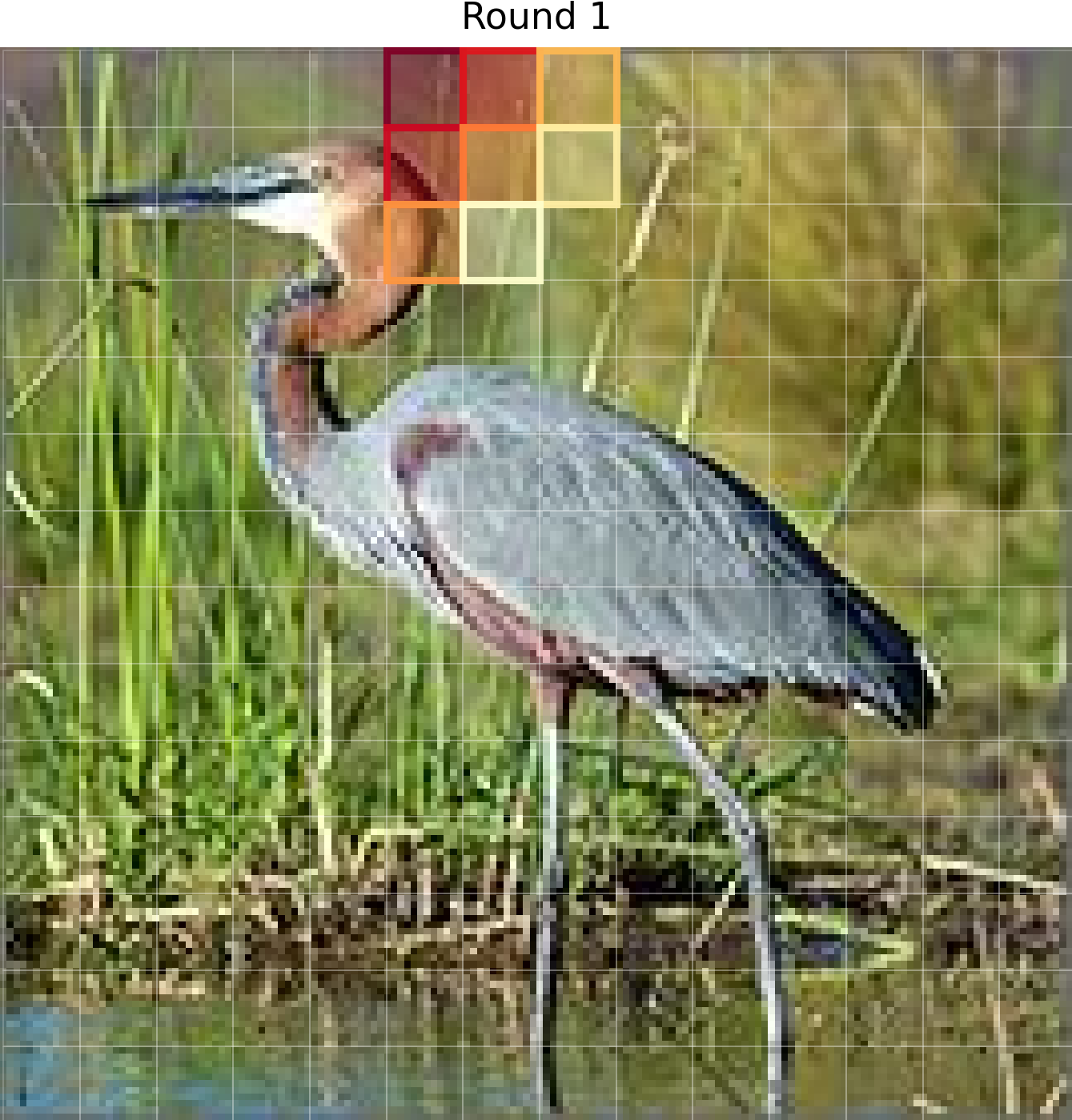}
    
    \vspace{2pt}
    \textbf{Round 1 routing}
\end{minipage}
&
\begin{minipage}[c]{\linewidth}
\textbf{Prefix:} Select the organism in the same genus as the Goliath heron. Options: 0) Falco sparverius 1) Ardea herodias 2) Tigrisoma mexicanum Answer: \\[2pt]
\textbf{Draft block:} The answer is Ardea \\
\textbf{Accepted prefix:} The answer is Ardea \\
\textbf{Committed output:} Select the organism in the same genus as the Goliath heron. Options: 0) Falco sparverius 1) Ardea herodias 2) Tigrisoma mexicanum Answer: The answer is Ardea
\end{minipage}
\\
\midrule

\begin{minipage}[c]{\linewidth}
    \centering
    \includegraphics[width=0.92\linewidth]{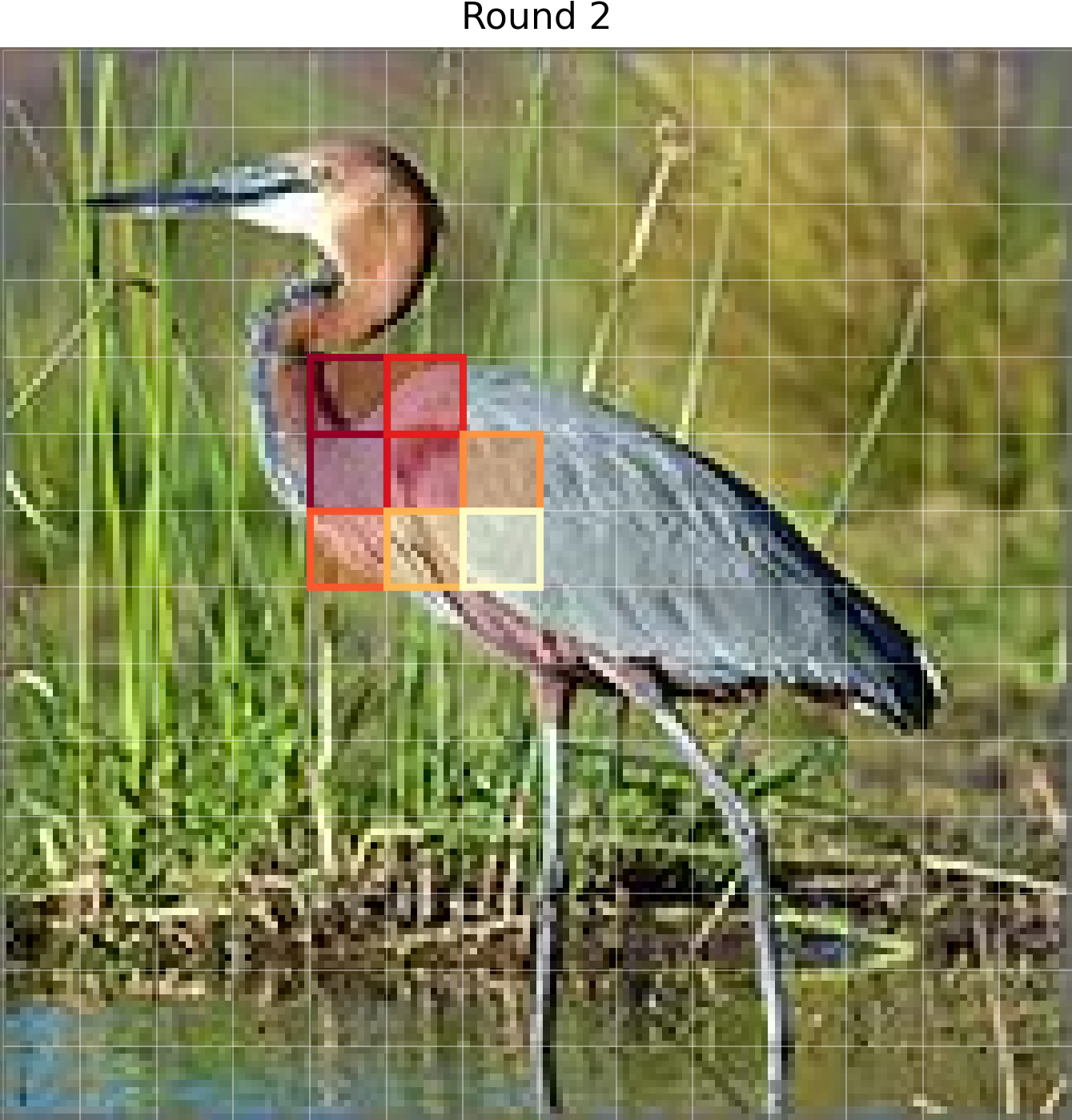}
    
    \vspace{2pt}
    \textbf{Round 2 routing}
\end{minipage}
&
\begin{minipage}[c]{\linewidth}
\textbf{Prefix:} Select the organism in the same genus as the Goliath heron. Options: 0) Falco sparverius 1) Ardea herodias 2) Tigrisoma mexicanum Answer: The answer is Ardea \\[2pt]
\textbf{Draft block:} herodias, which shares \\
\textbf{Accepted prefix:} herodias \\
\textbf{Rejected suffix:} , which shares \\
\textbf{Verifier replacement:} . \\
\textbf{Committed output:} Select the organism in the same genus as the Goliath heron. Options: 0) Falco sparverius 1) Ardea herodias 2) Tigrisoma mexicanum Answer: The answer is Ardea herodias.
\end{minipage}
\\
\midrule

\begin{minipage}[c]{\linewidth}
    \centering
    \includegraphics[width=0.92\linewidth]{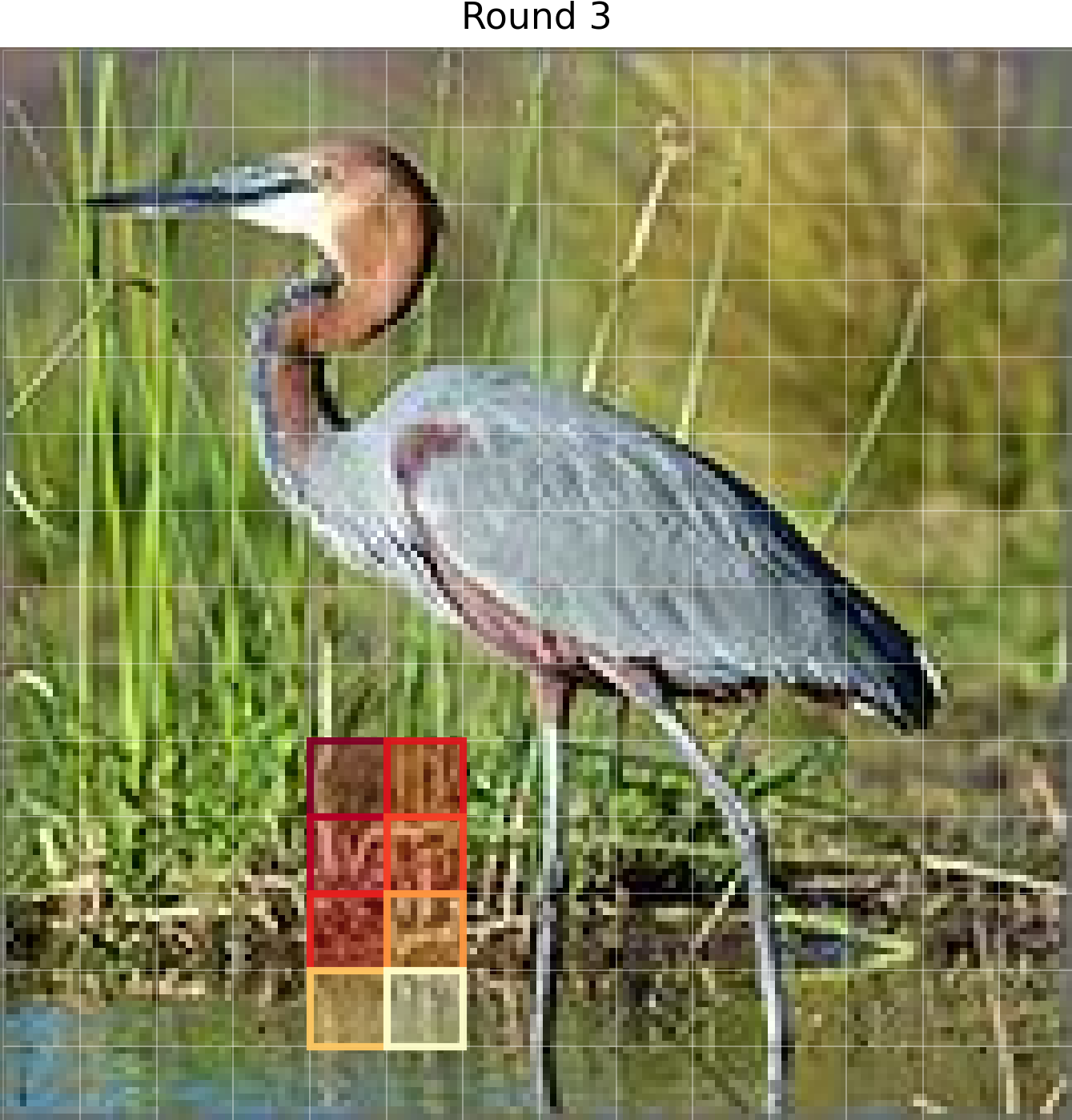}
    
    \vspace{2pt}
    \textbf{Round 3 routing}
\end{minipage}
&
\begin{minipage}[c]{\linewidth}
\textbf{Prefix:} Select the organism in the same genus as the Goliath heron. Options: 0) Falco sparverius 1) Ardea herodias 2) Tigrisoma mexicanum Answer: The answer is Ardea herodias. \\[2pt]
\textbf{Draft block:} It belongs to genus \\
\textbf{Accepted prefix:} [none] \\
\textbf{Rejected suffix:} It belongs to genus \\
\textbf{Verifier replacement:} [EOS] \\
\textbf{Committed output:} Select the organism in the same genus as the Goliath heron. Options: 0) Falco sparverius 1) Ardea herodias 2) Tigrisoma mexicanum Answer: The answer is Ardea herodias. [EOS]
\end{minipage}
\\
\bottomrule
\end{tabular}

\caption{
Execution trace of TIGER on a genus-classification example involving the Goliath heron.
Each row shows the routed visual tokens selected at one speculative round together with the current prefix, draft block, verifier decision, and committed output.
TIGER first accepts the visually and taxonomically grounded answer ``Ardea herodias,'' then rejects an unnecessary explanatory continuation and terminates generation.
}
\label{fig:tiger_heron_trace}
\end{figure*}

\begin{figure*}[t]
\centering
\footnotesize
\renewcommand{\arraystretch}{1.16}
\setlength{\tabcolsep}{6pt}

\begin{tabular}{m{0.28\textwidth} m{0.67\textwidth}}
\toprule

\multicolumn{2}{l}{\textbf{Question:} How many dogs are touching their muzzles together in one of the images?} \\
\multicolumn{2}{l}{\textbf{Options:} A) Three \quad B) None of the choices provided \quad C) Four \quad D) Two \quad E) One} \\
\midrule

\begin{minipage}[c]{\linewidth}
    \centering
    \includegraphics[width=0.92\linewidth]{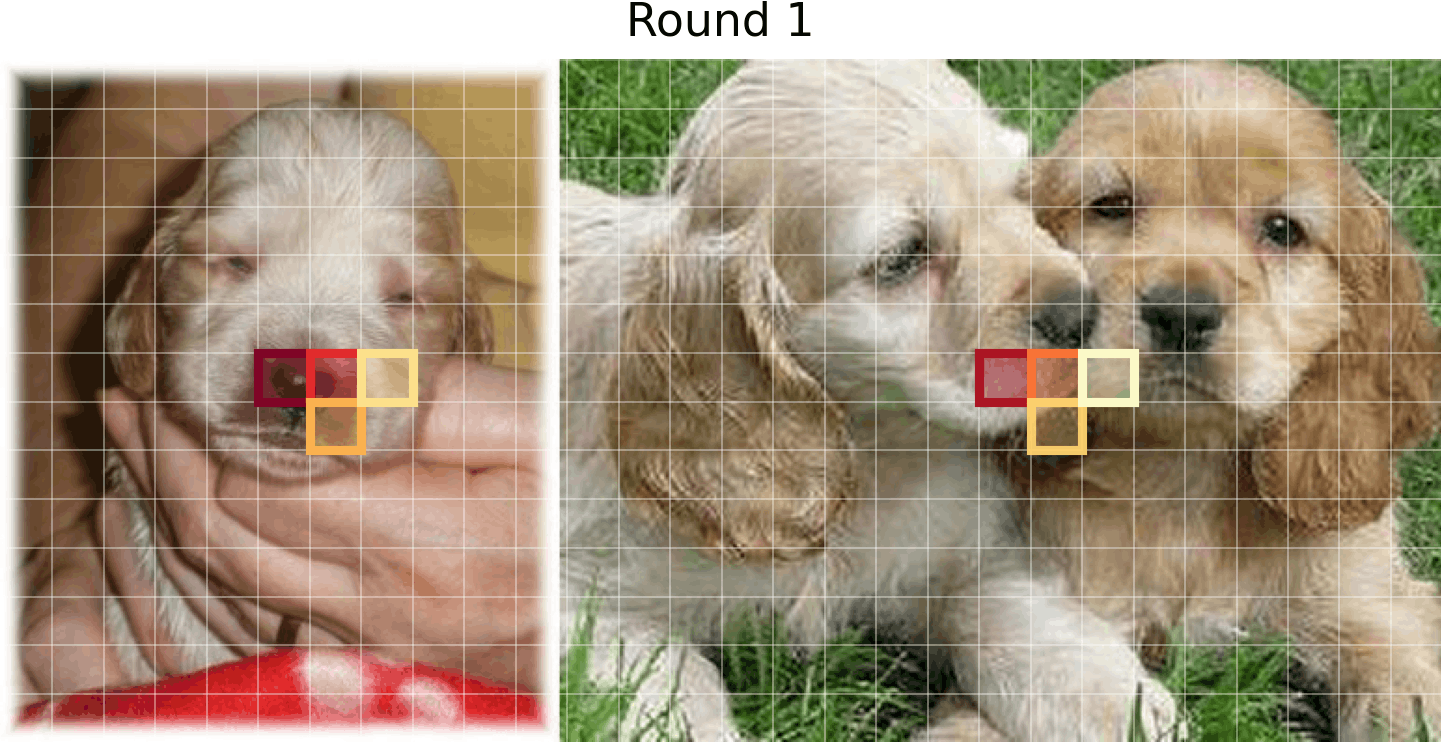}
    
    \vspace{2pt}
    \textbf{Round 1 routing}
\end{minipage}
&
\begin{minipage}[c]{\linewidth}
\textbf{Prefix:} How many dogs are touching their muzzles together in one of the images? Options: A) Three B) None of the choices provided C) Four D) Two E) One Answer: \\[2pt]
\textbf{Draft block:} The answer is Two \\
\textbf{Accepted prefix:} The answer is Two \\
\textbf{Committed output:} How many dogs are touching their muzzles together in one of the images? Options: A) Three B) None of the choices provided C) Four D) Two E) One Answer: The answer is Two
\end{minipage}
\\
\midrule

\begin{minipage}[c]{\linewidth}
    \centering
    \includegraphics[width=0.92\linewidth]{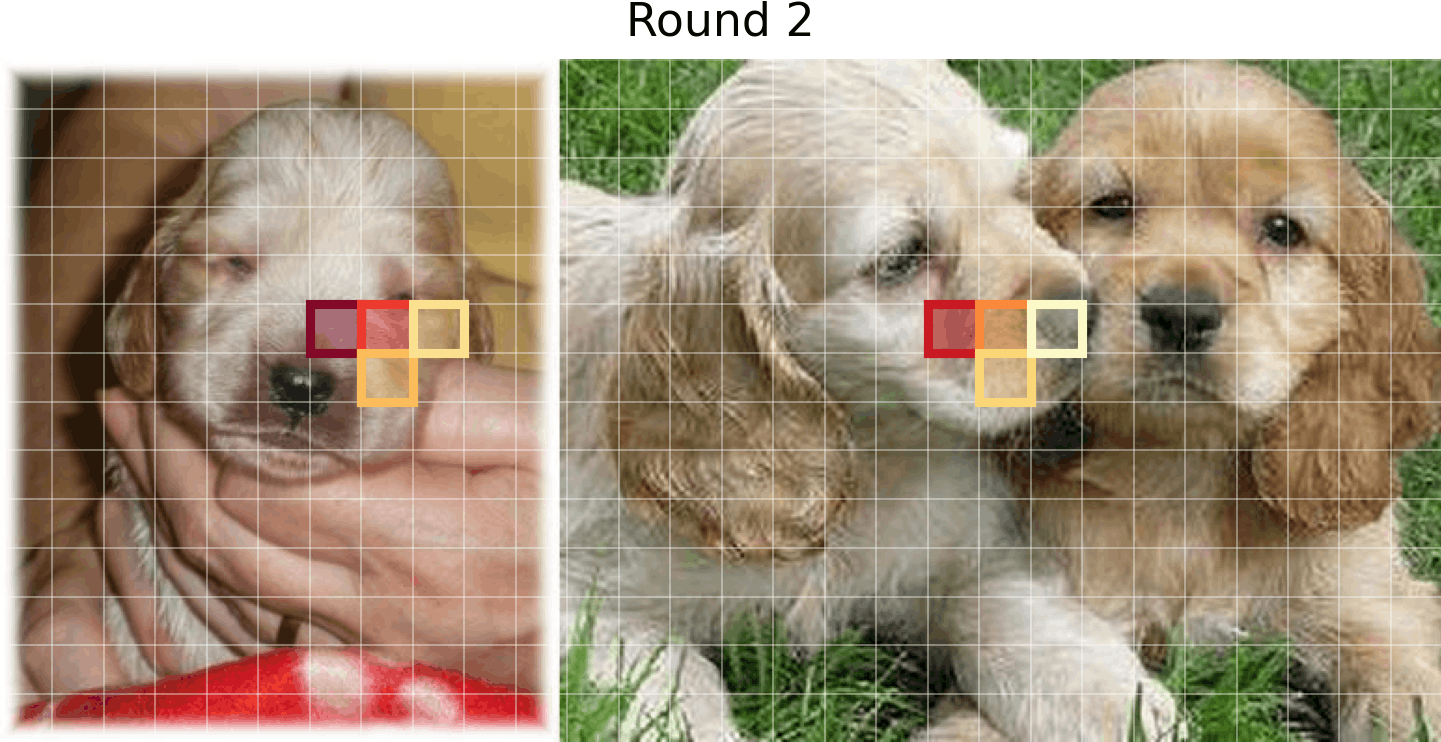}
    
    \vspace{2pt}
    \textbf{Round 2 routing}
\end{minipage}
&
\begin{minipage}[c]{\linewidth}
\textbf{Prefix:} How many dogs are touching their muzzles together in one of the images? Options: A) Three B) None of the choices provided C) Four D) Two E) One Answer: The answer is Two \\[2pt]
\textbf{Draft block:} dogs are touching muzzles \\
\textbf{Accepted prefix:} dogs \\
\textbf{Rejected suffix:} are touching muzzles \\
\textbf{Verifier replacement:} . \\
\textbf{Committed output:} How many dogs are touching their muzzles together in one of the images? Options: A) Three B) None of the choices provided C) Four D) Two E) One Answer: The answer is Two dogs.
\end{minipage}
\\
\midrule

\begin{minipage}[c]{\linewidth}
    \centering
    \includegraphics[width=0.92\linewidth]{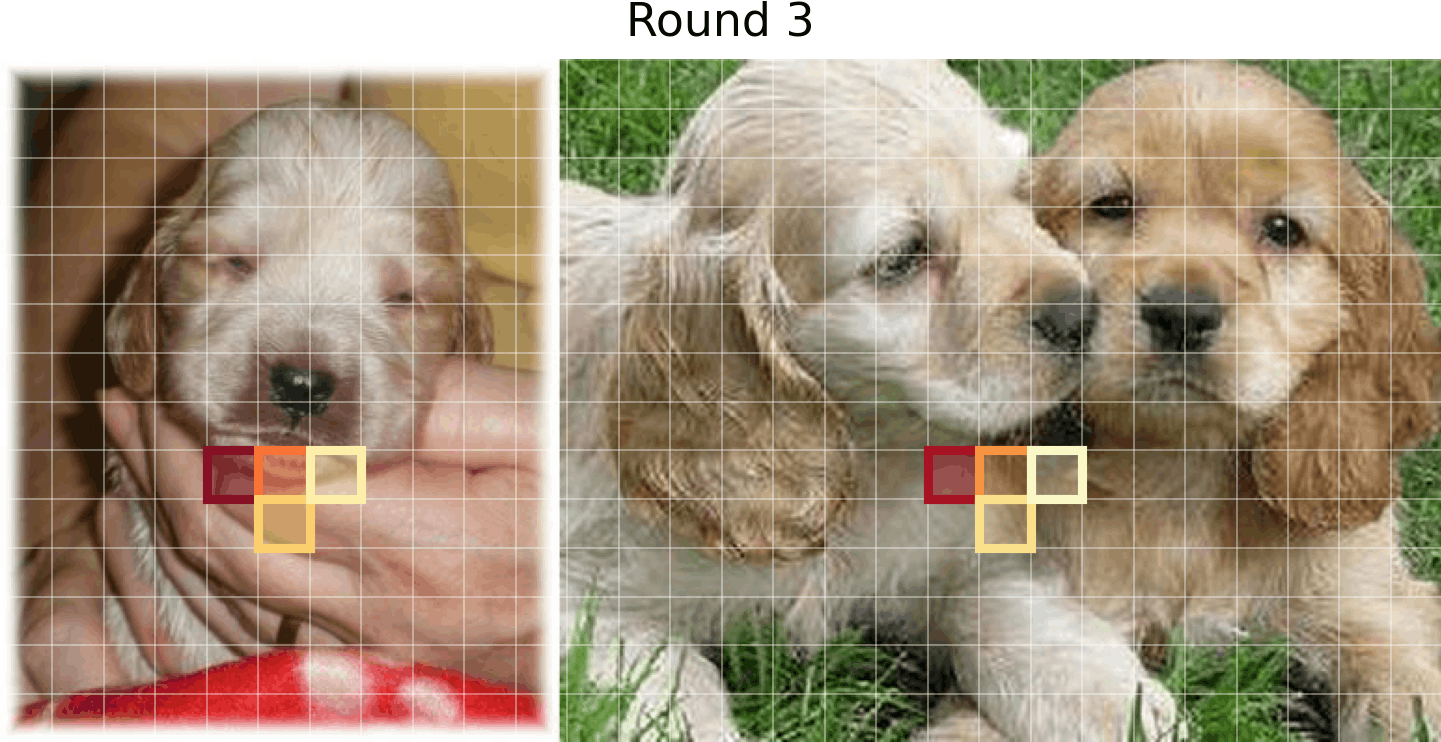}
    
    \vspace{2pt}
    \textbf{Round 3 routing}
\end{minipage}
&
\begin{minipage}[c]{\linewidth}
\textbf{Prefix:} How many dogs are touching their muzzles together in one of the images? Options: A) Three B) None of the choices provided C) Four D) Two E) One Answer: The answer is Two dogs. \\[2pt]
\textbf{Draft block:} There are two dogs \\
\textbf{Accepted prefix:} [none] \\
\textbf{Rejected suffix:} There are two dogs \\
\textbf{Verifier replacement:} [EOS] \\
\textbf{Committed output:} How many dogs are touching their muzzles together in one of the images? Options: A) Three B) None of the choices provided C) Four D) Two E) One Answer: The answer is Two dogs. [EOS]
\end{minipage}
\\
\bottomrule
\end{tabular}

\caption{
Execution trace of TIGER on a two-image visual reasoning example.
Each row shows the routed visual tokens selected at one speculative round together with the current prefix, draft block, verifier decision, and committed output.
TIGER routes evidence across both images, accepts the answer ``Two dogs,'' and rejects an unnecessary explanatory continuation before terminating generation.
}
\label{fig:tiger_dogs_multiimage_trace}
\end{figure*}

\begin{figure*}[t]
\centering
\scriptsize
\renewcommand{\arraystretch}{1.14}
\setlength{\tabcolsep}{6pt}

\begin{tabular}{m{0.28\textwidth} m{0.67\textwidth}}
\toprule

\multicolumn{2}{l}{\textbf{Question:} Which animal is also adapted to be camouflaged in the snow?} \\
\multicolumn{2}{l}{\textbf{Options:} 0) short-tailed weasel \quad 1) common hawk-cuckoo} \\
\multicolumn{2}{l}{\textbf{Context:} Arctic wolves live in the Canadian Arctic and Greenland. The Arctic wolf is adapted to be camouflaged in the snow.} \\
\midrule

\begin{minipage}[c]{\linewidth}
    \centering
    \includegraphics[width=0.92\linewidth]{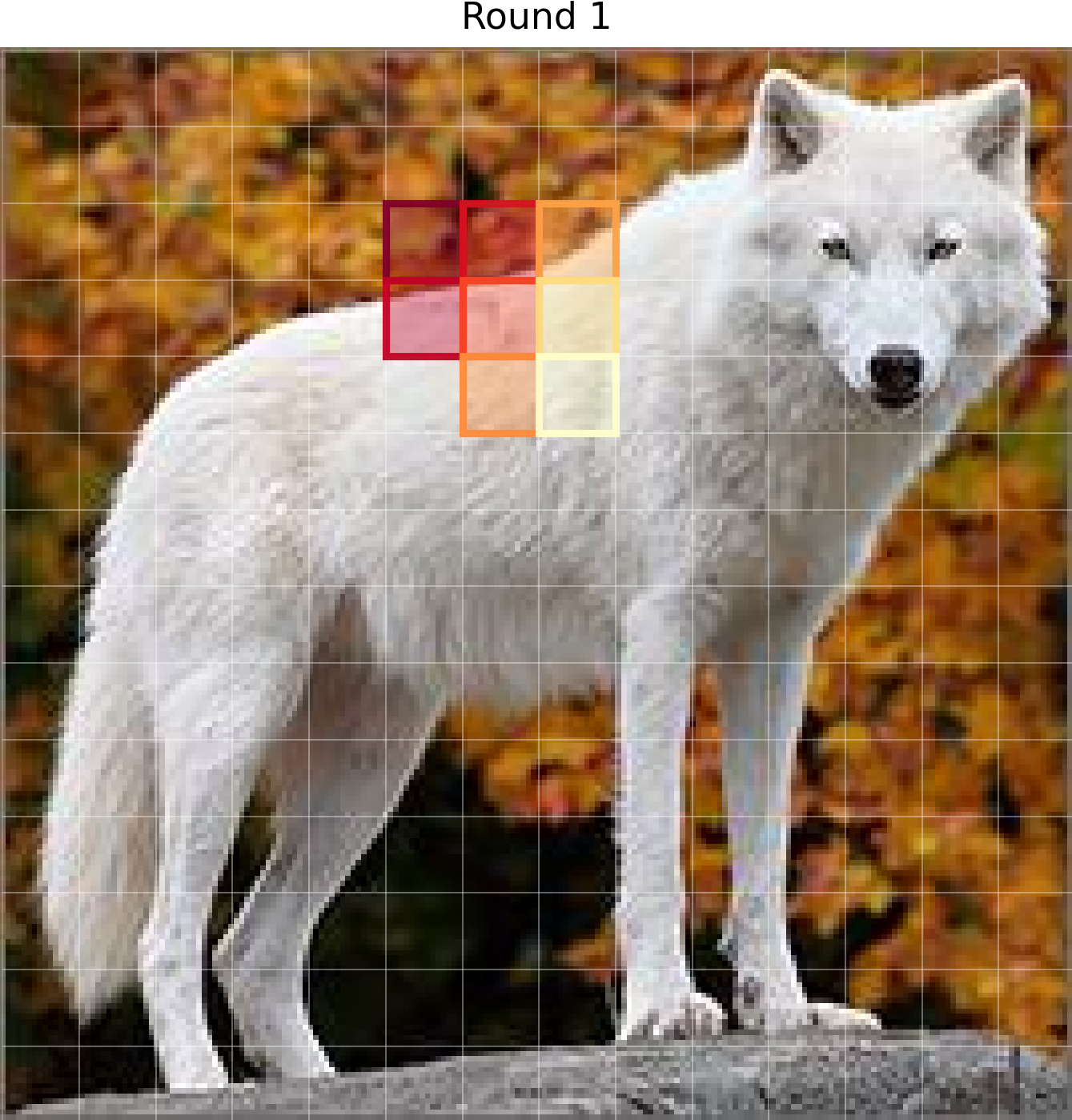}
    
    \vspace{2pt}
    \textbf{Round 1 routing}
\end{minipage}
&
\begin{minipage}[c]{\linewidth}
\textbf{Prefix:} Which animal is also adapted to be camouflaged in the snow? Options: 0) short-tailed weasel 1) common hawk-cuckoo Context: Arctic wolves live in the Canadian Arctic and Greenland. The Arctic wolf is adapted to be camouflaged in the snow. Answer: \\[2pt]
\textbf{Draft block:} The answer is short-tailed \\
\textbf{Accepted prefix:} The answer is short-tailed \\
\textbf{Committed output:} Which animal is also adapted to be camouflaged in the snow? Options: 0) short-tailed weasel 1) common hawk-cuckoo Context: Arctic wolves live in the Canadian Arctic and Greenland. The Arctic wolf is adapted to be camouflaged in the snow. Answer: The answer is short-tailed
\end{minipage}
\\
\midrule

\begin{minipage}[c]{\linewidth}
    \centering
    \includegraphics[width=0.92\linewidth]{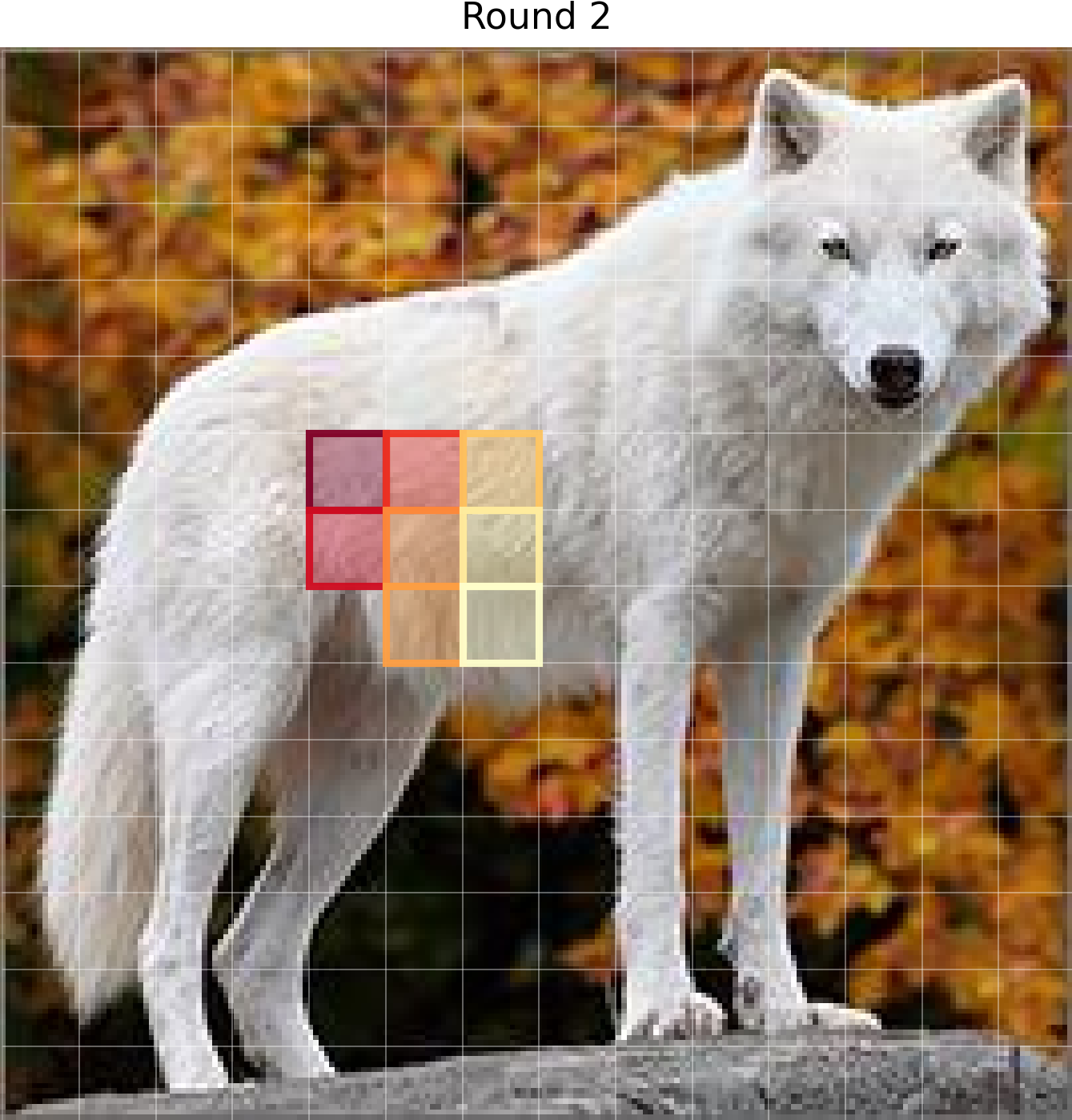}
    
    \vspace{2pt}
    \textbf{Round 2 routing}
\end{minipage}
&
\begin{minipage}[c]{\linewidth}
\textbf{Prefix:} Which animal is also adapted to be camouflaged in the snow? Options: 0) short-tailed weasel 1) common hawk-cuckoo Context: Arctic wolves live in the Canadian Arctic and Greenland. The Arctic wolf is adapted to be camouflaged in the snow. Answer: The answer is short-tailed \\[2pt]
\textbf{Draft block:} weasel, which also \\
\textbf{Accepted prefix:} weasel \\
\textbf{Rejected suffix:} , which also \\
\textbf{Verifier replacement:} . \\
\textbf{Committed output:} Which animal is also adapted to be camouflaged in the snow? Options: 0) short-tailed weasel 1) common hawk-cuckoo Context: Arctic wolves live in the Canadian Arctic and Greenland. The Arctic wolf is adapted to be camouflaged in the snow. Answer: The answer is short-tailed weasel.
\end{minipage}
\\
\midrule

\begin{minipage}[c]{\linewidth}
    \centering
    \includegraphics[width=0.92\linewidth]{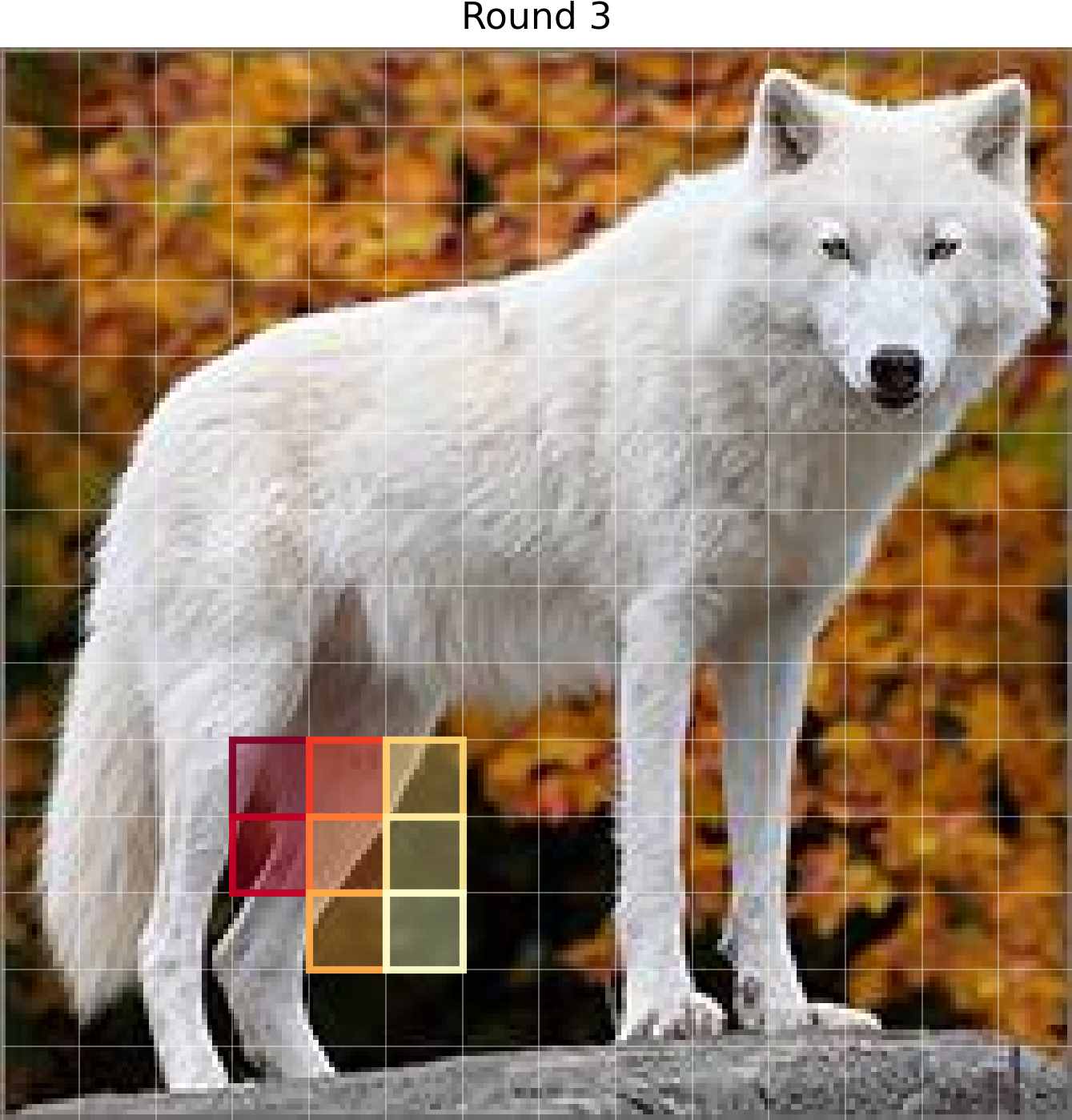}
    
    \vspace{2pt}
    \textbf{Round 3 routing}
\end{minipage}
&
\begin{minipage}[c]{\linewidth}
\textbf{Prefix:} Which animal is also adapted to be camouflaged in the snow? Options: 0) short-tailed weasel 1) common hawk-cuckoo Context: Arctic wolves live in the Canadian Arctic and Greenland. The Arctic wolf is adapted to be camouflaged in the snow. Answer: The answer is short-tailed weasel. \\[2pt]
\textbf{Draft block:} Both turn white in \\
\textbf{Accepted prefix:} [none] \\
\textbf{Rejected suffix:} Both turn white in \\
\textbf{Verifier replacement:} [EOS] \\
\textbf{Committed output:} Which animal is also adapted to be camouflaged in the snow? Options: 0) short-tailed weasel 1) common hawk-cuckoo Context: Arctic wolves live in the Canadian Arctic and Greenland. The Arctic wolf is adapted to be camouflaged in the snow. Answer: The answer is short-tailed weasel. [EOS]
\end{minipage}
\\
\bottomrule
\end{tabular}

\caption{
Execution trace of TIGER on a snow-camouflage reasoning example involving an Arctic wolf.
Each row shows the routed visual tokens selected at one speculative round together with the current prefix, draft block, verifier decision, and committed output.
TIGER accepts the answer ``short-tailed weasel,'' rejects an unnecessary explanatory continuation, and terminates generation after the verifier emits \texttt{[EOS]}.
}
\label{fig:tiger_wolf_trace}
\end{figure*}

\begin{figure*}[t]
\centering
\scriptsize
\renewcommand{\arraystretch}{1.14}
\setlength{\tabcolsep}{6pt}

\begin{tabular}{m{0.28\textwidth} m{0.67\textwidth}}
\toprule

\multicolumn{2}{l}{\textbf{Question:} When does the man take a long exhaled breath in this video?} \\
\multicolumn{2}{l}{\textbf{Options:} A) When he is running a marathon \quad B) When he is solving a difficult math problem} \\
\multicolumn{2}{l}{\textbf{\phantom{Options:}} C) When he is releasing the tension \quad D) When he is pulling apart} \\
\midrule

\begin{minipage}[c]{\linewidth}
    \centering
    \includegraphics[width=0.92\linewidth]{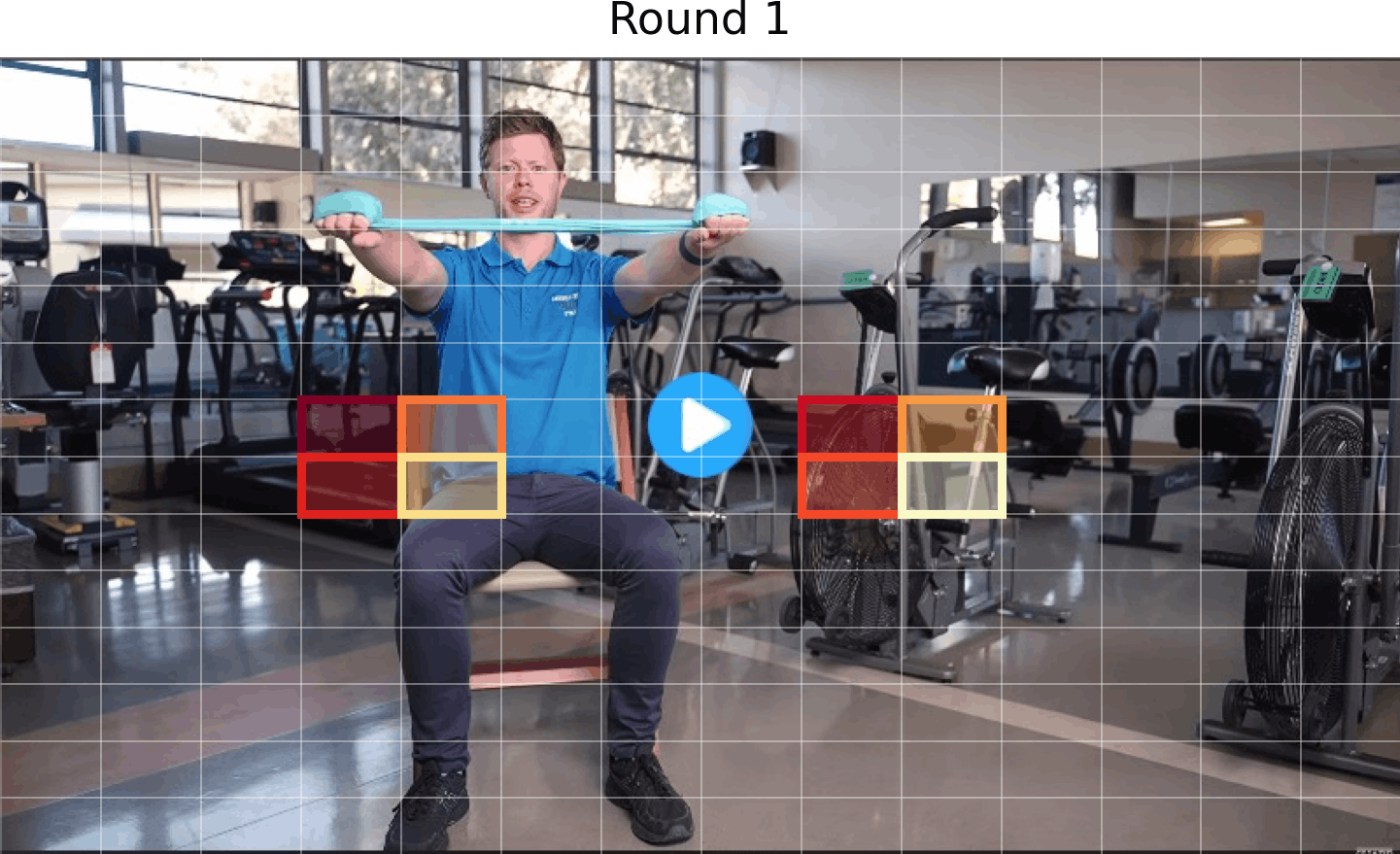}
    
    \vspace{2pt}
    \textbf{Round 1 routing}
\end{minipage}
&
\begin{minipage}[c]{\linewidth}
\textbf{Prefix:} When does the man take a long exhaled breath in this video? 
Options: A) When he is running a marathon. 
B) When he is solving a difficult math problem. 
C) When he is releasing the tension. 
D) When he is pulling apart. 
Answer: \\[2pt]
\textbf{Draft block:} The answer is D \\
\textbf{Accepted prefix:} The answer is D \\
\textbf{Committed output:} When does the man take a long exhaled breath in this video? 
Options: A) When he is running a marathon. 
B) When he is solving a difficult math problem. 
C) When he is releasing the tension. 
D) When he is pulling apart. 
Answer: The answer is D
\end{minipage}
\\
\midrule

\begin{minipage}[c]{\linewidth}
    \centering
    \includegraphics[width=0.92\linewidth]{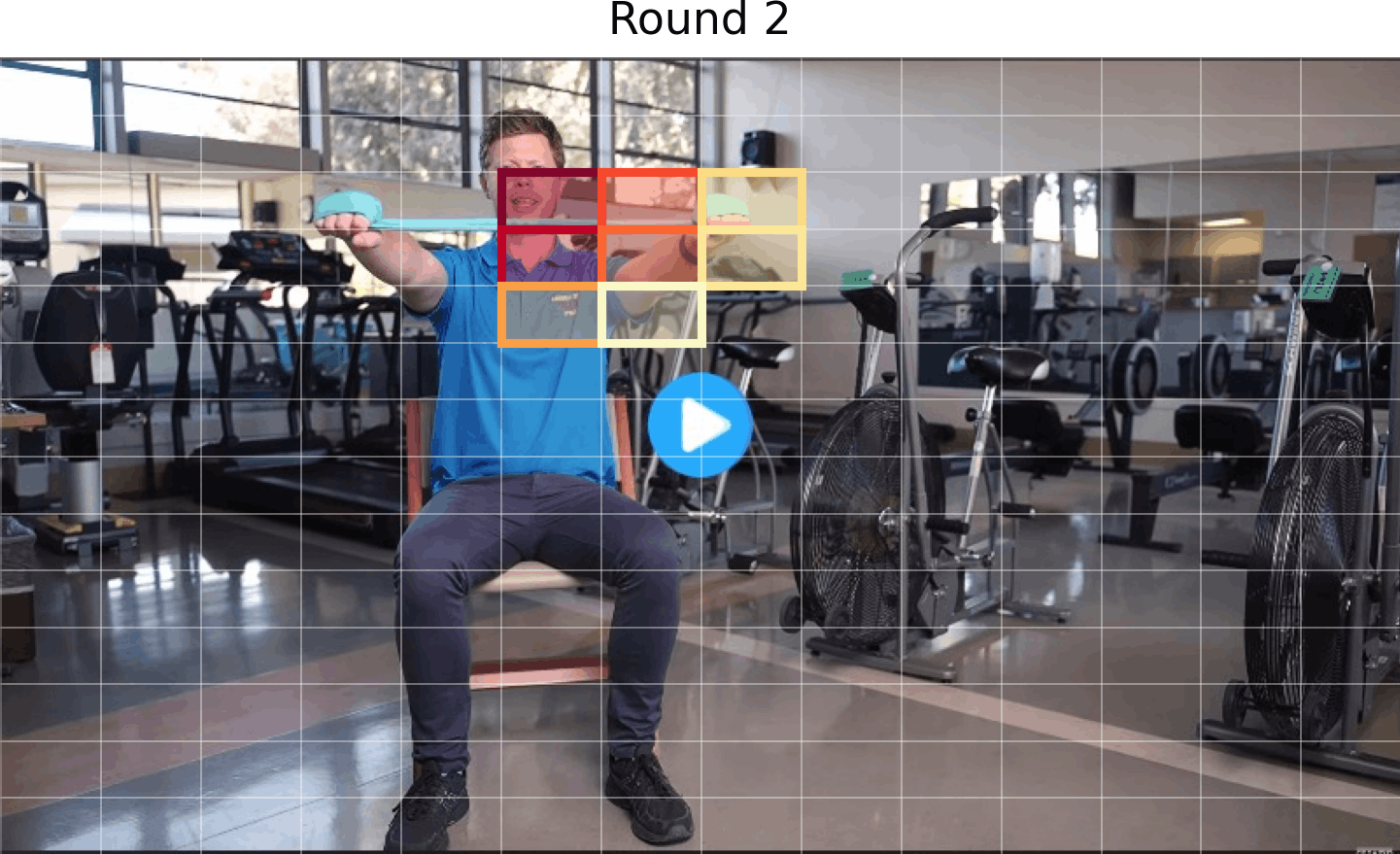}
    
    \vspace{2pt}
    \textbf{Round 2 routing}
\end{minipage}
&
\begin{minipage}[c]{\linewidth}
\textbf{Prefix:} When does the man take a long exhaled breath in this video? 
Options: A) When he is running a marathon. 
B) When he is solving a difficult math problem. 
C) When he is releasing the tension. 
D) When he is pulling apart. 
Answer: The answer is D \\[2pt]
\textbf{Draft block:} When he is pulling \\
\textbf{Accepted prefix:} When \\
\textbf{Rejected suffix:} he is pulling \\
\textbf{Verifier replacement:} . \\
\textbf{Committed output:} When does the man take a long exhaled breath in this video? 
Options: A) When he is running a marathon. 
B) When he is solving a difficult math problem. 
C) When he is releasing the tension. 
D) When he is pulling apart. 
Answer: The answer is D When.
\end{minipage}
\\
\midrule

\begin{minipage}[c]{\linewidth}
    \centering
    \includegraphics[width=0.92\linewidth]{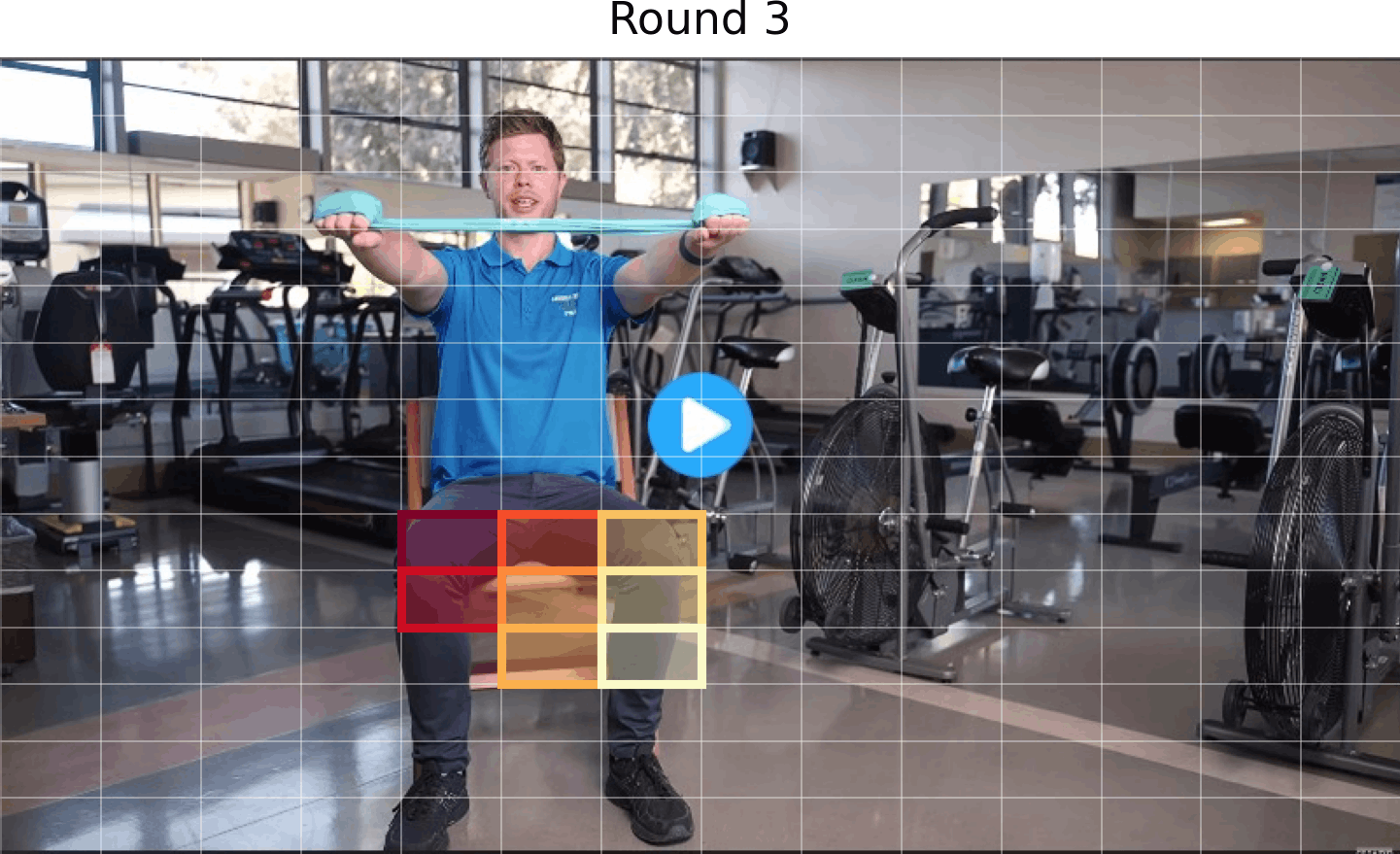}
    
    \vspace{2pt}
    \textbf{Round 3 routing}
\end{minipage}
&
\begin{minipage}[c]{\linewidth}
\textbf{Prefix:} When does the man take a long exhaled breath in this video? 
Options: A) When he is running a marathon. 
B) When he is solving a difficult math problem. 
C) When he is releasing the tension. 
D) When he is pulling apart. 
Answer: The answer is D When. \\[2pt]
\textbf{Draft block:} apart and exhales deeply \\
\textbf{Accepted prefix:} [none] \\
\textbf{Rejected suffix:} apart and exhales deeply \\
\textbf{Verifier replacement:} [EOS] \\
\textbf{Committed output:} When does the man take a long exhaled breath in this video? 
Options: A) When he is running a marathon. 
B) When he is solving a difficult math problem. 
C) When he is releasing the tension. 
D) When he is pulling apart. 
Answer: The answer is D When he is pulling apart. [EOS]
\end{minipage}
\\
\bottomrule
\end{tabular}

\caption{
Execution trace of TIGER on a video reasoning example about exhaled breath while pulling apart.
Each row shows the routed visual tokens selected at one speculative round together with the current prefix, draft block, verifier decision, and committed output.
TIGER initially accepts the answer option, then rejects redundant or unsupported continuation as the verifier enforces the final accepted response.
}
\label{fig:tiger_breath_video_trace}
\end{figure*}

\begin{figure*}[t]
\centering
\scriptsize
\renewcommand{\arraystretch}{1.14}
\setlength{\tabcolsep}{6pt}

\begin{tabular}{m{0.28\textwidth} m{0.67\textwidth}}
\toprule

\multicolumn{2}{l}{\textbf{Question:} Which person placed their leg on the table?} \\
\multicolumn{2}{l}{\textbf{Options:} A) The one on top \quad B) The one on the right \quad C) The one on the left \quad D) The one below} \\
\midrule

\begin{minipage}[c]{\linewidth}
    \centering
    \includegraphics[width=0.92\linewidth]{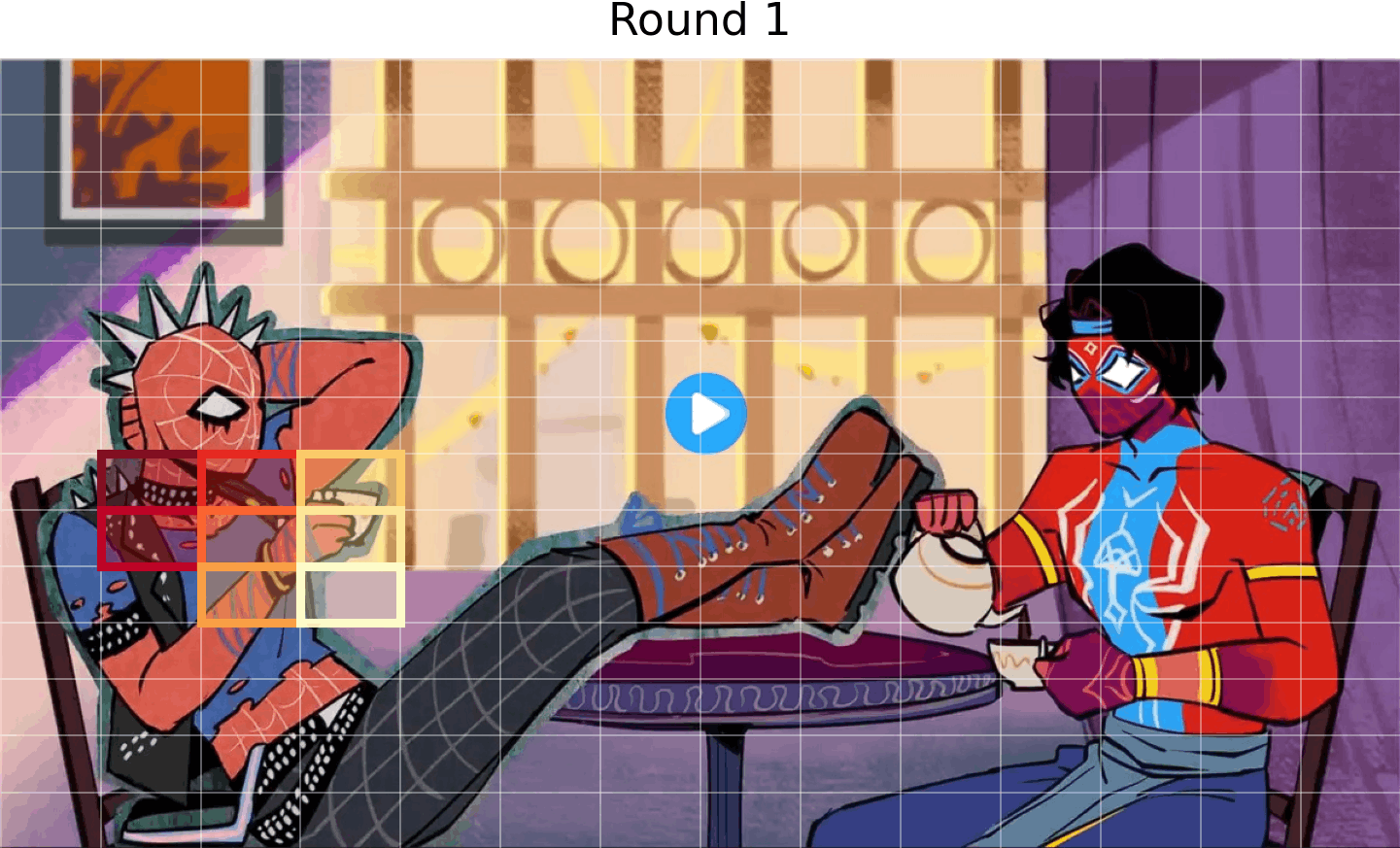}
    
    \vspace{2pt}
    \textbf{Round 1 routing}
\end{minipage}
&
\begin{minipage}[c]{\linewidth}
\textbf{Prefix:} Which person placed their leg on the table? 
Options: A) The one on top. 
B) The one on the right. 
C) The one on the left. 
D) The one below. 
Answer: \\[2pt]
\textbf{Draft block:} The answer is C \\
\textbf{Accepted prefix:} The answer is C \\
\textbf{Committed output:} Answer: The answer is C
\end{minipage}
\\
\midrule

\begin{minipage}[c]{\linewidth}
    \centering
    \includegraphics[width=0.92\linewidth]{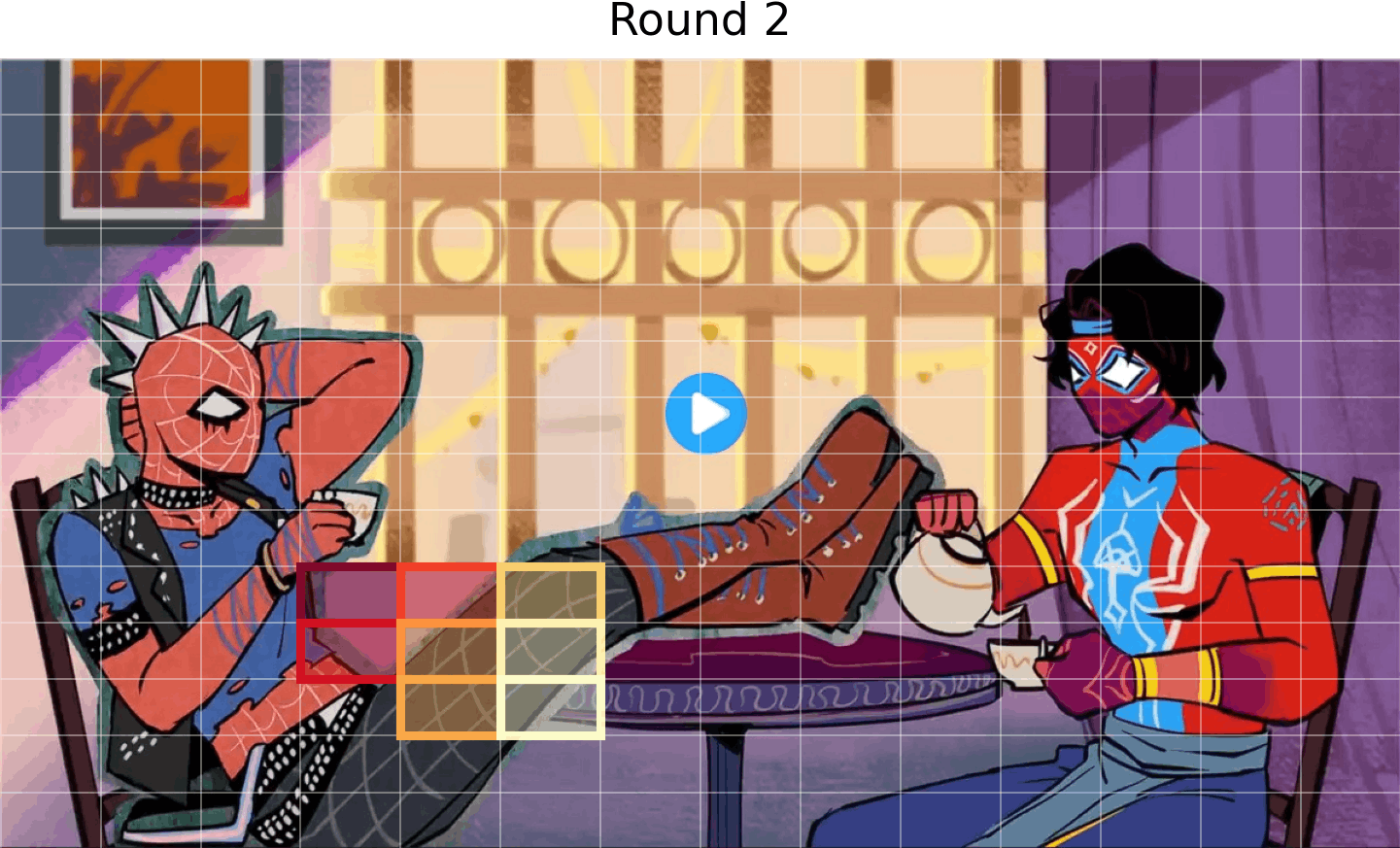}
    
    \vspace{2pt}
    \textbf{Round 2 routing}
\end{minipage}
&
\begin{minipage}[c]{\linewidth}
\textbf{Prefix:} Answer: The answer is C \\[2pt]
\textbf{Draft block:} The one on the left \\
\textbf{Accepted prefix:} The one on the left \\
\textbf{Verifier replacement:} . \\
\textbf{Committed output:} Answer: The answer is C) The one on the left.
\end{minipage}
\\
\midrule

\begin{minipage}[c]{\linewidth}
    \centering
    \includegraphics[width=0.92\linewidth]{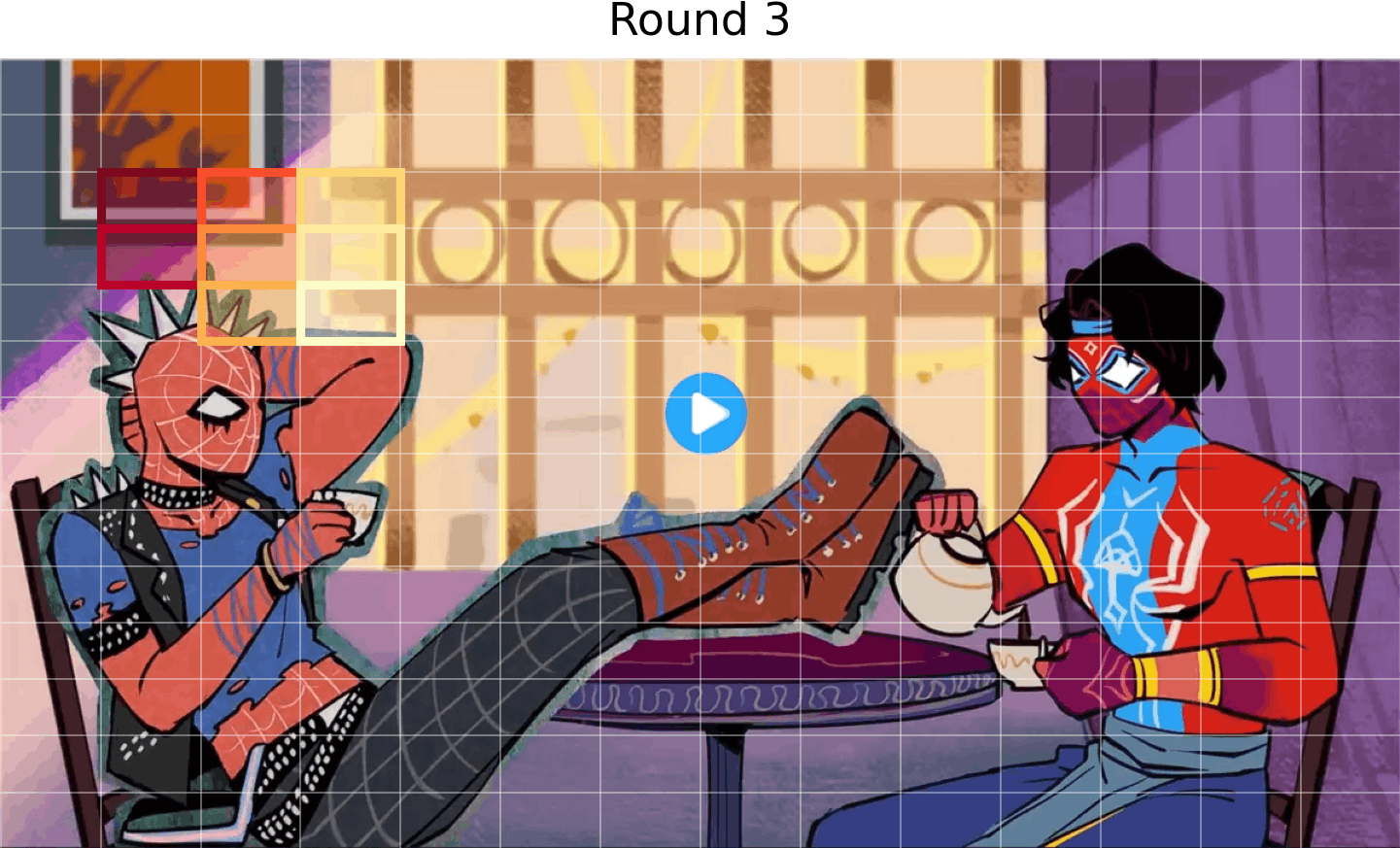}
    
    \vspace{2pt}
    \textbf{Round 3 routing}
\end{minipage}
&
\begin{minipage}[c]{\linewidth}
\textbf{Prefix:} Answer: The answer is C) The one on the left. \\[2pt]
\textbf{Draft block:} their leg resting on \\
\textbf{Accepted prefix:} [none] \\
\textbf{Rejected suffix:} their leg resting on \\
\textbf{Verifier replacement:} [EOS] \\
\textbf{Committed output:} Answer: The answer is C) The one on the left. [EOS]
\end{minipage}
\\
\bottomrule
\end{tabular}

\caption{
Execution trace of TIGER on a visual reasoning example about identifying the person with a leg on the table.
Each row shows the routed visual tokens selected at one speculative round together with the current prefix, draft block, verifier decision, and committed output.
TIGER first accepts the answer option corresponding to the person on the left, then rejects an unnecessary explanatory continuation and terminates generation.
}
\label{fig:tiger_table_leg_trace}
\end{figure*}

We provide qualitative execution traces to illustrate how TIGER's routed visual context evolves across speculative decoding rounds.
All examples are generated under the Qwen3-VL-8B setting and are drawn from a diverse set of benchmarks, including COCO Caption, ScienceQA, MuirBench, and Video-MME.
Figures~\ref{fig:tiger_cinnamon_trace}, \ref{fig:tiger_heron_trace}, \ref{fig:tiger_dogs_multiimage_trace}, \ref{fig:tiger_wolf_trace}, \ref{fig:tiger_breath_video_trace}, and~\ref{fig:tiger_table_leg_trace} cover image captioning, science-oriented visual reasoning, multi-image reasoning, snow-camouflage reasoning, video temporal reasoning, and video spatial identification, respectively.
We also include an additional open-ended visual reasoning example in Figure~\ref{fig:tiger_execution_trace}, where the model answers a question about the emotions evoked by a bouquet of red roses.

Across these examples, TIGER does not rely on a fixed visual subset throughout generation.
Instead, the routed visual tokens are recomputed at each speculative round according to the current textual prefix.
In the COCO Caption example, TIGER routes evidence around the main object and surrounding context to produce a visually grounded caption.
In the ScienceQA examples, TIGER accepts grounded answers such as \emph{Ardea herodias} and \emph{short-tailed weasel}, while rejecting unnecessary explanatory continuation.
In the MuirBench multi-image example, TIGER routes evidence across both images, accepts the answer \emph{Two dogs}, and terminates after the verifier rejects redundant continuation.
In the Video-MME examples, TIGER handles both temporal and spatial visual reasoning: one trace routes evidence around the pulling-apart motion and related body cues, while another identifies the person on the left whose leg is placed on the table.
In the open-ended rose example, TIGER progressively shifts routing toward flower-related regions as the committed response becomes more visually grounded.

These traces illustrate two recurring behaviors.
First, text-conditioned routing adapts the drafter-side visual evidence to the current continuation, rather than relying on globally selected visual tokens.
Second, verifier feedback filters unsupported or redundant draft suffixes, while visually grounded prefixes are committed.
Although qualitative, these examples are consistent with the quantitative results showing that TIGER improves verifier-accepted prefix length by strengthening drafter--verifier agreement on visually grounded continuations.

\subsection{Routing Overhead, Caching, and KV-Cache Reuse}
\label{app:routing-overhead}

TIGER recomputes the routed visual context once per speculative round because the routing query depends on the current textual prefix. However, this does not require re-encoding the image at every round. The vision encoder is executed once per input to obtain the visual token bank \(V\), and the visual-side projections used for routing can be cached because the image remains fixed throughout decoding.

Concretely, for each input image, TIGER first computes and stores the visual token bank \(V\) and the projected visual tokens \(W_v V\). At speculative round \(j\), TIGER only computes the current text-conditioned query \(q_j = W_q h_j\), scores the cached visual tokens, and performs Top-\(k\) selection:
\begin{equation}
s_i = \frac{q_j^\top W_v \bar v_i}{\sqrt d},
\qquad
\mathcal S_j = \operatorname{TopK}(s_1,\dots,s_{r+1};k).
\end{equation}
The selected routed context \(V_j^{\mathrm{route}}\) is then reused for the entire draft block of length \(\gamma\). Thus, routing is performed once per speculative round, not once per drafted token.

This design adds a lightweight routing cost while reducing the drafter-side visual interaction. If \(r\) is the number of full visual tokens and \(k \ll r\) is the routing sparsity, TIGER introduces a routing term \(C_{\mathrm{route}}(r)\), but replaces the drafter-side visual cost \(C_{\mathrm{vis}}(r)\) with \(C_{\mathrm{vis}}(k)\). The routing cost consists mainly of relevance scoring and Top-\(k\) selection, which are much cheaper than re-encoding the image or attending to the full visual token bank throughout drafting.

Importantly, dynamic routing does not change the verifier. The verifier still operates with the full visual token bank \(V\) and follows the standard speculative decoding acceptance protocol. Therefore, verifier-side KV-cache reuse is unchanged from standard speculative decoding. Routing only affects the drafter-side visual interface. On the drafter side, the routed context is fixed within each speculative block and can be reused while generating that block.

The reported end-to-end latency includes the routing step. For longer decoding trajectories, routing overhead scales with the number of speculative rounds rather than the number of generated tokens. Since TIGER increases the accepted prefix length, it reduces the expected number of verifier rounds, which helps amortize the additional routing cost.

\section*{Supplementary Materials Availability}

To facilitate reproducibility, we have included the source code of \textsc{TIGER} in the submission. We have also included the two training datasets developed in this work, namely the KD training mixture and the GRPO training pool, together with data preparation scripts and documentation.

\end{document}